\documentclass{article}

    \usepackage[preprint,nonatbib]{neurips_2024}

\usepackage[utf8]{inputenc} 
\usepackage[T1]{fontenc}    
\usepackage{hyperref}       
\usepackage{url}            
\usepackage{booktabs}       
\usepackage{amsfonts}       
\usepackage{nicefrac}       
\usepackage{microtype}      
\usepackage{xcolor}         

\usepackage{amssymb}
\usepackage{amsmath}
\usepackage{amsthm}

\newcommand{\dualstepsize}{\zeta}
\newcommand{\gcost}{expert} 
\newcommand{\Gcost}{Expert} 
\newcommand{\xmark}{\textcolor{gray!50}{\ding{55}}} 
\newcommand{\sigmoid}{S}
\newcommand\norm[1]{\left\lVert#1\right\rVert}

\usepackage{algorithm}
\usepackage[noend]{algpseudocode}
\newcommand\gy{u}
\def\abs#1{|{#1}|}

\ifodd 0
    \newcommand\rev[1]{{\color{blue}#1}}
    \newcommand{\com}[1]{\textbf{\color{red} (COMMENT: #1)}} 
\else
    \newcommand\rev[1]{{#1}}
    \newcommand{\com}[1]{}
\fi

\newcommand{\covNg}{\mathcal{N}(\mathcal{B}_g,\epsilon,\|\cdot\|_\infty)}

\newcommand{\defeq}{\vcentcolon=}

\newcommand\regvar{r}
\newcommand\fl{f}
\newcommand\gl{g}

\newcommand\lb{\left(}
\newcommand\rb{\right)}
\newcommand\lig{\left\{}
\newcommand\rig{\right\}}
\newcommand\gthr{g_\textrm{thr}}

\theoremstyle{plain}
\newtheorem{theorem}{Theorem}[section]

\newtheorem{lemma}[theorem]{Lemma}

\theoremstyle{definition}

\newtheorem{assump}[theorem]{Assumption}
\newtheorem{example}[theorem]{Example}
\theoremstyle{remark}
\newtheorem{remark}[theorem]{Remark}

\newcommand{\E}{\mathbb{E}}
\newcommand\ag{1}
\newcommand\ft{f_t}
\newcommand\fT{f_T}
\newcommand\gt{g_t}
\newcommand\gtp{g_{t+1}}
\newcommand\fg{\tilde{g}}
\newcommand\betag{\alpha}
\newcommand\tg{g}

\newcommand\Gnota{Q}
\newcommand\GT{\mathcal{\Gnota}^f_T}
\newcommand\aGT{|\mathcal{\Gnota}^f_T|}
\newcommand\Gt{\mathcal{\Gnota}^f_t}
\newcommand\Gtm{\mathcal{\Gnota}^f_{t-1}}
\newcommand\aGt{|\mathcal{\Gnota}^f_t|}
\newcommand\aGtm{|\mathcal{\Gnota}^f_{t-1}|}

\newcommand\Cnota{Q}
\newcommand\CT{\mathcal{\Cnota}^g_T}
\newcommand\CoT{\mathcal{C}_T}
\newcommand\aCT{|\mathcal{\Cnota}^g_T|}
\newcommand\Ct{\mathcal{\Cnota}^g_t}
\newcommand\Ctm{\mathcal{\Cnota}^g_{t-1}}
\newcommand\aCt{|\mathcal{\Cnota}^g_t|}

\newcommand\CTc{\Cnota^g_T}

\newcommand\one{\mathbf{1}}
\newcommand{\onetau}{\one_\tau}

\newcommand{\Prob}{\mathbb{P}}
\usepackage[acronym]{glossaries} 
\usepackage{pifont}              
\usepackage{paralist}            
\usepackage{comment}             
\usepackage{mathtools}           
\usepackage{multirow}            
\usepackage{wrapfig}             
\usepackage{caption}             
\usepackage{minitoc}             
\usepackage{wrapfig}             
\usepackage{subfig}              
\newcommand{\bse}{\begin{subequations}}
\newcommand{\ese}{\end{subequations}}

\newcommand\Hil{\mathcal{H}}

\usepackage{todonotes}

\title{Principled Bayesian Optimisation\\ in Collaboration with Human Experts}

\author{
    Wenjie Xu\thanks{Equal contribution}$\,\,\,^{, 1, {2}}$, 
    Masaki Adachi$^{*, 3, 4}$, 
    Colin N. Jones$^{1}$,
    Michael A. Osborne$^{3}$,\\
    \small{$^1$ Automatic Control Laboratory, EPFL}\quad
     \small{$^{2}$ Urban Energy Systems Laboratory, Empa}\\
    \small{$^3$Machine Learning Research Group, University of Oxford}\\
    \small{$^4$ Toyota Motor Corporation}\\
    \small{\texttt{\{wenjie.xu, colin.jones\}@epfl.ch}, \texttt{\{masaki, mosb\}@robots.ox.ac.uk}}\\
}

\begin{document}

\maketitle

\begin{abstract}
  Bayesian optimisation for real-world problems is often performed interactively with human experts, and integrating their domain knowledge is key to accelerate the optimisation process. We consider a setup where experts provide advice on the next query point through binary accept/reject recommendations (labels). Experts' labels are often costly, requiring efficient use of their efforts, and can at the same time be unreliable, requiring careful adjustment of the degree to which any expert is trusted. We introduce the first principled approach that provides two key guarantees. (1) Handover guarantee: similar to a no-regret property, we establish a sublinear bound on the cumulative number of experts' binary labels. Initially, multiple labels per query are needed, but the number of expert labels required asymptotically converges to zero, saving both expert effort and computation time. (2) No-harm guarantee with data-driven trust level adjustment: our adaptive trust level ensures that the convergence rate will not be worse than the one without using advice, even if the advice from experts is adversarial. Unlike existing methods that employ a user-defined function that hand-tunes the trust level adjustment, our approach enables data-driven adjustments. Real-world applications empirically demonstrate that our method not only outperforms existing baselines, but also maintains robustness despite varying labelling accuracy, in tasks of battery design with human experts.
\end{abstract}

\doparttoc 
\faketableofcontents 
\section{Introduction}
Bayesian optimisation (BO) \cite{mockus1978application, osborne2009gaussian, garnett2023bayesian} is a successful approach to black-box optimiation that has been applied across a wide array of applications. BO is often praised for `taking the human out of the loop' \cite{shahriari2015taking} by automating laborious optimisation processes, such as hyperparameter optimisation \cite{feurer2015efficient, wu2020practical} and neural architecture search \cite{ru2020interpretable, white2021bananas}, thus relieving human users from these tasks. Nonetheless, a growing trend involves the opposite direction, which brings humans back into the loop and leverages human expertise as an adviser to the optimiser \cite{adachi2023looping}. This human-in-the-loop approach is particularly relevant to scientific and explorative tasks, such as materials discovery \cite{cisse2023hypbo, adachi2021high} and product design \cite{kanarik2023human, jetton2023constraining, adachi2023looping}. Experts have accumulated domain knowledge and should be helpful in accelerating the optimisation process, yet their experience and knowledge are often qualitative---they can struggle to express their knowledge in a functional form or to pinpoint the best candidates as an absolute quantity \cite{kahneman1979interpretation}. At the forefront of science, experts are also in the middle of trial and error; demanding well-defined and error-free inputs can limit the applicable range of BO. As such, a human-AI collaborative setting in BO has emerged, driven by practical demands, and has been gaining popularity in the literature \cite{av2022human, hvarfner2022pi, gupta2023bo, khoshvishkaie2023cooperative, cisse2023hypbo, adachi2023looping, rodemann2024explaining, av2024enhanced, hvarfner2024a}.

A prevalent issue in this domain is the lack of both shared assumptions and theoretical guarantees, making fair comparisons challenging. 
Our community has yet to reach a consensus on acceptable assumptions, particularly in the following areas. \textbf{(a) The level of effectiveness of experts' knowledge:} assuming near oracle-like knowledge,
e.g. in \cite{av2022human, gupta2023bo, av2024enhanced}, collaborative settings can significantly surpass vanilla BO. However, if experts are entirely erroneous (yet confident)---which can happen  \cite{hvarfner2022pi, khoshvishkaie2023cooperative, cisse2023hypbo, adachi2023looping}---overreliance on experts' input cannot guarantee the global optimum convergence.
\textbf{(b) Human interaction method:} ideally, humans prefer minimising interaction with machines for convenience. Minimising interaction leads to maximising the information at each query to human, which often ends up requesting error-free and quantitative information for humans \cite{souza2021bayesian, av2022human, hvarfner2022pi, hvarfner2024a}. However, accurate knowledge elicitation remains a long-standing quest \cite{shadbolt2015knowledge, o2019expert, mikkola2021prior}. Inversely, when we assume human belief is also a black-box function and require the elicitation of the belief function through statistical modelling, e.g. \cite{rousseau2001schema, garthwaite2005statistical, adachi2023looping}, we will demand excessive queries of the experts.

\textbf{Contributions.} The contributions of this paper are summarised below:
\begin{compactenum}
    \item  \textbf{Handover guarantee}: we model the expert's role as cognitively simple and qualitative---the expert serves as a black-box classifier, providing binary labels on the desirability of the next query location. Similar to the no-regret property, we establish a sublinear bound on the cumulative number of binary labels needed. Initially, multiple labels per query are needed, but the frequency of querying binary labels asymptotically converges to zero, thus saving both expert effort and computation time.
    \item \textbf{No-harm guarantee}: we show that the convergence rate of our expert-advised algorithm will not be worse than that of vanilla BO (i.e. without expert advice), even if the advice from experts is adversarial. Our convergence is achieved through data-driven trust level adjustments, and is unlike existing methods that rely on hand-tuned user-defined functions.
    \item \textbf{Real-world contribution}: empirically, our algorithm provides both fast convergence and resilience against erroneous inputs. It outperformed existing methods in both popular synthetic, and new real-world, tasks in designing lithium-ion batteries.
\end{compactenum}

\section{Problem Statement}
\begin{wrapfigure}{r}{0.48\textwidth}
  \vspace{-1em}
  \includegraphics[width=0.48\textwidth]{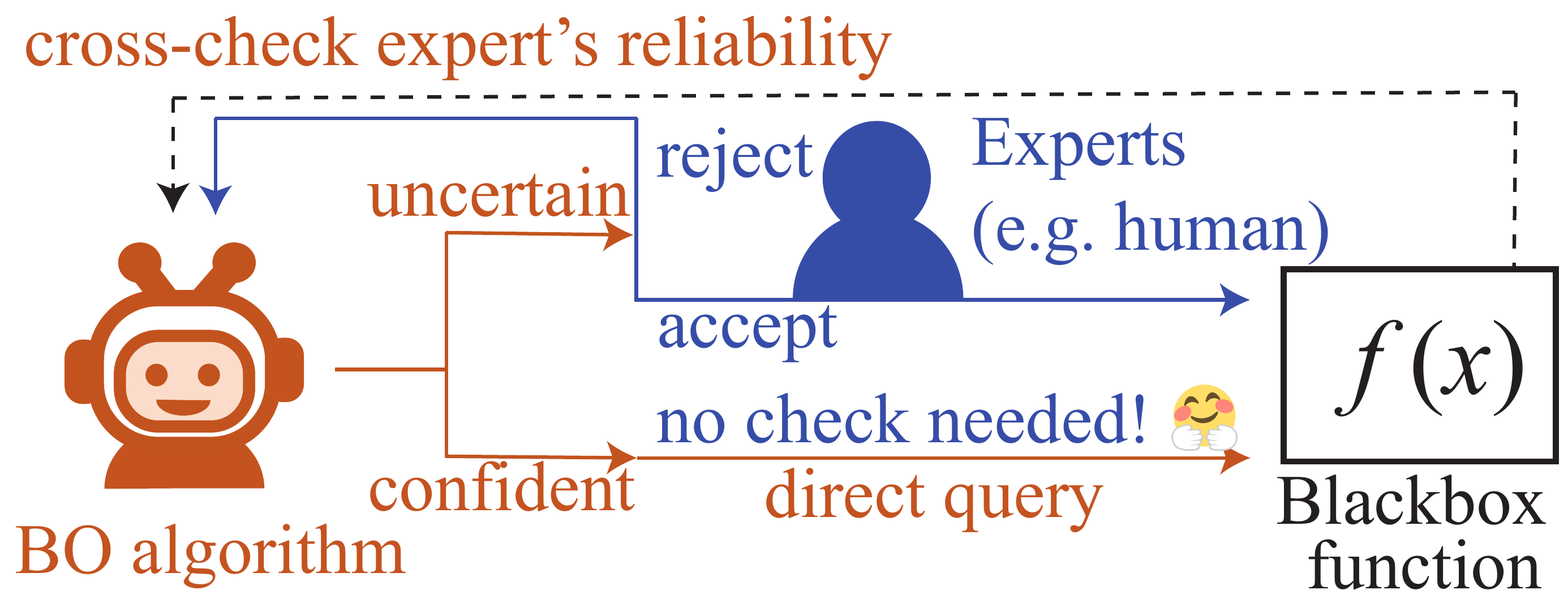}
  \caption{\textbf{BO-expert collaboration framework}: The algorithm (red) decides if an expert's (blue) label is necessary. If rejected, it generates a different candidate; otherwise, it directly queries.}
  \label{fig:concept}
\end{wrapfigure}
We address the black-box optimization problem,
\begin{equation}
    x^\star \in \arg\min_{x\in\mathcal{X}}\; f(x)\;, \label{eq:prob_to_solve}
\end{equation}
while collaborating with an expert, where $\mathcal{X}\subset\mathbb{R}^d$ and $d$ is the dimension. 

\textbf{Expert labelling model.} 
We model an expert as a binary labeller (see Fig.~\ref{fig:concept}). An expert labels a point $x\in\mathcal{X}$ as either `accept' or `reject'. An `accept' label indicates that the point is worth sampling, while `reject' label indicates it is not. These labels are binary, with $0$ for `accept' and $1$ for `reject'. 
In practice, the labelling process can be noisy, since humans may find some points hard to \rev{classify}. Non-expert or incorrect belief may label the optimum $x^\star$ `reject'. The distribution of the labels is determined by the expert's prior belief about the black-box function $f$, and we model the expert's belief through another unknown black-box function $g$.
\begin{assump}
   \label{assump:label_model}
   The notation $x\succ_g 0$ denotes the event where $x$ is labelled as `reject', based on the expert's belief function $g$. Additionally, the random indicator $\one_{x\succ_g 0}\in\{0,1\}$ takes value $1$ if $x\succ_g 0$ and $0$ otherwise. The probability distribution of $\one_{x\succ_g 0}\in\{0,1\}$ follows the Bernoulli distribution with $\Prob(\one_{x\succ_g 0}=1)=p_{x\succ_g0}=\sigmoid(g(x))$, where $\sigmoid(u)=\nicefrac{1}{(1+e^{-u})}$ is the sigmoid function. 
\end{assump}
\begin{example}\label{example:g_func}
Let us define an example `synthetic' expert's labelling response as $p_{x\succ_g0} = \sigmoid(a \rho(f(x)))$, where $a$ is the accuracy coefficient and $\rho$ is the linear scaling function from bound $[\min_{x \in \mathcal{X}}f(x), \max_{x \in \mathcal{X}}f(x)]$ to $[-3, 3]$. When $a=1$, $\rho(f(x^*)) = -3, \sigmoid(-3)\approx 0.05$, resulting in a Bernoulli distribution that yields an acceptance label of 0 with a 95\% chance at the global minimum $x^*$. In this case, the sharpness of the belief $p_{x\succ_g0}$ is influenced by both the shape of $f(x)$ and $a$; if $f(x)$ is peaky or $a \gg 1$, the expert can nearly pinpoint $x^*$.
\end{example}

However, in reality, the expert does not know the exact true $f$ and therefore, we consider $g$ to be a `subjective' belief function representing $f$. This differs from a typical surrogate model $\hat{f}$ of $f$, which infers an `objective' belief function from oracle queries. If $g$ has better predictive ability than the surrogate model $\hat{f}$, exploiting $g$ can accelerate convergence; otherwise it may decelerate the process. In the optimisation process, $g$ \rev{may} act as \rev{a} \rev{regularizer} function in addition to the objective function $f$. \rev{
For simplicity, we use this Ex.~\ref{example:g_func} as synthetic human feedback. Readers interested in other examples are encouraged to refer to Appendix~\ref{app:feedback}.
}
\begin{assump}\label{assump:support_set}
$\mathcal{X}$ is compact and non-empty.
\end{assump}
Assumption~\ref{assump:support_set} is reasonable because in many applications~(e.g., continuous hyperparameter tuning) of BO, we are able to restrict the optimisation into certain ranges based on domain knowledge.
Regarding the black-box function $f$ and the \rev{function} $g$, we assume that,
\begin{assump}\label{assump:bounded_norm}
$f\in\Hil_{k_f}, g\in\Hil_{k_g}$, where $k:\mathbb{R}^d\times\mathbb{R}^d\to \mathbb{R}$, representing $k_f$ or $k_g$, is a symmetric, positive-semidefinite kernel function, and $\Hil_{k}$ is its corresponding reproducing kernel Hilbert space~(RKHS, see~\cite{scholkopf2001generalized}). Furthermore, we assume $\|f\|_{k_f}\leq B_\fl$ and $\|g\|_{k_g}\leq B_g$, where $\|\cdot\|_{k}$ is the norm induced by the inner product in the corresponding RKHS $\Hil_{k}$. We use $\mathcal{B}_g$ to denote the set $\lig\fg\in\Hil_{k_g} \mid \|{\fg}\|_{k_g}\leq B_g\rig$.
\end{assump}
Assumption~\ref{assump:bounded_norm} requires that the objective $f$ and the \rev{function} $g$ are regular in the sense that they have bounded norms in the corresponding RKHS, which is a common assumption.
\begin{assump}
\label{assump:kernel_bound}
$k(x,x^\prime)\leq 1$, $x,x^\prime\in \mathcal{X}$, and $k(x,x^\prime)$ is continuous on $\mathbb{R}^d\times \mathbb{R}^d$. 
\end{assump}
\begin{assump}
\label{assump:obj_obs}
At step $t$, if query point $x_t$ is evaluated, we get a noisy evaluation of $f$ (we refer to an oracle query),
$y_{t}=f(x_t)+\xi_{t}\, ,$
where $\xi_{t}$ is i.i.d $\sigma$-sub-Gaussian noise with fixed $\sigma>0$.
\end{assump}
\textbf{Notation}. We refer to $\onetau$ as data realisation of $\one_{x_\tau\succ_g 0}$ at step $\tau$. We denote the following sequences of steps: iterations as $[t] := \{1, 2, \cdots, t\}$, $f$ queries as $\mathcal{Q}_t^f \defeq \{\tau \in [t-1] \mid \text{if} f \text{ is queried in step }\tau \}$, and expert queries as $\mathcal{Q}_t^g$, respectively $\big(t \geq |\mathcal{Q}_t^g|, t \geq |\mathcal{Q}_t^f| \big)$. We use capitals, e.g. $X_{\mathcal{Q}^f_t}$, for the set $(x_\tau)_{\tau \in \mathcal{Q}^f_t}$.

\section{Confidence Set of the Surrogate Models}
We introduce surrogate models for the objective $f$ and the \rev{function} $g$. We opted for a Gaussian process (GP; \cite{stein1999interpolation, williams2006gaussian}) for $f$ and the likelihood ratio model \cite{owen1990empirical, emmenegger2023likelihood} for $g$.

\subsection{Surrogate Model of the Objective \texorpdfstring{$f$}{}: Gaussian Process}
\textbf{Definitions.} We employ a zero-mean GP regression model, with predictive posterior $\tilde{f}_t \mid D_t^f \sim \mathcal{GP}(\mu_{\ft}, \sigma^2_{\ft})$, 
\bse
\label{eq:mean_cov}
\begin{align}
\mu_{\ft}(x) &=k_f(X_{\Gt}, x)^\top\left(K_{\Gt}+\regvar I\right)^{-1} Y_{\Gt},\\
\sigma^2_{\ft}\left(x\right) &=k_f \left(x, x\right)-k_f(X_{\Gt}, x)^\top\left(K_{\Gt}+ \regvar I\right)^{-1} k_f\left(X_{\Gt}, x\right), \label{eq:f_sigma_def}
\end{align}
\ese
where $K_{\Gt}=\big( k_f(x_{\tau_1},x_{\tau_2}) \big)_{\tau_1,\tau_2\in \Gt}$, $D_t^f := (X_{\Gt}, Y_{\Gt})$, $r$ is the regularisation term \cite{mohammadi2016analytic}.\footnote{We follow the definition from \cite{chowdhury2017kernelized}.} The maximum information gain \cite{srinivas2012information} for the objective $f$ is,
\begin{equation}
\label{eq:max_inf_gain}
\gamma_{\aGt}^f:=\max_{X \subset \mathcal{X};\, |X|=|\Gt|} \frac{1}{2} \log \left|I+\regvar^{-1}K_{f,X}\right|, \quad\text{where}\quad K_{f,X} :=(k_f(x,x^\prime))_{x,x^\prime\in X}.
\end{equation}

\begin{lemma}[Theorem 2,~\cite{chowdhury2017kernelized}]
Let Assumptions~\ref{assump:support_set}, \ref{assump:bounded_norm} and \ref{assump:obj_obs} hold. For {any $\delta\in(0, 1)$}, with probability at least $1-\delta/2$, the following holds for all $x \in \mathcal{X}$ and $1\leq t \leq T$, $T\in\mathbb{N}$, 
\begin{equation*}
\left|\mu_{\ft}(x)-f(x)\right| \leq \beta_{\ft} \sigma_{\ft}(x), \quad\,\, \beta_{\ft} := \left(B_f+\sigma \sqrt{2\left(\gamma_{\aGtm}^f+1+\ln (2 / \delta)\right)}\right),
\end{equation*}
where $\mu_{\ft}(x), \sigma_{\ft}(x)$ and $\gamma^f_{\aGtm}$ are as given in Eq.~\eqref{eq:mean_cov} and Eq.~\eqref{eq:max_inf_gain}, and $\gamma^f_{0}=0$.
\label{lem:conf_int} 
\end{lemma}

For brevity, we denote the lower/upper confidence bound (LCB/UCB) functions $\underline{f}_t(x)$ and $\bar{f}_t(x)$ as,
\begin{equation*}
 \underline{f}_t(x)=\mu_{\ft}(x)-\beta_{\ft}\sigma_{\ft}(x)\;,\quad\quad \bar{f}_t(x)=\mu_{\ft}(x)+\beta_{\ft}\sigma_{\ft}(x).
\end{equation*}

\subsection{Surrogate Model of the \rev{\Gcost} Function \texorpdfstring{$g$}{}: Likelihood Ratio Model}
While a GP classifier \cite{nickisch2008approximations} is a popular choice, we opted for likelihood ratio model \cite{owen1990empirical, emmenegger2023likelihood}. The combination of a Gaussian prior with a Bernoulli likelihood in GP models presents challenges in estimating the posterior confidence bound both theoretically and computationally. Moreover, GPs assume strong rankability \cite{guo2010gaussian, chu2005preference}, presuming humans can rank their preferences accurately in all cases, which often leads to inconsistent results \cite{chau2022learning}.
To address these issues, we drew inspiration from classic expert elicitation methods using imprecise probability theory \cite{augustin2014introduction, hullermeier2021aleatoric}. Instead of estimating the predictive distribution, we estimate the `interval' of the worst-case prediction only. This approach does not assume any distribution within the interval, thereby relaxing the rankability assumption \cite{senge2014reliable}. This method is particularly well-suited to the GP-UCB algorithm \cite{srinivas2009gaussian}, which only requires a confidence interval. We developed a kernel-based method to provably estimate the predictive interval.

\textbf{Definitions.} First, we introduce the function, $p_{\hat{g}}(x_\tau, \onetau)\defeq \onetau\sigmoid \left( \hat{g}(x_\tau) \right) + (1-\one_\tau) \left[1-\sigmoid(\hat{g}(x_\tau))\right]$, 
which is the likelihood of $\hat{g}$ over the event when $\one_{x_\tau\succ_g0}=\one_\tau$ under the Assumption~\ref{assump:label_model}, 
and $\hat{g}$ is an estimate function of $g\in\Hil_{k_g}$ under the Assumption~\ref{assump:bounded_norm}. We can then derive the likelihood function of a fixed function $\hat{g}$ over the historical dataset $\mathcal{D}^g_t\defeq\{(x_\tau, \onetau)\}_{\tau\in\Ct}$, which becomes the product,
$\Prob_{\hat{g}}((x_\tau,\onetau)_{\tau\in\Ct})\defeq \prod_{\tau\in\Ct}p_{\hat{g}}(x_\tau,\onetau)$. 
The log-likelihood (LL) function, 
\rev{\begin{equation}
   \text{LL value:}\;\; \ell_t(\hat{g})\defeq\log\Prob_{\hat{g}}((x_\tau, \onetau)_{\tau\in\Ct}),
    \end{equation} 
    } 
reduces to $\ell_t(\hat{g}) =\sum_{\tau\in\Ct} z_\tau\one_\tau-\sum_{\tau\in\Ct}\log\left(1+e^{z_\tau}\right)$, 
where $z_\tau=\hat{g}(x_\tau)$ (this equality can be checked as correct for either $\onetau=1$ or $\onetau=0$). We then introduce the maximum likelihood estimator~(MLE), 
$\hat{g}^\mathrm{MLE}_{t}\in\arg\max_{\fg\in\mathcal{B}_g}\log\Prob_{\fg}((x_\tau,\onetau)_{\tau\in\Ct})$.
Similar to~\cite{liu2023optimistic,emmenegger2023likelihood,xu2024principled}, the \emph{confidence set} can be derived as shown in Lemma~\ref{thm:conf_set_g}.

\begin{lemma}[\textbf{Likelihood-based confidence set}]
\label{thm:conf_set_g}
$\forall \epsilon, \delta>0$, let,
\begin{equation*}\mathcal{B}^{t+1}_g\defeq\lig\tilde{g}\in\mathcal{B}_g \mid \ell_t(\fg)\geq\ell_t(\hat{g}^{\mathrm{MLE}}_t)-\betag_\ag(\epsilon, \delta, \aCt, t)\rig,
\end{equation*}
where $\betag_\ag(\epsilon, \delta, \aCt, t)\defeq\sqrt{32\aCt B_g^2\log\frac{\pi^2t^2\covNg}{6\delta}}+2\epsilon t.$
We have, 
\begin{equation*}
\Prob\left(\tg\in\mathcal{B}^{t+1}_g,\forall t\geq1\right)\geq1-\delta.
\end{equation*}
\end{lemma}
The proof is in Appendix~\ref{proof:conf_set_g}. 
As introduced in Assumption~\ref{assump:bounded_norm}, 
while the function $g$ was originally in a broader set of RKHS functions $g \in \mathcal{B}_g$, it is now in a smaller set defined as 
$g \in \mathcal{B}_g^{t+1}$ conditioned on the expert labels $\mathcal{D}^g_t$.
Intuitively, with limited data, the MLE may be imperfect. Hence, it is reasonable to suppose that $\mathcal{B}_g^{t+1}$, bounded by LL values `slightly worse' than the MLE, contains the ground truth with high probability.

\begin{remark}[\textbf{Choice of $\epsilon$}]
In Lemma.~\ref{thm:conf_set_g}, $\betag_\ag(\epsilon, \delta,\aCt, t)$ depends on a small positive value $\epsilon$. It will be seen that $\epsilon$ can be selected to be $\nicefrac{1}{T}$, where $T$ is the running horizon of the algorithm.  \end{remark}
\begin{remark}[\textbf{Confidence bound}]
By Lemma~\ref{thm:conf_set_g}, we define the pointwise confidence bound for unknown $g \in \Hil_{k_g}$, 
$\underline{g}_t(x)\leq g(x)\leq \bar{g}_t(x)$, 
where $\underline{g}_t(x)\defeq\inf_{\fg\in\mathcal{B}^t_g}\fg(x)$ and $\bar{g}_t(x)\defeq\sup_{\fg\in\mathcal{B}^t_g}\fg(x)$.
\end{remark}
\begin{remark}[\textbf{Pointwise predictive interval estimation}]
At a given prediction point $x$, the predictive interval $\big[ \underline{g}_t(x), \,\, \bar{g}_t(x) \big]$ can be estimated through two individual finite-dimensional optimisation problems~(See Appendix~\ref{app_sec:comp_conf_g} for details). Subsequently, applying the sigmoid function yields the predictive interval in probability space $\big[ \sigmoid(\underline{g}_t(x)), \,\, \sigmoid(\bar{g}_t(x)) \big]$ (see Fig.~\ref{fig:demo} for visualisation).
\end{remark}

\section{Algorithm and Theoretical Guarantees}
\subsection{Mixing Two Surrogate Models \texorpdfstring{$f$}{} and \texorpdfstring{$g$}{} via Primal-Dual Method}\label{sec:primal_dual}
\begin{figure}[t]
  \centering
  \includegraphics[width=0.8\hsize]{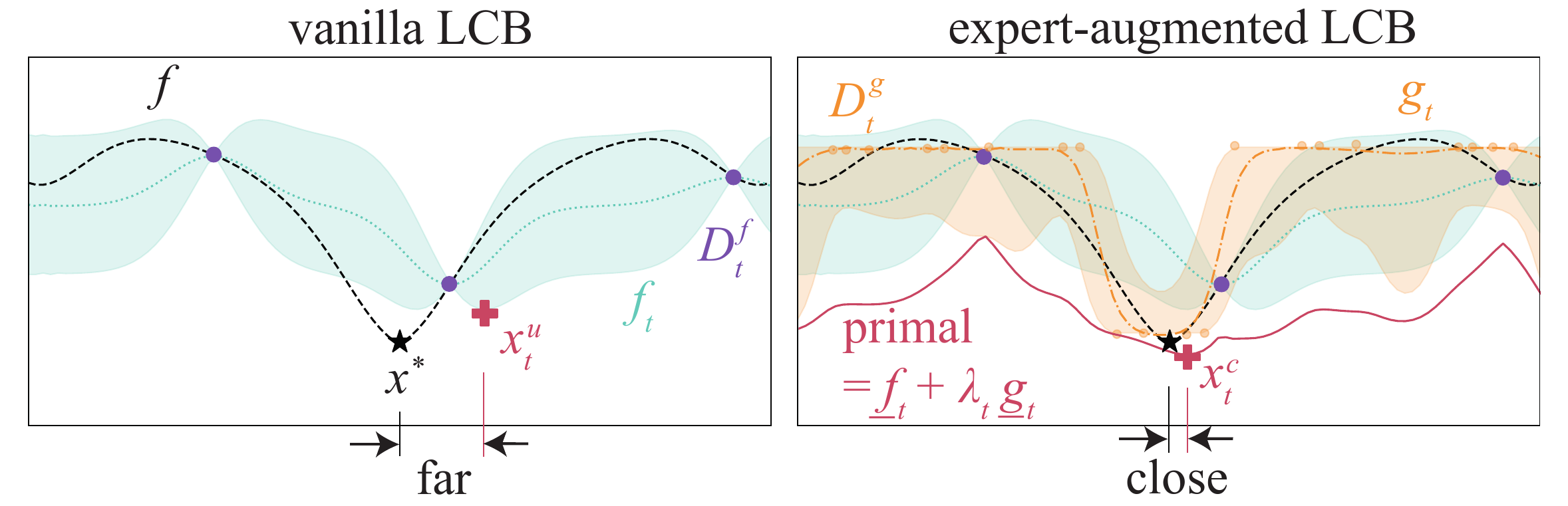}
  \caption{Visual explanation: While the vanilla LCB returns $x^u_t$, a far point from global minimum $x^*$, expert-augmented LCB can successfully navigate to closer point $x^c_t$ by mixing $f_t$ and $g_t$ \rev{with $\underline{f}_t+\lambda_t\underline{g}_t$, where $\lambda_t$ is the dual variable}. \rev{In the figure, $D_t^f$ is the set of the sample points of the objective function $f$ and $D_t^g$ is the set of human feedback.}}
  \label{fig:demo}
  \vspace{-1em}
\end{figure}
\textbf{Primal dual.} 
We introduce the following primal-dual problem~\eqref{eq:primal_dual} as our acquisition policy,
\begin{equation}\label{eq:primal_dual}
\textbf{Primal}: x^c_t\in\arg\min_{x\in\mathcal{X}}\underline{f}_t(x)+\lambda_t\underline{g}_t(x), \textbf{ Dual}: \lambda_{t+1}=[\lambda_t + \rev{\dualstepsize} \underline{g}_t(x_t^c)]^+, 
\end{equation}
where $\lambda_t$ is the primal-dual weight at the $t$-th iteration and $\rev{\dualstepsize}$ is the step size for dual update. See Fig.~\ref{fig:demo} for the intuition: we prioritise the sample in the expert-preferred region (i.e., the region with small $\underline{g}_t(x)$). The primal-dual method is a classical algorithm for constrained optimisation~\cite{nocedal1999numerical} and has recently been applied to, for example, the constrained bandit problem~\cite{zhou2022kernelized}. In terms of constrained optimisation,  Prob.~\eqref{eq:primal_dual} can be understood as solving 
$\min_{x \in \mathcal{X}} \underline{f}(x) \,\, \text{s.t.} \,\, \underline{g}_t(x)\leq0$.  Interestingly, the primal-dual approach is also roughly analogous to Bayesian inference \cite{dai2018coupled}. Just as the prior acts as a regulariser to the LL maximiser \cite{vapnik1998statistical}, expert belief $\underline{g}_t(x)$ regularises the $\underline{f}_t(x)$ minimiser. 
More specifically, the weight $\lambda_{t+1}$ increases when $\underline{g}_t(x_t^c)>0$; otherwise, $\lambda_{t+1}$ decreases. The condition $\underline{g}_t(x_t^c)>0$ indicates that the primal solution $x_t^c$ is more likely to be rejected.\footnote{Recall that $\sigmoid({g(x_t^c)}) > \sigmoid(0) = 0.5$ implies a higher chance of rejection than random (=0.5).} Under such a risk of rejection, increasing the weights $\lambda_{t+1}$ is natural because it more strongly regularises the $\underline{f}_t$ minimiser to enhance feasibility in the next round, and vice versa.

\textbf{Level of trust.} Note that the primal-dual method is not the primary reason we achieve the no-harm guarantee. Indeed, its proof (detailed later) does not rely on the primal-dual formulation. Therefore, technically speaking, our algorithm could employ a more aggressive exploitation of $g_t$  (e.g., simply minimising $g_t$). Nevertheless, the primal-dual approach is our recommended policy for generating the expert-augmented candidate $x^c_t$ to enhance resilience to erroneous inputs.
The initial level of trust on $g_t$ is determined by the initial weight $\lambda_0$, where larger $\lambda_0$ values correspond to greater trust in the expert. We compared the effect of $\lambda_0$ in the later experimental section.

\textbf{Efficient computation.} 
Leveraging the representer theorem \cite{scholkopf2001generalized, xu2024principled} due to the RKHS property, we further reformulate Prob.~\eqref{eq:primal_dual} to a $(|\Ct| + d + 1)$-dimensional, tractable optimisation problem~\eqref{eqn:reform_inner_prob_to_fin}.

\begin{equation}
\begin{aligned}
    \label{eqn:reform_inner_prob_to_fin}
\min_{Z_{\Ct}\in\mathbb{R}^{|\Ct|}, \,\, z\in\mathbb{R}, \,\, x\in\mathcal{X}}&\quad\underline{f}_{t}(x)+\lambda_t z \\
   \text{subject to}&\quad \left[\begin{array}{l}
Z_{\Ct} \\
z
\end{array}\right]^{\top}K_{\Ct,x}^{-1}\left[\begin{array}{l}
Z_{\Ct} \\
z
\end{array}\right]   \leq B_g^2, \\
   &\quad \ell(Z_{\Ct} \mid \mathcal{D}^g_t)\geq \ell_t(\hat{g}^\mathrm{MLE}_t)-\betag_1(\epsilon, \delta, \aCt, t),
\end{aligned}
\end{equation}
where $K_{\Ct,x} := (k_g(\tilde{x}, \tilde{x}^\prime))_{\tilde{x}, \tilde{x}^\prime\in X_{\Ct}\cup\{x\}}$, and $\ell(Z_{\Ct} \mid \mathcal{D}^g_t)=\sum_{\tau\in\Ct} Z_\tau\one_\tau-\sum_{\tau\in\Ct}\log\left(1+e^{Z_\tau}\right)$ is the LL value when $\hat{g}(x_\tau) = Z_\tau$, 
$\forall\tau\in\Ct$. We update $\lambda_{t+1}=\lambda_t +\rev{\dualstepsize} z^\star$, where $z^\star = \underline{g}_t(x_t^c)$ is the optimal $z$ of Prob.~\eqref{eqn:reform_inner_prob_to_fin}.

\textbf{Key hyperparameter estimation.} 
A key hyperparameter in Prob.~\eqref{eqn:reform_inner_prob_to_fin} is the norm bound $B_g$ in the first constraint. Another hyperparameter, $\alpha_\ag$, in the second constraint, also scales with $B_g$, (see Lemma~\ref{thm:conf_set_g}). However, $B_g$ may be unknown in practice, and its mis-specification leads to mis-calibrated uncertainty. We estimate $B_g$ by starting with a small initial guess (e.g., 1) and doubling it when the following condition is met based on newly observed expert labels: $\betag_1(\epsilon, \delta, \aCt, t \mid 2\hat{B}_g) < \ell_t(\hat{g}^{\mathrm{MLE}}_{t \mid 2\hat{B}_g}) - \ell_t(\hat{g}^{\mathrm{MLE}}_{t \mid \hat{B}_g})$, where $\hat{B}_g$ is our current guess. Intuitively, if the new likelihood $\ell_t(\hat{g}^{\mathrm{MLE}}_{t \mid 2\hat{B}_g})$ is significantly larger, then $2\hat{B}_g$ is more likely a valid bound. We iterate this estimation online during optimisation and in pre-training with the initial dataset (see details in Appendix~\ref{sec:est_norm_bound}).

\subsection{Algorithm and Theoretical Guarantee} 
\textbf{Algorithm.} Our algorithm in Alg.~\ref{alg:HAIBO} generates two candidates: the vanilla LCB $x^u_t$ and the expert-augmented LCB $x^c_t$.\rev{~(See App.~\ref{app:ext_acq} on extension to other acquisition functions.)} Always selecting the vanilla LCB guarantees no-harm but misses the chance to accelerate convergence using the expert's belief. Intuitively, this can be seen as a bandit problem regarding which arm to select.
Line~\ref{alg_line:no-elicitation guarantee} corresponds to the \emph{handover guarantee}, stating that our algorithm stops asking the expert once our model $g$ becomes more confident than the predefined $g_\text{thr}$. Line~\ref{alg_line:no-harm guarantee} outlines the conditions for achieving the \emph{no-harm guarantee} by assessing the reliability of the expert-augmented candidate $x^c_t$. The first condition ensures $x^c_t$ is at least possibly better than the worst-case estimation of the optimal value. The second condition acts as active learning of human belief, exploring uncertain points to avoid inaccurate yet confident expert beliefs. The hyperparameter $\eta\geq1$ represents the initial level of trust in the expert. A larger $\eta$ indicates greater priority in exploring expert-preferred regions.

\begin{figure}[t]
\begin{algorithm}[H]
\caption{\textbf{CO}llaborative \textbf{B}ayesian \textbf{O}ptimization with \textbf{L}abelling Experts (\textbf{COBOL}).}
\label{alg:HAIBO}
\begin{algorithmic}[1]
\normalsize
\State \rev{\textbf{Input and Initialization}: function space ball $\mathcal{B}_g$, trust weight $\eta$, and uncertainty threshold $\gthr$. }
\State Set $\mathcal{B}^1_g=\mathcal{B}_g$\rev{, $\mathcal{Q}^f_0=\emptyset$, and $\mathcal{Q}^g_0=\emptyset$}. 
\For{$t\in[T]$} 

\State Solve Prob.~\eqref{eq:primal_dual} via Prob.~\eqref{eqn:reform_inner_prob_to_fin} to generate $x^c_t$. \Comment{Expert-augmented LCB}
\label{alg_line:colbo_loop_start}
\State Solve the unconstrained problem, 
$x^u_t\in\arg\min_{x\in\mathcal{X}}\underline{f}_t(x).$
\Comment{Vanilla LCB}
\label{alg_line:uncon_opt_lcb}

\If{$\underline{f}_t(x_t^c)\leq\min_{x\in\mathcal{X}}\bar{f}_t(x)$ \textbf{and} $\sigma_{\ft}(x^u_t) \leq \eta \sigma_{\ft}(x^c_t)$ \label{alg_line:constr_cond} } \Comment{No-harm guarantee}
\label{alg_line:no-harm guarantee}
  \State Set $x_t=x_t^c$.
 \If{$\bar{g}_t(x_t)-\underline{g}_t(x_t)>\gthr$} \Comment{Handover guarantee}
\label{alg_line:no-elicitation guarantee}
 \State  Query the expert's label to get the feedback $\mathbf{1}_t$.
 \State Update $\Ct=\Ctm\cup\{t\}$ and the posterior confidence set $\mathcal{B}^{t+1}_g$. 
 \If{$\one_t=1$}
 \State Set $\Gt=\Gtm$, and continue the loop at line~\ref{alg_line:colbo_loop_start}.
 \EndIf
\EndIf  
\Else
\State Set $\Ct=\Ctm$, and $x_t=x_t^u$.
\EndIf

 
\State Evaluate the black-box function at the point $x_t$, and set $\Gt=\Gtm\cup\{t\}$.  
 \State Update the posterior mean/variance of the objective $f$.
\EndFor
\end{algorithmic}
\end{algorithm}
\vspace{-2em}
\end{figure}

\textbf{Theoretical guarantee.} For Alg.~\ref{alg:HAIBO}, we mainly care about two metrics: cumulative regret $R_{\GT}\defeq\sum_{t\in\GT}(f(x_t)-f(x^*))$ and cumulative queries $\CTc\defeq|\CT|$. $R_{\GT}$ captures the cumulative regret over the query points to the black-box function. $\CTc$ captures the number of queries to the expert. Since intuitively each query to the expert causes inconvenience, ideally, the frequency of query to an expert should be low (e.g., $\CTc$ grows sublinearly in $T$).

\begin{theorem}
\label{thm:RQ_bound}
Under Assumptions~\ref{assump:label_model} to \ref{assump:obj_obs}, with probability at least $1-\delta$, Alg~\ref{alg:HAIBO} satisfies, 
\label{eqn:cumu_reg_bound}
\noindent
\bse
\begin{center}
\vspace{-1em}
\begin{tabular}{p{5cm}p{7.5cm}}
  \begin{equation}
  R_{\GT}\leq\mathcal{O}\left(\gamma_{\aGT}^f\sqrt{\aGT}\right),
  \label{eq:no-harm}
  \end{equation} &
  \begin{equation}
  \CTc\leq \mathcal{O}\left(\left(\gamma^{g}_{T}\right)^2{\log\frac{T\mathcal{N}(\mathcal{B}_g, \nicefrac{1}{T},\|\cdot\|_\infty)}{\delta}}\right).
    \label{eq:elic_eff}
  \end{equation}
\end{tabular}
\end{center}
\ese
\end{theorem}
See Appendix~\ref{app:rq_bound_proof} for the proof of Thm.~\ref{thm:RQ_bound}. Intuitively, Eq.~\eqref{eq:no-harm} shows the \textbf{no-harm guarantee}, since it provides a cumulative regret bound independent of the latent \rev{function} $g$. Eq.~\eqref{eq:elic_eff} shows the \textbf{handover guarantee}, since the bound on cumulative queries to the expert is sublinear for commonly-used kernel functions~(See Table~\ref{tab:kern_spec_bounds}). This means that the frequency of querying the expert asymptotically converges to zero. Since $\CT\cup\GT=[T]$, $\aGT$ grows linearly in $T$. \rev{
It should be noted that there is a trade-off in $\eta$ selection. A larger $\eta$ can accelerate convergence when feedback is informative, but it may also cause the worse convergence rate for adversarial feedback (see Appendix~\ref{app:rq_bound_proof}, which includes an additional constant factor of $\nicefrac{(2 + \eta)}{4}$ compared to the original UCB). In practice, setting $\eta=3$ is sufficiently effective (see Figure~\ref{fig:robust}). 
}

By plugging in the maximum information gain bounds~\cite{srinivas2012information,vakili2021information} and covering number bounds~\cite{wu2017lecture,xu2024lower,bull2011convergence,zhou2002covering}, we apply Thm.~\ref{thm:RQ_bound} to derive the kernel-specific bounds in Table~\ref{tab:kern_spec_bounds}. \rev{In practice, kernel choice and scalability to high dimensions are common challenges for BO. Existing generic techniques, such as decomposed kernels \cite{kandasamy2015high}, can be applied in our algorithm to choose kernel functions and achieve scalability in high-dimensional spaces.}
\begin{table}[t]
    \centering
   \caption{Kernel-specific bounds where $\nu$ is the smoothness parameter of the Mat\'ern kernel that is assumed to satisfy $\nu>\frac{d}{4}(3+d+\sqrt{d^2+14d+17})=\Theta(d^2)$.}
    \label{tab:kern_spec_bounds}
    {
    \begin{tabular}{lccc}
    \toprule
         Metric & Linear &  Squared Exponential & Mat\'ern  \\ 
    \midrule
    $R_{\GT}$ & $\mathcal{O}\left(\sqrt{\aGT}\log \aGT\right)$ & $\mathcal{O}\left(\sqrt{\aGT}(\log \aGT)^{d+1} \right)$ & $\mathcal{O}\left(\aGT^{\frac{2\nu+3d}{4\nu+2d}}\log^{\frac{2\nu}{2\nu+d}}(\aGT)\right)$ \\ 
        $\CTc$  & $\mathcal{O}\left((\log T)^3\right)$& $\mathcal{O}\left((\log T)^{3(d+1)}\right)$ & $\mathcal{O}(T^{\frac{2d(d+1)}{2\nu+d(d+1)}}T^{\frac{d}{\nu}}(\log T)^3)$ \\
    \bottomrule
    \end{tabular}
    }
    \vspace{-1em}
\end{table}



\subsection{Related Works}
\textbf{Human-AI Collaborative BO.} 
There are two primary approaches: the first approach assumes that human experts can express their beliefs through \emph{quantitative} labels, such as well-defined distributions \cite{ramachandran2020incorporating, li2020incorporating, souza2021bayesian, hvarfner2022pi, cisse2023hypbo, hvarfner2024a} or pinpoint querying locations \cite{av2022human, gupta2023bo, khoshvishkaie2023cooperative, av2024enhanced, rodemann2024explaining}. While this strong assumption is valid in specific cases, such as physics simulations \cite{gupta2023bo}, many experimental tasks---such as chemistry, which lacks the consensus on numerical representations of, e.g. molecules---require more relaxed assumptions \cite{cisse2023hypbo, jordan2019artificial}. The \textit{qualitative} approach, on the other hand, involves human experts providing pairwise comparisons \cite{adachi2023looping} or binary recommendations (ours). The algorithm trains a surrogate model from experts' labels, thereby expanding applicable scenarios. Ours is the \emph{first-of-its-kind} principled method with both no-harm and handover guarantee on a continuous domain.

\textbf{Related BO tasks.} Eliciting human preference from labels has been explored in preferential BO \cite{brochu2007active, gonzalez2017preferential, mikkola2020projective, takeno2023towards, astudillo2023qeubo, xu2024principled}. However, this approach treats human preference as the main objective of BO, whereas our work uses experts' belief as an additional information source. Constrained BO \cite{gardner2014bayesian,gelbart2014bayesian,sui2015safe,sui2018stagewise,zhou2022kernelized,xu2023constrained,nguyen2024optimistic,jetton2023constraining, wang2024constrained, pmlr-v238-losalka24a} is another line of research that investigates BO under unknown constraints, placing another surrogate model on the constraint inferred from queried labels. However, our approach does not treat human belief as a constraint that must be satisfied or a reward to maximise, given that expert knowledge can sometimes be unreliable (see details in Appendix~\ref{app:related}).

\section{Experiments}\label{sec:exp}
We benchmarked the performance of the proposed algorithm against existing baselines in a collaborative setting with human experts. We employed an ARD RBF kernel for both $f$ and $g$. In each iteration of the optimisation loop, the inputs were rescaled to the unit cube $[0,1]^d$, and the outputs were standardised to have zero mean and unit variance. The initial datasets consisted of three random data points sampled uniformly from within the domain, and in each iteration, one data point was queried. Additionally, we collected initial expert labels by asking an expert to label `accept' (= 0) or `reject' (= 1) for 10 uniformly random points. All experiments were repeated ten times with different initial datasets and random seeds. 
We tuned hyperparameters online at each iteration. The GP hyperparameters were tuned by maximising the marginal likelihood on observed datasets using a multi-start L-BFGS-B method \cite{liu1989limited} (the default BoTorch optimiser \cite{balandat2020botorch}). The key hyperparameters of the confidence set, $B_\gl$ and $\alpha_1$, were optimised via the online method in Appendix~\ref{sec:est_norm_bound}. Other hyperparameters were set as $\eta = 3$, $\lambda_0 = 1$, and $g_\text{thr} = 0.1$ by default throughout the experiments, with their sensitivity discussed later (see also Appendix~\ref{app:hypers}). The constrained optimisation in Prob.~\eqref{eqn:reform_inner_prob_to_fin} was solved using the interior-point nonlinear optimiser IPOPT \cite{wachter2006implementation}\rev{, which is highly scalable for solving the primal problem,} via the symbolic interface CasADi \cite{Andersson2019}. The unconstrained optimisation (line~\ref{alg_line:uncon_opt_lcb}) was solved using the default BoTorch optimiser \cite{balandat2020botorch}. More details for reproducing results are available on GitHub.\footnote{\url{https://github.com/ma921/COBOL/}} The models were implemented in GPyTorch \cite{gardner2018gpytorch}. All experiments were conducted on a laptop PC.\footnote{MacBook Pro 2019, 2.4 GHz 8-Core Intel Core i9, 64 GB 2667 MHz DDR4} Computational time is discussed in Appendix~\ref{appendix:experiments}. In addition to cumulative regret and queries, we also consider simple regret defined as $\text{SR}_{t} := \min_{\tau\in\Gt}(f(x_\tau)-f(x^\star))$.

\textbf{Robustness and sensitivity.} 
\begin{figure}
  \centering
  \includegraphics[width=0.8\hsize]{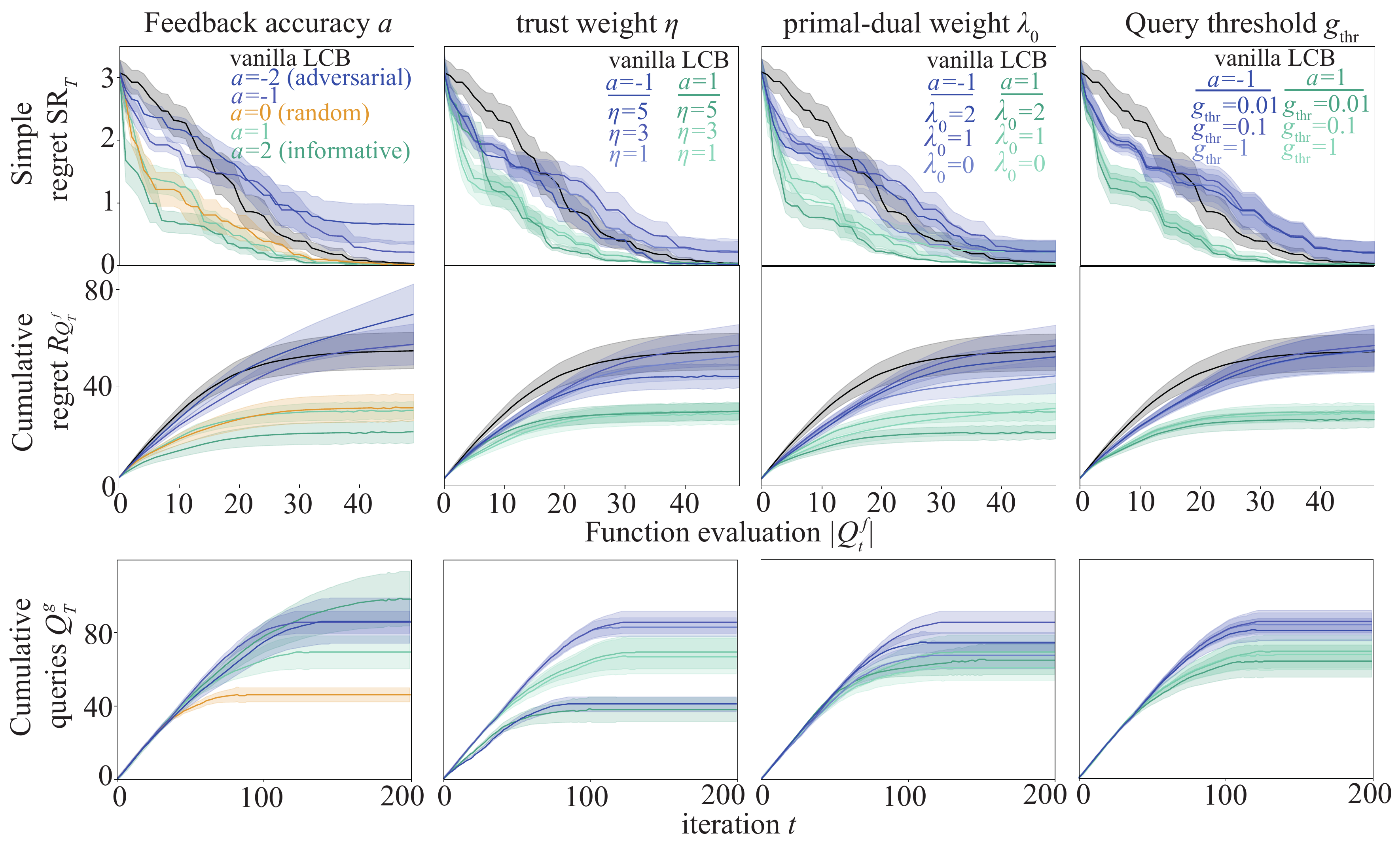}
  \caption{Robustness and sensitivity analysis using the Ackley function. Lines and shaded areas denote mean $\pm$ 1 standard error. The no-harm guarantee ensures the convergence rate is on par with vanilla LCB even in adversarial cases. Handover guarantee ensures that $\Ct$ plateau, allowing optimisation without expert intervention once sufficient information has been elicited.}
  \label{fig:robust}
  \vspace{-1em}
\end{figure}
First, we tested the robustness of our algorithm to the accuracy of the expert's labels using the 4-dimensional Ackley function \cite{ackley2012connectionist}. We modelled the synthetic agent response  according to Example~\ref{example:g_func}. In particular, we examine the impact of feedback accuracy, denoted as $a$.
Fig.~\ref{fig:robust} illustrates the robustness of our algorithm. When labels are informative ($a = 1, 2$), the convergence rate for both simple and cumulative regrets is accelerated in accordance with the accuracy. Even if the feedback is completely random ($a=0$) or adversarial ($a=-1,-2$), the no-harm guarantee ensures that the algorithm converges at a rate on par with vanilla LCB by adjusting the level of trust to be lower over iterations. \rev{Refer to Appendix~\ref{app:no-harm} for additional confirmation of the no-harm guarantee based on more extensive experimental results.} Handover guarantee ensures that our algorithm stops seeking label feedback once sufficient information has been elicited, as indicated by the plateau in the cumulative queries $\CTc$. We also tested the sensitivity to the optimisation parameters $\eta, \lambda_0$, and $g_\text{thr}$. The change in convergence at those parameters were varied mostly within the standard error, indicating that our algorithm is insensitive to these hyperparameters and that feedback accuracy is more dominant. For the primal-dual weight, $\lambda_0 = 0$ corresponds to constrained optimisation without a primal-dual mixing objective, which performs worse than mixing cases ($\lambda_0 = 1, 2$), demonstrating the efficacy of incorporating the primal-dual mixing objective.

\textbf{Synthetic dataset.} 
\begin{figure}
  \centering
  \includegraphics[width=1\hsize]{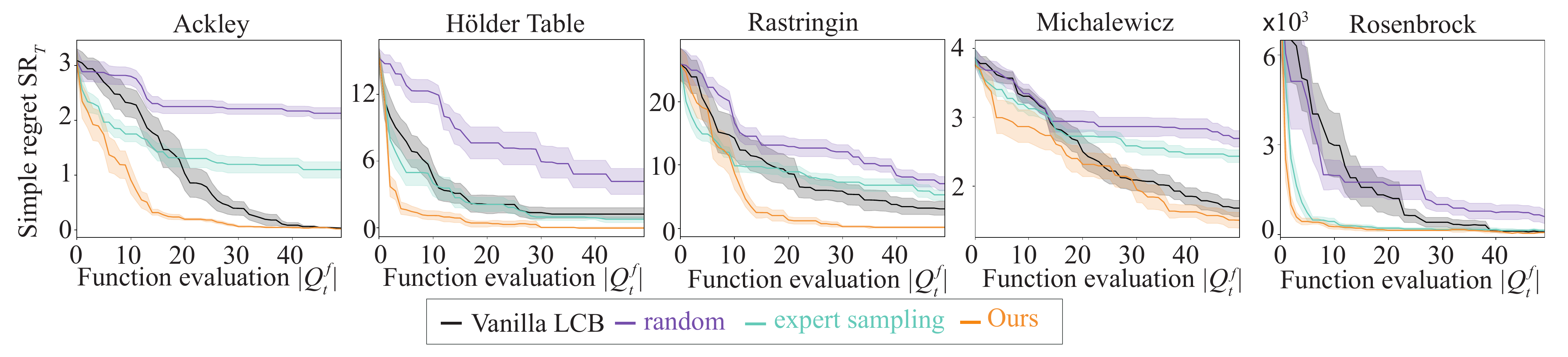}
  \caption{Ablation study on five common synthetic functions with synthetic expert labels ($a = 1$).}
  \label{fig:synth}
\end{figure}
We compared our algorithm against five common synthetic functions \cite{simulationlib} (see details in Appendix~\ref{sec:synthetic}), using simple baselines for an ablation study: random sampling, vanilla LCB (unconstrained optimisation), and expert sampling. Expert sampling involves direct sampling from the expert belief distribution $p_{x\succ_g0}$. We employ rejection sampling by generating a uniform random sample over the domain and then accepting it with the probability $1-p_{x\succ_g0}$. We fixed the feedback accuracy at $a = 1$\rev{~(as in Example~\ref{example:g_func}.)}. The efficacy of expert labels is roughly estimated by how much faster expert sampling converges compared to random sampling. In all synthetic experiments, our algorithm outperformed the baselines. While expert sampling is at least more effective than random sampling, it is not always better than vanilla LCB. For functions with a very sharp global optimum, such as Rosenbrock \cite{rosenbrock1960automatic}, $p_{x\succ_g0}$ nearly pinpoints the global minimum. Still, our algorithm performs slightly better than expert sampling. See Appendix~\ref{app:synthetic_time} for computation time and query frequency.
\rev{
The overhead of our algorithm is comparable to that of other baselines.
}

\begin{figure}
  \centering
  \includegraphics[width=1\hsize]{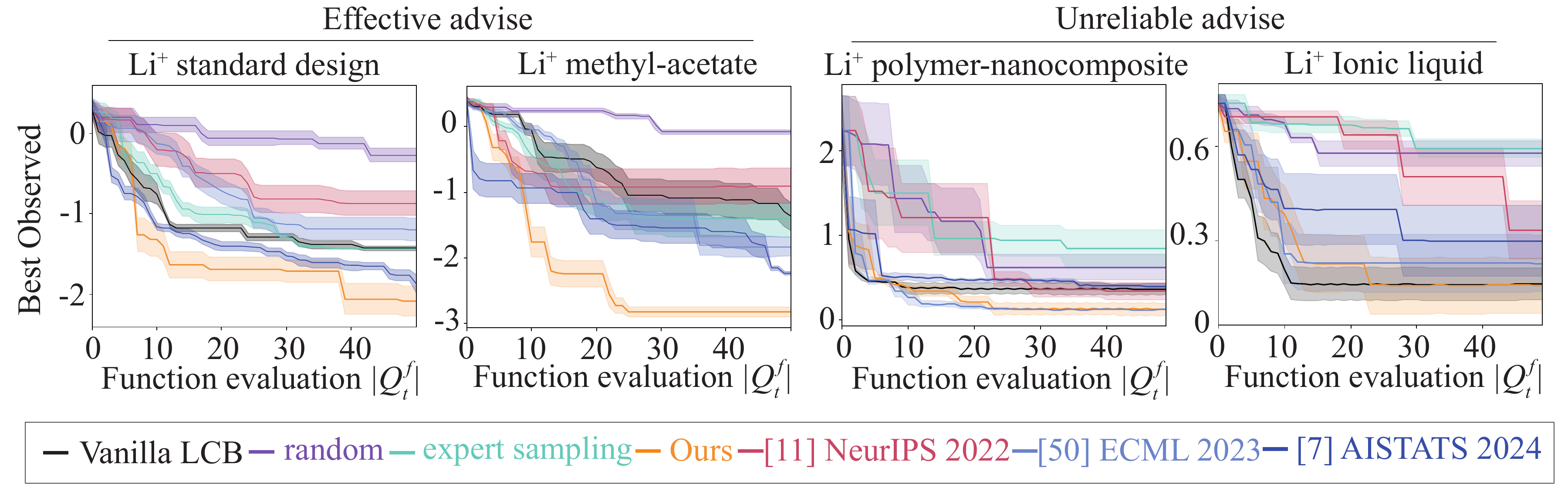}
  \caption{Real-world experiments with four human experts of lithium-ion batteries.}
  \label{fig:human}
  \vspace{-1em}
\end{figure}
\textbf{Real-world experiments with human experts.} We conducted real-world experiments in collaboration with four human experts who possess post-doctoral level knowledge on lithium-ion batteries. 
\rev{In this experiment, human labelling costs vary among experts but typically range from a few seconds to several minutes. In the real-world development of lithium-ion batteries, creating and testing a prototype cell requires at least a week, making the labelling cost negligible by comparison.}

Lithium-ion batteries are crucial for realising a green society, a rapidly growing field where knowledge is continuously updated at an unprecedented rate. This field typically suffers from data scarcity \cite{jordan2019artificial} due to the ongoing development of new materials synthesised by chemists. Consequently, transfer learning approaches, e.g., \cite{swersky2013multi, wistuba2020few, feurer2018scalable, chen2022towards}, are not effective in this setting.
We prepared four cases for the experiments: the first is a standard task where we optimise the standard electrolyte composition \cite{dave2022autonomous, gering2017prediction}, and the second involves a slight modification of the first setup by changing one solvent material \cite{logan2018study}.\footnote{This slight change makes optimal design challenging enough \cite{gering2017prediction}. See Appendix~\ref{app:reason} for details.} We expect the experts to have informative knowledge on these two tasks. The remaining two cases involve emerging new categories of materials: one is a polymer-nanocomposite electrolyte \cite{zhang2016flexible}, and the other is an ionic liquid \cite{rosol2009solubility}. We anticipate that the experts' knowledge on these new materials will not be as effective as in the first two tasks (see more details in Appendix~\ref{sec:human}).
Given the scarcity of real experts, we conducted a pre-experimental step to elicit their knowledge for a fair baseline comparison. We asked them to label 50 random points uniformly from the domain, for all experiments before seeing the results. Then we fit the confidence set model to these results and used $\hat{g}_t^\text{MLE}$ as the \emph{estimated} human response. Additionally, we asked the participants to manually select the next query point without any assistance from BO, which we refer to as `expert sampling' in the baseline. We also compared against state-of-the-art algorithms \cite{av2022human, khoshvishkaie2023cooperative, adachi2023looping}. These methods have predefined levels of trust, roughly ranked from strong to weak: \cite{av2022human} $\rightarrow$ \cite{adachi2023looping} $\rightarrow$ \cite{khoshvishkaie2023cooperative}. Ours can adjust the level of trust based on data, so we expect it to perform well in both effective and ineffective cases.

Fig.~\ref{fig:human} summarises the results. For the first two tasks, our algorithm outperformed all baselines. Particularly in the second task, human sampling was better than vanilla LCB, indicating that we should trust their advice aggressively. Our algorithm can adapt to trust them over time, resulting in significantly accelerated convergence. On the other hand, expert sampling for the new materials tasks was, although unintentionally, worse than random, thereby discouraging trust. While trustful algorithms \cite{av2022human, adachi2023looping} struggled to converge, the distrustful algorithm \cite{khoshvishkaie2023cooperative} was able to converge on par with vanilla LCB. Our no-harm guarantee worked in this situation, gradually equating to LCB, and showed identical performance to the distrustful algorithm \cite{khoshvishkaie2023cooperative}. 
See also Appendix~\ref{app:real_time} for the complete experimental results on the number of queries and computation time.

\section{Discussion}
\rev{
\textbf{Feedback form.} Other forms of feedback, such as pairwise comparisons \cite{adachi2023looping} or preferential rankings \cite{av2024enhanced}, can be incorporated into our algorithm with slight modifications. However, we empirically found that the binary labelling approach performs best (see Fig.~\ref{fig:human}), and therefore, we recommend using binary feedback as the primary choice. For those interested in using alternative feedback forms, detailed instructions on how to adapt them to our algorithm are provided in Appendix~\ref{app:feedback}.

\textbf{Time-varying human knowledge.} 
We assume that expert knowledge is stationary, although it can be time-varying, e.g., experts' knowledge often evolves as more data is gathered. A simple extension to accommodate this is the use of windowing, where past queried data is forgotten. This can be easily implemented in our algorithm by removing old data beyond a predefined iteration window. However, our initial trials did not show significant performance gains from this approach, so it was not included in the main text. We suggest a dynamic model as a potential future direction, which is discussed in Appendix~\ref{app:dynamic} with additional experimental results. Similarly, we kept the trust weight $\eta$ fixed throughout the optimization process. Since human knowledge can improve over time, an adaptive $\eta$ could be employed to enhance both convergence and robustness. Nevertheless, our no-harm guarantee remains valid even without this adaptation. Further details are provided in Appendix~\ref{app:adapt_eta}.


\textbf{Acceleration vs. Robustness.} 
One might seek to derive a theoretical guarantee on the acceleration of convergence when the feedback is helpful. However, we want to emphasize that theoretically guaranteeing both acceleration and robustness may be incompatible. From a theoretical perspective, they are in a trade-off relationship \cite{tsipras2018robustness}. This can be intuitively explained by the no-free-lunch theorem \cite{wolpert1997no}: if algorithm A outperforms B, it does so by exploiting `biased' information. The `bias' inherent in the acceleration is contradictory to robustness. Our setting is unbiased, meaning we do not have prior knowledge of helpful or adversarial human expert. Therefore, we must make a design choice between prioritizing robustness or acceleration as a theoretical contribution, depending on whether we assume that expert input can be adversarial (weak bias) or that it will always be helpful (strong bias). Indeed, there are lower bound results for the average-case regret of Bayesian optimization in the literature (e.g., see \cite{scarlett2017lower}). GP-UCB is already nearly rate-optimal in achieving this lower bound. This means theoretical acceleration is obtained in the price of worse robustness. In Appendix~\ref{app:improvement}, we present a slightly modified version, Algorithm~\ref{alg:cobohl}, which offers an improvement guarantee based on strong bias. Our Algorithm~\ref{alg:HAIBO} can be seen as a relaxed version of this algorithm (soft constraint), which helps explain the empirical success in accelerating convergence.
}

\section{Conclusion}
Our algorithm, with its data-driven adjustment of the level of trust, successfully accelerated convergence from effective advice while ensuring a no-harm guarantee from unreliable inputs. The handover guarantee also ensures that the BO can automate the optimisation process without assistance from human experts at a later stage. These features are particularly valuable for scientific applications, where researchers often face trial and error, making it challenging to determine the effectiveness of their prior knowledge before starting experiments. Our flexible and robust framework is also expected to be effective in collaboration with large language models (LLMs), which demonstrate remarkable sample-efficient performance by exploiting encoded priors \cite{liu2024large, sanh2022multitask, ouyang2022trainig}, and can be regarded as `expert knowledge'. Our safeguard features would be particularly effective for shared challenges, such as difficulty in eliciting knowledge \cite{jiang2020can, bommasani2021opportunities} and varying accuracy of advice due to hallucinations \cite{shin-etal-2020-autoprompt, wei2022chain, srivastava2022beyond}.
Although ours is the \emph{first-of-its-kind} algorithm with a general theoretical guarantee in the expert-collaborative setting, it is still based on the GP-UCB algorithm~\footnote{\rev{Maximization formulation is adopted in GP-UCB paper~\cite{srinivas2012information}, while we consider minimization. So LCB in our paper essentially corresponds to UCB in GP-UCB algorithm.}} and shares its limitations (e.g., high dimensionality). \rev{One future direction is combining our approach with the high-dimensional BO methods~\cite{wang2016bayesian,kirschner2019adaptive}.} Additionally, our current setting does not consider the batch setting, yet one can easily extend with existing approaches, e.g.~\cite{adachi2022fast, adachi2023bayesian, adachi2023sober, adachi2024adaptive}. Multiple expert scenario is also a promising future extension. While a simple expert aggregation approach (e.g., majority vote, adding multiple \rev{\gcost s} $g$) could work without modifications to the current algorithm, more advanced methods, such as choice functions \cite{benavoli2023learning}, present promising directions for future work. 
Our method can positively influence human experts by empirically demonstrating the value of their expertise, even amidst concerns about job security in the AI era \cite{bakhshi2017future}. On the negative side, more powerful LLMs may eventually replace the expert role in our algorithm in areas where data is sufficiently shared on websites or in papers, such as hyperparameter tuning \cite{liu2024large}.
\begin{ack}
We would like to thank Ondrej Bajgar, Juliusz Ziomek, and anonymous reviewers for their helpful comments about improving the paper. 
Wenjie Xu and Colin N. Jones were supported by the Swiss National Science Foundation under NCCR Automation, grant agreement 51NF40\_180545.
Masaki Adachi was supported by the Clarendon Fund, the Oxford Kobe Scholarship, the Watanabe Foundation, and Toyota Motor Corporation.
\end{ack}

\bibliographystyle{plain}
\sloppy
\bibliography{references}

\begin{thebibliography}{100}

\bibitem{ackley2012connectionist}
David Ackley.
\newblock {\em A connectionist machine for genetic hillclimbing}, volume~28.
\newblock Springer science \& business media, 1987.

\bibitem{adachi2021high}
Masaki Adachi.
\newblock High-dimensional discrete {B}ayesian optimization with self-supervised representation learning for data-efficient materials exploration.
\newblock In {\em NeurIPS 2021 AI for Science Workshop}, 2021.

\bibitem{adachi2023sober}
Masaki Adachi, Satoshi Hayakawa, Martin J{\o}rgensen, Saad Hamid, Harald Oberhauser, and Michael~A Osborne.
\newblock A quadrature approach for general-purpose batch {B}ayesian optimization via probabilistic lifting.
\newblock {\em arXiv preprint arXiv:2404.12219}, 2024.

\bibitem{adachi2022fast}
Masaki Adachi, Satoshi Hayakawa, Martin J{\o}rgensen, Harald Oberhauser, and Michael~A Osborne.
\newblock Fast {B}ayesian inference with batch {B}ayesian quadrature via kernel recombination.
\newblock {\em Advances in Neural Information Processing Systems (NeurIPS)}, 35:16533--16547, 2022.

\bibitem{adachi2024adaptive}
Masaki Adachi, Satoshi Hayakawa, Martin J{\o}rgensen, Xingchen Wan, Vu~Nguyen, Harald Oberhauser, and Michael~A Osborne.
\newblock Adaptive batch sizes for active learning: A probabilistic numerics approach.
\newblock In {\em International Conference on Artificial Intelligence and Statistics (AISTATS)}, pages 496--504. PMLR, 2024.

\bibitem{adachi2023bayesian}
Masaki Adachi, Yannick Kuhn, Birger Horstmann, Arnulf Latz, Michael~A Osborne, and David~A Howey.
\newblock Bayesian model selection of lithium-ion battery models via {B}ayesian quadrature.
\newblock {\em IFAC-PapersOnLine}, 56(2):10521--10526, 2023.

\bibitem{adachi2023looping}
Masaki Adachi, Brady Planden, David~A Howey, Michael~A Osborne, Sebastian Orbell, Natalia Ares, Krikamol Muandet, and Siu~Lun Chau.
\newblock Looping in the human: Collaborative and explainable {B}ayesian optimization.
\newblock In {\em International Conference on Artificial Intelligence and Statistics (AISTATS)}, 2024.

\bibitem{Andersson2019}
Joel A~E Andersson, Joris Gillis, Greg Horn, James~B Rawlings, and Moritz Diehl.
\newblock {CasADi} -- {A} software framework for nonlinear optimization and optimal control.
\newblock {\em Mathematical Programming Computation}, 11(1):1--36, 2019.

\bibitem{astudillo2023qeubo}
Raul Astudillo, Zhiyuan~Jerry Lin, Eytan Bakshy, and Peter Frazier.
\newblock q{EUBO}: {A} decision-theoretic acquisition function for preferential {B}ayesian optimization.
\newblock In {\em International Conference on Artificial Intelligence and Statistics (AISTATS)}, pages 1093--1114. PMLR, 2023.

\bibitem{augustin2014introduction}
Thomas Augustin, Frank~PA Coolen, Gert De~Cooman, and Matthias~CM Troffaes.
\newblock {\em Introduction to imprecise probabilities}, volume 591.
\newblock John Wiley \& Sons, 2014.

\bibitem{av2022human}
Arun~Kumar AV, Santu Rana, Alistair Shilton, and Svetha Venkatesh.
\newblock Human-{AI} collaborative {B}ayesian {O}ptimisation.
\newblock {\em Advances in Neural Information Processing Systems (NeurIPS)}, 35:16233--16245, 2022.

\bibitem{av2024enhanced}
Arun~Kumar AV, Alistair Shilton, Sunil Gupta, Santu Rana, Stewart Greenhill, and Svetha Venkatesh.
\newblock Enhanced {B}ayesian optimization via preferential modeling of abstract properties.
\newblock {\em arXiv preprint arXiv:2402.17343}, 2024.

\bibitem{bakhshi2017future}
Hasan Bakhshi, Jonathan Downing, Michael Osborne, and Philippe Schneider.
\newblock {\em The future of skills: Employment in 2030}.
\newblock Pearson, 2017.

\bibitem{balandat2020botorch}
Maximilian Balandat, Brian Karrer, Daniel Jiang, Samuel Daulton, Ben Letham, Andrew~G Wilson, and Eytan Bakshy.
\newblock Bo{T}orch: a framework for efficient {M}onte-{C}arlo {B}ayesian optimization.
\newblock {\em Advances in Neural Information Processing Systems (NeurIPS)}, 33:21524--21538, 2020.

\bibitem{benavoli2023learning}
Alessio Benavoli, Dario Azzimonti, and Dario Piga.
\newblock Learning choice functions with {G}aussian processes.
\newblock In {\em Uncertainty in Artificial Intelligence (UAI)}, pages 141--151. PMLR, 2023.

\bibitem{bommasani2021opportunities}
Rishi Bommasani, Drew~A Hudson, Ehsan Adeli, Russ Altman, Simran Arora, Sydney von Arx, Michael~S Bernstein, Jeannette Bohg, Antoine Bosselut, Emma Brunskill, et~al.
\newblock On the opportunities and risks of foundation models.
\newblock {\em arXiv preprint arXiv:2108.07258}, 2021.

\bibitem{bradley1952rank}
Ralph~Allan Bradley and Milton~E Terry.
\newblock Rank analysis of incomplete block designs: I. the method of paired comparisons.
\newblock {\em Biometrika}, 39(3/4):324--345, 1952.

\bibitem{bull2011convergence}
Adam~D Bull.
\newblock Convergence rates of efficient global optimization algorithms.
\newblock {\em Journal of Machine Learning Research (JMLR)}, 12(10), 2011.

\bibitem{casteel1972specific}
Jerry~F Casteel and Edward~S Amis.
\newblock Specific conductance of concentrated solutions of magnesium salts in water-ethanol system.
\newblock {\em Journal of Chemical and Engineering Data}, 17(1):55--59, 1972.

\bibitem{chau2022learning}
Siu~Lun Chau, Javier Gonzalez, and Dino Sejdinovic.
\newblock Learning inconsistent preferences with {G}aussian processes.
\newblock In {\em International Conference on Artificial Intelligence and Statistics (AISTATS)}, pages 2266--2281. PMLR, 2022.

\bibitem{chen2022towards}
Yutian Chen, Xingyou Song, Chansoo Lee, Zi~Wang, Richard Zhang, David Dohan, Kazuya Kawakami, Greg Kochanski, Arnaud Doucet, Marc'aurelio Ranzato, et~al.
\newblock Towards learning universal hyperparameter optimizers with transformers.
\newblock {\em Advances in Neural Information Processing Systems (NeurIPS)}, 35:32053--32068, 2022.

\bibitem{chowdhury2017kernelized}
Sayak~Ray Chowdhury and Aditya Gopalan.
\newblock On kernelized multi-armed bandits.
\newblock In {\em International Conference on Machine Learning (ICML)}, pages 844--853. PMLR, 2017.

\bibitem{chu2005preference}
Wei Chu and Zoubin Ghahramani.
\newblock Preference learning with {G}aussian processes.
\newblock In {\em Proceedings of the 22nd international conference on Machine learning}, pages 137--144, 2005.

\bibitem{cisse2023hypbo}
Abdoulatif Cisse, Xenophon Evangelopoulos, Sam Carruthers, Vladimir~V Gusev, and Andrew~I Cooper.
\newblock {HypBO}: Expert-guided chemist-in-the-loop {B}ayesian search for new materials.
\newblock {\em arXiv preprint arXiv:2308.11787}, 2023.

\bibitem{dai2018coupled}
Bo~Dai, Hanjun Dai, Niao He, Weiyang Liu, Zhen Liu, Jianshu Chen, Lin Xiao, and Le~Song.
\newblock Coupled variational {B}ayes via optimization embedding.
\newblock {\em Advances in Neural Information Processing Systems (NeurIPS)}, 31, 2018.

\bibitem{dave2022autonomous}
Adarsh Dave, Jared Mitchell, Sven Burke, Hongyi Lin, Jay Whitacre, and Venkatasubramanian Viswanathan.
\newblock Autonomous optimization of non-aqueous {L}i-ion battery electrolytes via robotic experimentation and machine learning coupling.
\newblock {\em Nature communications}, 13(1):5454, 2022.

\bibitem{emmenegger2023likelihood}
Nicolas Emmenegger, Mojmir Mutny, and Andreas Krause.
\newblock Likelihood ratio confidence sets for sequential decision making.
\newblock In {\em Thirty-seventh Conference on Neural Information Processing Systems}, 2023.

\bibitem{brochu2007active}
Brochu Eric, Nando Freitas, and Abhijeet Ghosh.
\newblock Active preference learning with discrete choice data.
\newblock In {\em Advances in Neural Information Processing Systems (NeurIPS)}, volume~20, 2007.

\bibitem{feurer2015efficient}
Matthias Feurer, Aaron Klein, Katharina Eggensperger, Jost Springenberg, Manuel Blum, and Frank Hutter.
\newblock Efficient and robust automated machine learning.
\newblock {\em Advances in Neural Information Processing Systems (NeurIPS)}, 28, 2015.

\bibitem{feurer2018scalable}
Matthias Feurer, Benjamin Letham, and Eytan Bakshy.
\newblock Scalable meta-learning for {B}ayesian optimization using ranking-weighted {G}aussian process ensembles.
\newblock In {\em AutoML Workshop at ICML}, volume~7, page~5, 2018.

\bibitem{gardner2018gpytorch}
Jacob Gardner, Geoff Pleiss, Kilian~Q Weinberger, David Bindel, and Andrew~G Wilson.
\newblock {GPyTorch}: Blackbox matrix-matrix {G}aussian process inference with {GPU} acceleration.
\newblock In {\em Advances in Neural Information Processing Systems (NeurIPS)}, pages 7576--7586, 2018.

\bibitem{gardner2014bayesian}
Jacob~R Gardner, Matt~J Kusner, Zhixiang~Eddie Xu, Kilian~Q Weinberger, and John~P Cunningham.
\newblock Bayesian optimization with inequality constraints.
\newblock In {\em International Conference on Machine Learning~(ICML)}, volume 2014, pages 937--945, 2014.

\bibitem{garnett2023bayesian}
Roman Garnett.
\newblock {\em Bayesian optimization}.
\newblock Cambridge University Press, 2023.

\bibitem{garthwaite2005statistical}
Paul~H Garthwaite, Joseph~B Kadane, and Anthony O'Hagan.
\newblock Statistical methods for eliciting probability distributions.
\newblock {\em Journal of the American Statistical Association}, 100(470):680--701, 2005.

\bibitem{gering2017prediction}
Kevin~L Gering.
\newblock Prediction of electrolyte conductivity: results from a generalized molecular model based on ion solvation and a chemical physics framework.
\newblock {\em Electrochimica Acta}, 225:175--189, 2017.

\bibitem{gonzalez2017preferential}
Javier Gonz{\'a}lez, Zhenwen Dai, Andreas Damianou, and Neil~D Lawrence.
\newblock Preferential {B}ayesian optimization.
\newblock In {\em International Conference on Machine Learning (ICML)}, pages 1282--1291. PMLR, 2017.

\bibitem{guo2010gaussian}
Shengbo Guo, Scott Sanner, and Edwin~V Bonilla.
\newblock Gaussian process preference elicitation.
\newblock {\em Advances in Neural Information Processing Systems (NeurIPS)}, 23, 2010.

\bibitem{gupta2023bo}
Sunil Gupta, Alistair Shilton, Arun~Kumar AV, Shannon Ryan, Majid Abdolshah, Hung Le, Santu Rana, Julian Berk, Mahad Rashid, and Svetha Venkatesh.
\newblock {BO}-{M}use: A human expert and {AI} teaming framework for accelerated experimental design.
\newblock {\em arXiv preprint arXiv:2303.01684}, 2023.

\bibitem{hong2023optimization}
Kihyuk Hong, Yuhang Li, and Ambuj Tewari.
\newblock An optimization-based algorithm for non-stationary kernel bandits without prior knowledge.
\newblock In {\em International Conference on Artificial Intelligence and Statistics (AISTATS)}, pages 3048--3085. PMLR, 2023.

\bibitem{hullermeier2021aleatoric}
Eyke H{\"u}llermeier and Willem Waegeman.
\newblock Aleatoric and epistemic uncertainty in machine learning: An introduction to concepts and methods.
\newblock {\em Machine learning}, 110(3):457--506, 2021.

\bibitem{hvarfner2024a}
Carl Hvarfner, Frank Hutter, and Luigi Nardi.
\newblock A general framework for user-guided {B}ayesian optimization.
\newblock In {\em International Conference on Learning Representations (ICLR)}, 2024.

\bibitem{hvarfner2022pi}
Carl Hvarfner, Danny Stoll, Artur Souza, Marius Lindauer, Frank Hutter, and Luigi Nardi.
\newblock $\pi${BO}: Augmenting acquisition functions with user beliefs for {B}ayesian optimization.
\newblock In {\em International Conference on Learning Representations (ICLR)}, 2022.

\bibitem{jetton2023constraining}
Cole Jetton, Matthew Campbell, and Christopher Hoyle.
\newblock Constraining the feasible design space in {B}ayesian optimization with user feedback.
\newblock {\em Journal of Mechanical Design}, 146(4):041703, 2023.

\bibitem{jiang2020can}
Zhengbao Jiang, Frank~F Xu, Jun Araki, and Graham Neubig.
\newblock How can we know what language models know?
\newblock {\em Transactions of the Association for Computational Linguistics}, 8:423--438, 2020.

\bibitem{jordan2019artificial}
Michael~I Jordan.
\newblock Artificial intelligence—the revolution hasn’t happened yet.
\newblock {\em Harvard Data Science Review}, 1(1):1--9, 2019.

\bibitem{kahneman1979interpretation}
Daniel Kahneman and Amos Tversky.
\newblock {On the interpretation of intuitive probability: A reply to Jonathan Cohen}.
\newblock {\em Cognition}, 7(4):409--411, 1979.

\bibitem{kanarik2023human}
Keren~J Kanarik, Wojciech~T Osowiecki, Yu~Lu, Dipongkar Talukder, Niklas Roschewsky, Sae~Na Park, Mattan Kamon, David~M Fried, and Richard~A Gottscho.
\newblock Human--machine collaboration for improving semiconductor process development.
\newblock {\em Nature}, 616(7958):707--711, 2023.

\bibitem{kandasamy2015high}
Kirthevasan Kandasamy, Jeff Schneider, and Barnab{\'a}s P{\'o}czos.
\newblock High dimensional {B}ayesian optimisation and bandits via additive models.
\newblock In {\em International Conference on Machine Learning (ICML)}, pages 295--304. PMLR, 2015.

\bibitem{khoshvishkaie2023cooperative}
Ali Khoshvishkaie, Petrus Mikkola, Pierre-Alexandre Murena, and Samuel Kaski.
\newblock Cooperative {B}ayesian optimization for imperfect agents.
\newblock In {\em Joint European Conference on Machine Learning and Knowledge Discovery in Databases (ECML)}, pages 475--490. Springer, 2023.

\bibitem{kirschner2019adaptive}
Johannes Kirschner, Mojmir Mutny, Nicole Hiller, Rasmus Ischebeck, and Andreas Krause.
\newblock Adaptive and safe {B}ayesian optimization in high dimensions via one-dimensional subspaces.
\newblock In {\em International Conference on Machine Learning (ICML)}, pages 3429--3438. PMLR, 2019.

\bibitem{li2020incorporating}
Cheng Li, Sunil Gupta, Santu Rana, Vu~Nguyen, Antonio Robles-Kelly, and Svetha Venkatesh.
\newblock Incorporating expert prior knowledge into experimental design via posterior sampling.
\newblock {\em arXiv preprint arXiv:2002.11256}, 2020.

\bibitem{liu1989limited}
Dong~C Liu and Jorge Nocedal.
\newblock On the limited memory {BFGS} method for large scale optimization.
\newblock {\em Mathematical programming}, 45(1-3):503--528, 1989.

\bibitem{liu2023optimistic}
Qinghua Liu, Praneeth Netrapalli, Csaba Szepesvari, and Chi Jin.
\newblock Optimistic {MLE}: A generic model-based algorithm for partially observable sequential decision making.
\newblock In {\em Proceedings of the 55th Annual ACM Symposium on Theory of Computing}, pages 363--376, 2023.

\bibitem{liu2024large}
Tennison Liu, Nicol{\'a}s Astorga, Nabeel Seedat, and Mihaela van~der Schaar.
\newblock Large language models to enhance {B}ayesian optimization.
\newblock In {\em The Twelfth International Conference on Learning Representations}, 2024.

\bibitem{logan2018study}
ER~Logan, Erin~M Tonita, KL~Gering, Jing Li, Xiaowei Ma, LY~Beaulieu, and JR~Dahn.
\newblock A study of the physical properties of {L}i-ion battery electrolytes containing esters.
\newblock {\em Journal of The Electrochemical Society}, 165(2):A21, 2018.

\bibitem{pmlr-v238-losalka24a}
Arpan Losalka and Jonathan Scarlett.
\newblock No-regret algorithms for safe {B}ayesian optimization with monotonicity constraints.
\newblock In Sanjoy Dasgupta, Stephan Mandt, and Yingzhen Li, editors, {\em Proceedings of The 27th International Conference on Artificial Intelligence and Statistics}, volume 238 of {\em Proceedings of Machine Learning Research}, pages 3232--3240. PMLR, 02--04 May 2024.

\bibitem{mikkola2021prior}
Petrus Mikkola, Osvaldo~A Martin, Suyog Chandramouli, Marcelo Hartmann, Oriol~Abril Pla, Owen Thomas, Henri Pesonen, Jukka Corander, Aki Vehtari, Samuel Kaski, et~al.
\newblock Prior knowledge elicitation: The past, present, and future.
\newblock {\em arXiv preprint arXiv:2112.01380}, 2112, 2021.

\bibitem{mikkola2020projective}
Petrus Mikkola, Milica Todorovi{\'c}, Jari J{\"a}rvi, Patrick Rinke, and Samuel Kaski.
\newblock Projective preferential {B}ayesian optimization.
\newblock In {\em International Conference on Machine Learning (ICML)}, pages 6884--6892. PMLR, 2020.

\bibitem{mockus1978application}
Jonas Mockus, Vytautas Tiesis, and Antanas Zilinskas.
\newblock The application of {B}ayesian methods for seeking the extremum.
\newblock {\em Towards global optimization}, 2(117-129):2, 1978.

\bibitem{mohammadi2016analytic}
Hossein Mohammadi, Rodolphe~Le Riche, Nicolas Durrande, Eric Touboul, and Xavier Bay.
\newblock An analytic comparison of regularization methods for {G}aussian processes.
\newblock {\em arXiv preprint arXiv:1602.00853}, 2016.

\bibitem{nguyen2024optimistic}
Quoc~Phong Nguyen, Wan Theng~Ruth Chew, Le~Song, Bryan Kian~Hsiang Low, and Patrick Jaillet.
\newblock Optimistic {B}ayesian optimization with unknown constraints.
\newblock In {\em The Twelfth International Conference on Learning Representations}, 2024.

\bibitem{nickisch2008approximations}
Hannes Nickisch and Carl~Edward Rasmussen.
\newblock Approximations for binary {G}aussian process classification.
\newblock {\em Journal of Machine Learning Research (JMLR)}, 9(Oct):2035--2078, 2008.

\bibitem{nocedal1999numerical}
Jorge Nocedal and Stephen~J Wright.
\newblock {\em Numerical optimization}.
\newblock Springer, 1999.

\bibitem{osborne2009gaussian}
Michael~A Osborne, Roman Garnett, and Stephen~J Roberts.
\newblock Gaussian processes for global optimization.
\newblock In {\em International Conference on Learning and Intelligent Optimization (LION3)}, 2009.

\bibitem{ouyang2022trainig}
Long Ouyang, Jeffrey Wu, Xu~Jiang, Diogo Almeida, Carroll Wainwright, Pamela Mishkin, Chong Zhang, Sandhini Agarwal, Katarina Slama, Alex Ray, John Schulman, Jacob Hilton, Fraser Kelton, Luke Miller, Maddie Simens, Amanda Askell, Peter Welinder, Paul~F Christiano, Jan Leike, and Ryan Lowe.
\newblock Training language models to follow instructions with human feedback.
\newblock In S.~Koyejo, S.~Mohamed, A.~Agarwal, D.~Belgrave, K.~Cho, and A.~Oh, editors, {\em Advances in Neural Information Processing Systems (NeurIPS)}, volume~35, pages 27730--27744, 2022.

\bibitem{owen1990empirical}
Art Owen.
\newblock Empirical likelihood ratio confidence regions.
\newblock {\em The Annals of Statistics}, 18(1):90--120, 1990.

\bibitem{o2019expert}
Anthony O’Hagan.
\newblock Expert knowledge elicitation: subjective but scientific.
\newblock {\em The American Statistician}, 73(sup1):69--81, 2019.

\bibitem{ramachandran2020incorporating}
Anil Ramachandran, Sunil Gupta, Santu Rana, Cheng Li, and Svetha Venkatesh.
\newblock Incorporating expert prior in {B}ayesian optimisation via space warping.
\newblock {\em Knowledge-Based Systems}, 195:105663, 2020.

\bibitem{rodemann2024explaining}
Julian Rodemann, Federico Croppi, Philipp Arens, Yusuf Sale, Julia Herbinger, Bernd Bischl, Eyke H{\"u}llermeier, Thomas Augustin, Conor~J Walsh, and Giuseppe Casalicchio.
\newblock Explaining {B}ayesian optimization by {S}hapley values facilitates human-{AI} collaboration.
\newblock {\em arXiv preprint arXiv:2403.04629}, 2024.

\bibitem{rosenbrock1960automatic}
Howard~Harry Rosenbrock.
\newblock An automatic method for finding the greatest or least value of a function.
\newblock {\em The computer journal}, 3(3):175--184, 1960.

\bibitem{rosol2009solubility}
Zachary~P Rosol, Natalie~J German, and Stephen~M Gross.
\newblock Solubility, ionic conductivity and viscosity of lithium salts in room temperature ionic liquids.
\newblock {\em Green Chemistry}, 11(9):1453--1457, 2009.

\bibitem{rousseau2001schema}
Denise~M Rousseau.
\newblock Schema, promise and mutuality: The building blocks of the psychological contract.
\newblock {\em Journal of occupational and organizational psychology}, 74(4):511--541, 2001.

\bibitem{ru2020interpretable}
Binxin Ru, Xingchen Wan, Xiaowen Dong, and Michael Osborne.
\newblock Interpretable neural architecture search via {B}ayesian optimisation with {W}eisfeiler-{L}ehman kernels.
\newblock In {\em International Conference on Learning Representations (ICLR)}, 2021.

\bibitem{sanh2022multitask}
Victor Sanh, Albert Webson, Colin Raffel, Stephen Bach, Lintang Sutawika, Zaid Alyafeai, Antoine Chaffin, Arnaud Stiegler, Arun Raja, Manan Dey, M~Saiful Bari, Canwen Xu, Urmish Thakker, Shanya~Sharma Sharma, Eliza Szczechla, Taewoon Kim, Gunjan Chhablani, Nihal Nayak, Debajyoti Datta, Jonathan Chang, Mike Tian-Jian Jiang, Han Wang, Matteo Manica, Sheng Shen, Zheng~Xin Yong, Harshit Pandey, Rachel Bawden, Thomas Wang, Trishala Neeraj, Jos Rozen, Abheesht Sharma, Andrea Santilli, Thibault Fevry, Jason~Alan Fries, Ryan Teehan, Teven~Le Scao, Stella Biderman, Leo Gao, Thomas Wolf, and Alexander~M Rush.
\newblock Multitask prompted training enables zero-shot task generalization.
\newblock In {\em International Conference on Learning Representations (ICLR)}, 2022.

\bibitem{scarlett2017lower}
Jonathan Scarlett, Ilija Bogunovic, and Volkan Cevher.
\newblock Lower bounds on regret for noisy {G}aussian process bandit optimization.
\newblock In {\em Conference on Learning Theory}, pages 1723--1742. PMLR, 2017.

\bibitem{scholkopf2001generalized}
Bernhard Sch{\"o}lkopf, Ralf Herbrich, and Alex~J Smola.
\newblock A generalized representer theorem.
\newblock In {\em International Conference on Computational Learning Theory}, pages 416--426. Springer, 2001.

\bibitem{senge2014reliable}
Robin Senge, Stefan B{\"o}sner, Krzysztof Dembczy{\'n}ski, J{\"o}rg Haasenritter, Oliver Hirsch, Norbert Donner-Banzhoff, and Eyke H{\"u}llermeier.
\newblock Reliable classification: Learning classifiers that distinguish aleatoric and epistemic uncertainty.
\newblock {\em Information Sciences}, 255:16--29, 2014.

\bibitem{shadbolt2015knowledge}
Nigel~R Shadbolt, Paul~R Smart, J~Wilson, and S~Sharples.
\newblock Knowledge elicitation.
\newblock {\em Evaluation of human work}, pages 163--200, 2015.

\bibitem{shahriari2015taking}
Bobak Shahriari, Kevin Swersky, Ziyu Wang, Ryan~P Adams, and Nando De~Freitas.
\newblock Taking the human out of the loop: A review of {B}ayesian optimization.
\newblock {\em Proceedings of the IEEE}, 104(1):148--175, 2015.

\bibitem{shin-etal-2020-autoprompt}
Taylor Shin, Yasaman Razeghi, Robert~L. Logan~IV, Eric Wallace, and Sameer Singh.
\newblock {A}uto{P}rompt: {E}liciting {K}nowledge from {L}anguage {M}odels with {A}utomatically {G}enerated {P}rompts.
\newblock In {\em Empirical Methods in Natural Language Processing (EMNLP)}, pages 4222--4235, Online, November 2020. Association for Computational Linguistics.

\bibitem{souza2021bayesian}
Artur Souza, Luigi Nardi, Leonardo~B Oliveira, Kunle Olukotun, Marius Lindauer, and Frank Hutter.
\newblock Bayesian optimization with a prior for the optimum.
\newblock In {\em Joint European Conference on Machine Learning and Knowledge Discovery in Databases (ECML)}, pages 265--296. Springer, 2021.

\bibitem{srinivas2009gaussian}
Niranjan Srinivas, Andreas Krause, Sham~M Kakade, and Matthias Seeger.
\newblock Gaussian process optimization in the bandit setting: No regret and experimental design.
\newblock In {\em International Conference on Machine Learning (ICML)}, pages 1015--1022, 2010.

\bibitem{srinivas2012information}
Niranjan Srinivas, Andreas Krause, Sham~M Kakade, and Matthias~W Seeger.
\newblock Information-theoretic regret bounds for {G}aussian process optimization in the bandit setting.
\newblock {\em IEEE Transactions on Information Theory}, 58(5):3250--3265, 2012.

\bibitem{srivastava2022beyond}
Aarohi Srivastava, Abhinav Rastogi, Abhishek Rao, Abu Awal~Md Shoeb, Abubakar Abid, Adam Fisch, Adam~R Brown, Adam Santoro, Aditya Gupta, Adri{\`a} Garriga-Alonso, et~al.
\newblock Beyond the imitation game: Quantifying and extrapolating the capabilities of language models.
\newblock {\em arXiv preprint arXiv:2206.04615}, 2022.

\bibitem{stein1999interpolation}
Michael~L Stein.
\newblock {\em Interpolation of spatial data}.
\newblock Springer Science \& Business Media, 1999.

\bibitem{sui2018stagewise}
Yanan Sui, Joel Burdick, Yisong Yue, et~al.
\newblock Stage-wise safe {B}ayesian optimization with {G}aussian processes.
\newblock In {\em Proc. of the Int. Conf. on Mach. Learn.}, pages 4781--4789, 2018.

\bibitem{sui2015safe}
Yanan Sui, Alkis Gotovos, Joel Burdick, and Andreas Krause.
\newblock Safe exploration for optimization with {G}aussian processes.
\newblock In {\em Proc. of the Int. Conf. on Mach. Learn.}, pages 997--1005, 2015.

\bibitem{simulationlib}
S.~Surjanovic and D.~Bingham.
\newblock Virtual library of simulation experiments: Test functions and datasets.
\newblock Retrieved May 17, 2024, from \url{http://www.sfu.ca/~ssurjano}, 2024.

\bibitem{swersky2013multi}
Kevin Swersky, Jasper Snoek, and Ryan~P Adams.
\newblock Multi-task {B}ayesian optimization.
\newblock {\em Advances in Neural Information Processing Systems (NeurIPS)}, 26, 2013.

\bibitem{takeno2023towards}
Shion Takeno, Masahiro Nomura, and Masayuki Karasuyama.
\newblock Towards practical preferential {B}ayesian optimization with skew {G}aussian processes.
\newblock In {\em International Conference on Machine Learning (ICML)}, pages 33516--33533. PMLR, 2023.

\bibitem{tsipras2018robustness}
Dimitris Tsipras, Shibani Santurkar, Logan Engstrom, Alexander Turner, and Aleksander Madry.
\newblock Robustness may be at odds with accuracy.
\newblock In {\em International Conference on Learning Representations (ICLR)}, 2019.

\bibitem{vakili2021information}
Sattar Vakili, Kia Khezeli, and Victor Picheny.
\newblock On information gain and regret bounds in {G}aussian process bandits.
\newblock In {\em International Conference on Artificial Intelligence and Statistics (AISTATS)}, pages 82--90. PMLR, 2021.

\bibitem{vapnik1998statistical}
Vladimir~Naumovich Vapnik, Vlamimir Vapnik, et~al.
\newblock {\em Statistical learning theory}.
\newblock wiley New York, 1998.

\bibitem{wachter2006implementation}
Andreas W{\"a}chter and Lorenz~T Biegler.
\newblock On the implementation of an interior-point filter line-search algorithm for large-scale nonlinear programming.
\newblock {\em Mathematical Programming}, 106(1):25--57, 2006.

\bibitem{wang2024constrained}
Shengbo Wang and Ke~Li.
\newblock Constrained {B}ayesian optimization under partial observations: Balanced improvements and provable convergence.
\newblock In {\em Proceedings of the AAAI Conference on Artificial Intelligence}, pages 15607--15615, 2024.

\bibitem{wang2016bayesian}
Ziyu Wang, Frank Hutter, Masrour Zoghi, David Matheson, and Nando De~Feitas.
\newblock Bayesian optimization in a billion dimensions via random embeddings.
\newblock {\em Journal of Artificial Intelligence Research}, 55:361--387, 2016.

\bibitem{wei2022chain}
Jason Wei, Xuezhi Wang, Dale Schuurmans, Maarten Bosma, Fei Xia, Ed~Chi, Quoc~V Le, Denny Zhou, et~al.
\newblock Chain-of-thought prompting elicits reasoning in large language models.
\newblock {\em Advances in Neural Information Processing Systems (NeurIPS)}, 35:24824--24837, 2022.

\bibitem{white2021bananas}
Colin White, Willie Neiswanger, and Yash Savani.
\newblock Bananas: {B}ayesian optimization with neural architectures for neural architecture search.
\newblock In {\em Proceedings of the AAAI Conference on Artificial Intelligence (AAAI)}, pages 10293--10301, 2021.

\bibitem{williams2006gaussian}
Christopher~KI Williams and Carl~Edward Rasmussen.
\newblock {\em Gaussian processes for machine learning}.
\newblock MIT press Cambridge, MA, 2006.

\bibitem{wistuba2020few}
Martin Wistuba and Josif Grabocka.
\newblock Few-shot {B}ayesian optimization with deep kernel surrogates.
\newblock In {\em International Conference on Learning Representations (ICLR)}, 2020.

\bibitem{wolpert1997no}
David~H Wolpert and William~G Macready.
\newblock No free lunch theorems for optimization.
\newblock {\em IEEE transactions on evolutionary computation}, 1(1):67--82, 1997.

\bibitem{wu2020practical}
Jian Wu, Saul Toscano-Palmerin, Peter~I Frazier, and Andrew~Gordon Wilson.
\newblock Practical multi-fidelity {B}ayesian optimization for hyperparameter tuning.
\newblock In {\em Uncertainty in Artificial Intelligence (UAI)}, pages 788--798. PMLR, 2020.

\bibitem{wu2017lecture}
Yihong Wu.
\newblock Lecture notes on information-theoretic methods for high-dimensional statistics.
\newblock {\em Lecture Notes for ECE598YW (UIUC)}, 16, 2017.

\bibitem{xu2024lower}
Wenjie Xu, Yuning Jiang, Emilio~T Maddalena, and Colin~N Jones.
\newblock Lower bounds on the noiseless worst-case complexity of efficient global optimization.
\newblock {\em Journal of Optimization Theory and Applications}, pages 1--26, 2024.

\bibitem{xu2023constrained}
Wenjie Xu, Yuning Jiang, Bratislav Svetozarevic, and Colin Jones.
\newblock Constrained efficient global optimization of expensive black-box functions.
\newblock In {\em International Conference on Machine Learning (ICML)}, pages 38485--38498. PMLR, 2023.

\bibitem{xu2024principled}
Wenjie Xu, Wenbin Wang, Yuning Jiang, Bratislav Svetozarevic, and Colin~N Jones.
\newblock Principled preferential {B}ayesian optimization.
\newblock {\em arXiv preprint arXiv:2402.05367}, 2024.

\bibitem{zhang2016flexible}
Jingxian Zhang, Ning Zhao, Miao Zhang, Yiqiu Li, Paul~K Chu, Xiangxin Guo, Zengfeng Di, Xi~Wang, and Hong Li.
\newblock Flexible and ion-conducting membrane electrolytes for solid-state lithium batteries: Dispersion of garnet nanoparticles in insulating polyethylene oxide.
\newblock {\em Nano Energy}, 28:447--454, 2016.

\bibitem{zhou2002covering}
Ding-Xuan Zhou.
\newblock The covering number in learning theory.
\newblock {\em Journal of Complexity}, 18(3):739--767, 2002.

\bibitem{zhou2022kernelized}
Xingyu Zhou and Bo~Ji.
\newblock On kernelized multi-armed bandits with constraints.
\newblock {\em Advances in Neural Information Processing Systems (NeurIPS)}, 35, 2022.

\end{thebibliography}

\appendix
\newpage
\addcontentsline{toc}{section}{Appendix} 
\part{Appendix} 
\parttoc 

\section{Proof of Lem.~\ref{thm:conf_set_g}}\label{proof:conf_set_g}
To prepare for the proof of the lemma, we first prove several preliminary lemmas.

\begin{lemma} For any fixed $\hat{g}\in\mathcal{B}_g$, we have,
\begin{equation}
  \Prob\left(\log\Prob_{\hat{g}}((x_\tau, \onetau)_{\tau\in\Ct})-\log\Prob_{{g}}((x_\tau, \onetau)_{\tau\in\Ct})\leq \sqrt{8\aCt B_\fl^2\log\frac{1}{\delta_{t}}}\right)\geq 1-\delta_{t}. 
\end{equation}
\end{lemma}
\begin{proof}
We use $\gy_\tau$ to denote $g(x_\tau)$, $z_\tau$ to denote $\hat{g}(x_\tau)$, and $p_\tau$ to denote $\sigmoid(g(x_\tau))$. 
\begin{align}
&\Prob\left(\log\Prob_{\hat{g}}((x_\tau, \onetau)_{\tau\in\Ct})-\log\Prob_{{g}}((x_\tau, \onetau)_{\tau\in\Ct})\leq \xi\right)\\
=& \Prob\left(\sum_{\tau\in\Ct} \left((z_\tau-\gy_\tau)\onetau-\log(1+e^{z_\tau})+\log(1+e^{\gy_\tau})\right) \leq \xi\right) \\
=&\Prob\left(\sum_{\tau\in\Ct} (z_\tau-\gy_\tau)\onetau -\sum_{\tau\in\Ct} (z_\tau-\gy_\tau)p_\tau \leq \xi^\prime\right) \label{eq:transform_likelihood_val}
\end{align}
where the probability $\Prob$ is taken over the randomness from the feedback expert/oracle and the randomness from the algorithm, and $\xi^\prime=\xi+\sum_{\tau\in\Ct}\log\left(1+e^{z_\tau}\right)-\sum_{\tau\in\Ct}\log\left(1+e^{\gy_\tau}\right)-\sum_{\tau\in\Ct} (z_\tau-\gy_\tau)p_\tau$.
Let the function $\psi_\tau(z_\tau)\defeq\log\left(1+e^{z_\tau}\right)-\log\left(1+e^{\gy_\tau}\right)-(z_\tau-\gy_\tau)p_\tau$. It can be checked that $\psi_\tau^{''}(z_\tau)=\nicefrac{e^{z_\tau}}{(1+e^{z_\tau})^2}\geq0,\forall z_\tau\in\mathbb{R}$ and $\psi_\tau^{'}(\gy_\tau)=0$. Therefore, $\psi_\tau$ is a convex function and achieves the optimal value at the point $\gy_\tau$. Hence, $\psi_\tau(z_\tau)\geq\psi_\tau(\gy_\tau)=0$, which implies $\xi^\prime\geq\xi$. Therefore,  
\begin{equation}
    \Prob\left(\sum_{\tau\in\Ct} (z_\tau-\gy_\tau)\onetau -\sum_{\tau\in\Ct} (z_\tau-\gy_\tau)p_\tau \leq \xi^\prime\right)\geq\Prob\left(\sum_{\tau\in\Ct} (z_\tau-\gy_\tau)\onetau -\sum_{\tau\in\Ct} (z_\tau-\gy_\tau)p_\tau \leq \xi\right). \label{eq:bound_prob}
\end{equation}
Furthermore, it is easy to see that $(z_\tau-\gy_\tau)\onetau\in[-2B_\gl, 2B_\gl]$, and thus, by applying Azuma-Hoeffding inequality, we have, 
\begin{equation} 
    \Prob\left(\sum_{\tau\in\Ct} (z_\tau-\gy_\tau)\onetau -\sum_{\tau\in\Ct}(z_\tau-\gy_\tau)p_\tau \leq \xi\right)\geq1-\exp\left\{-\frac{\xi^2}{8|\Ct|B_\gl^2}\right\}
\end{equation}
Let $\exp\left\{-\frac{\xi^2}{8|\Ct|B_\gl^2}\right\}\leq\delta_{t}$, we need,
\begin{equation}
    \xi\geq \sqrt{8|\Ct|B_\gl^2\log\frac{1}{\delta_{t}}}.
\end{equation}
It is sufficient to pick $\xi=\sqrt{8|\Ct|B_\gl^2\log\frac{1}{\delta_{t}}}$. 
Therefore,
\begin{align*}
 & \Prob\left(\log\Prob_{\hat{g}}((x_\tau, \onetau)_{\tau\in\Ct})-\log\Prob_{{g}}((x_\tau, \onetau)_{\tau\in\Ct})\leq \sqrt{8\aCt B_\gl^2\log\frac{1}{\delta_{t}}}\right)\\
\geq&\Prob\left(\sum_{\tau\in\Ct} (z_\tau-\gy_\tau)\onetau -\sum_{\tau\in\Ct}(z_\tau-\gy_\tau)p_\tau \leq \sqrt{8|\Ct|B_\gl^2\log\frac{1}{\delta_{t}}} \right) \\
\geq & 1-\delta_t, 
\end{align*}
where the first inequality follows by combining Eq.~\eqref{eq:transform_likelihood_val} and Eq.~\eqref{eq:bound_prob}. 

\end{proof}

We then have the following high probability confidence set lemma.
\begin{lemma}
\label{lem:fixed_f_log_P}
For any fixed $\hat{g}$ that is independent of $((x_\tau, \onetau)_{\tau\in\Ct})$, we have, with probability at least $1-\delta$, $\forall t\geq1$, 
\begin{equation}
  \log\Prob_{\hat{g}}((x_\tau, \onetau)_{\tau\in\Ct})-\log\Prob_{{g}}((x_\tau, \onetau)_{\tau\in\Ct})\leq \sqrt{8\aCt B_\gl^2\log\frac{\pi^2t^2}{6\delta}}. 
\end{equation}
\end{lemma}
\begin{proof}
We use $\mathcal{E}_t$ to denote the event $\log\Prob_{\hat{g}}((x_\tau, \onetau)_{\tau\in\Ct})-\log\Prob_{{g}}((x_\tau, \onetau)_{\tau\in\Ct})\leq \sqrt{8\aCt B_\gl^2\log\frac{1}{\delta_t}}$. We pick $\delta_t=\nicefrac{(6\delta)}{(\pi^2 t^2)}$ and have,
\begin{align*}
  &\Prob\left(\log\Prob_{\hat{g}}((x_\tau, \onetau)_{\tau\in\Ct})-\log\Prob_{{g}}((x_\tau, \onetau)_{\tau\in\Ct})\leq \sqrt{8\aCt B_\gl^2\log\frac{1}{\delta_t}}, \forall t\geq1\right) \\
  =& 1-\Prob\left(\overline{\cap_{t=1}^\infty \mathcal{E}_t}\right)\\
  =& 1-\Prob\left({\cup_{t=1}^\infty \overline{\mathcal{E}_t}}\right)\\
  \geq& 1-\sum_{t=1}^\infty\Prob\left( \overline{\mathcal{E}_t}\right)\\
  =& 1-\sum_{t=1}^\infty\left(1-\Prob\left({\mathcal{E}_t}\right)\right)\\
 =& 1-\sum_{t=1}^\infty\left(1-\Prob\left({\log\Prob_{\hat{g}}((x_\tau, \onetau)_{\tau\in\Ct})-\log\Prob_{{g}}((x_\tau, \onetau)_{\tau\in\Ct})\leq \sqrt{8\aCt B_\gl^2\log\frac{1}{\delta_t}}}\right)\right)\\
 \geq&1-\sum_{t=1}^\infty\delta_t\\
 = &1-\frac{6\delta}{\pi^2}\sum_{t=1}^\infty\frac{1}{t^2}\\
= & 1-\delta. 
\end{align*}
\end{proof}

We then have a lemma to bound the difference of log likelihood when two functions are close in infinity-norm sense. 
\begin{lemma}
\label{lem:close_f}
$\forall \epsilon>0$, $\forall g_1, g_2\in\mathcal{B}_g$ that satisfies $\|g_1-g_2\|_\infty\leq\epsilon$, we have,
\begin{equation}
     \log\Prob_{{g}_1}((x_\tau, \onetau)_{\tau\in\Ct})-\log\Prob_{{g}_2}((x_\tau, \onetau)_{\tau\in\Ct})\leq 2\epsilon \aCt. 
\end{equation}
\end{lemma}
\begin{proof}

\begin{align*}
 &\log\Prob_{{g}_1}((x_\tau,\onetau)_{\tau\in\Ct})-\log\Prob_{{g}_2}((x_\tau, \onetau)_{\tau\in\Ct})\\
\leq& \sum_{\tau\in\Ct} \left((z_{1,\tau}-z_{2,\tau})\onetau-\log(1+e^{z_{1,\tau}})+\log(1+e^{z_{2,\tau}})\right)\\
\leq&\epsilon \aCt+\sum_{\tau\in\Ct}\max_{z\in[-B_\gl, B_\gl]}\left|\nabla_{z}\log\left(1+e^{z}\right)\right|\left|z_{1,\tau}-z_{2,\tau}\right|\\
\leq&\epsilon \aCt+\sum_{\tau\in\Ct}\epsilon\\
\leq&2\epsilon \aCt,
\end{align*}
where $z_{1,\tau}=g_1(x_\tau)$ and $z_{2,\tau}=g_2(x_\tau)$.
\end{proof}

We use $\mathcal{N}(\mathcal{B}_g,\epsilon,\|\cdot\|_\infty)$ to denote the covering number of the set $\mathcal{B}_g$, with $(g^\epsilon_i)_{i=1}^{\mathcal{N}(\mathcal{B}_g,\epsilon,\|\cdot\|_\infty)}$ be a set of $\epsilon$-covering for the set $\mathcal{B}_g$. Set the `$\delta$' in Lem.~\ref{lem:fixed_f_log_P} as $\nicefrac{\delta}{\mathcal{N}(\mathcal{B}_g,\epsilon,\|\cdot\|_\infty)}$ and applying the probability union bound, we have, with probability at least $1-\delta$, $\forall g_i^\epsilon$, 
\begin{equation}
  \log\Prob_{{g}_i^\epsilon}((x_\tau, \onetau)_{\tau\in\Ct})-\log\Prob_{{g}}((x_\tau, \onetau)_{\tau\in\Ct})\leq \sqrt{8\aCt B_\gl^2\log\frac{\pi^2t^2\mathcal{N}(\mathcal{B}_g,\epsilon,\|\cdot\|_\infty)}{6\delta}}. 
\end{equation}
By the definition of $\epsilon$-covering, there exists $j\in[\mathcal{N}(\mathcal{B}_g,\epsilon,\|\cdot\|_\infty)]$, such that,
\begin{equation}
   \|\hat{g}_{t+1}^\mathrm{MLE}-g_j^\epsilon\|_\infty\leq\epsilon. 
\end{equation}
Hence, with probability at least $1-\delta$, 
\begin{align*}
  &\log\Prob_{\hat{g}_{t+1}^\mathrm{MLE}}((x_\tau,\onetau)_{\tau\in\Ct})-\log\Prob_{{g}}((x_\tau, \onetau)_{\tau\in\Ct})\\
  =&\log\Prob_{\hat{g}_{t+1}^\mathrm{MLE}}((x_\tau, \onetau)_{\tau\in\Ct})-\log\Prob_{{g}_{j}^\epsilon}((x_\tau, \onetau)_{\tau\in\Ct})+\log\Prob_{{g}_{j}^\epsilon}((x_\tau, \onetau)_{\tau\in\Ct})-\log\Prob_{{g}}((x_\tau, \onetau)_{\tau\in\Ct})\\
  \leq& 2\epsilon \aCt+\sqrt{8\aCt B_\gl^2\log\frac{\pi^2t^2\mathcal{N}(\mathcal{B}_g,\epsilon,\|\cdot\|_\infty)}{6\delta}}, 
\end{align*}
where the inequality follows by Lem.~\ref{lem:close_f} and Lem.~\ref{lem:fixed_f_log_P}.

\section{Proof of Thm.~\ref{thm:RQ_bound}}
\label{app:rq_bound_proof}
\subsection{Bound Error over Historical Evaluations}
\label{sec:aug_RKHS}
Lem.~\ref{thm:conf_set_g} gives a high confidence set based on the likelihood function. The following Lem.~\ref{lem:obj_ell_po} further gives error bound over the historical sample points. Lem.~\ref{lem:obj_ell_po} highlights that with high probability, all the functions in the confidence set have values over the historical sample points that lie in a ball with the ground-truth function value as the center and $\sqrt{\betag(\epsilon, \nicefrac{\delta}{2}, \aCt, t)}$ as the radius. Before we proceed, we first introduce several constants that we will use, 
\begin{equation}
   \bar{\sigmoid}\defeq\max_{u\in[-B_\gl, B_\gl]}\sigmoid(u)=\frac{1}{1+e^{-B_\gl}}, \underline{\sigmoid}\defeq\min_{u\in[-B_\gl, B_\gl]}\sigmoid(u)=\frac{1}{1+e^{B_\gl}}.
\end{equation}

\begin{equation}
    \underline{\sigmoid^\prime}\defeq\min_{u\in[-B_\gl, B_\gl]}\sigmoid^\prime(u)=\frac{1}{e^{B_\gl}+e^{-B_\gl}+2},  \bar{\sigmoid^\prime}\defeq\max_{u\in[-B_\gl, B_\gl]}\sigmoid^\prime(u)=\frac{1}{4}.
\end{equation}

\begin{equation}
    H_\sigmoid\defeq\frac{1}{2\bar{\sigmoid}^2}, B_p=\frac{\sigmoid(B_\gl)}{\sigmoid(-B_\gl)}-\frac{\sigmoid(-B_\gl)}{\sigmoid(B_\gl)}.
\end{equation}

\begin{lemma}\label{lem:obj_ell_po} For any estimate $\hat{g}_{t+1}\in\mathcal{B}_g^{t+1}$ that is measurable with respect to the filtration $\mathcal{F}_{t}$, we have, with probability at least $1-\nicefrac{\delta}{2}$, $\forall t\geq1$, 
\begin{equation}
  \sum_{\tau\in\Ct}\left(\hat{g}_{t+1}(x_\tau)-{g}(x_\tau)\right)^2\leq\betag(\epsilon, \nicefrac{\delta}{2}, \aCt, t), \label{eq:bound_over_hist} 
\end{equation}
and 
\begin{equation}
   \tg\in\mathcal{B}_g^{t+1}, 
\end{equation}
where $\betag(\epsilon, \nicefrac{\delta}{2}, \aCt, t)=\frac{\underline{\sigmoid^\prime}^2}{H_\sigmoid}\left(\betag_2(\epsilon,\nicefrac{\delta}{2}, \aCt, t)+2\betag_1(\epsilon, \nicefrac{\delta}{2}, \aCt, t)\right)=\mathcal{O}\left(\sqrt{\aCt\log\frac{t\mathcal{N}(\mathcal{B}_g, \epsilon,\|\cdot\|_\infty)}{\delta}}+\epsilon t+\epsilon^2t\right)$, with $\betag_2(\epsilon, \delta, \aCt, t)=8H_\sigmoid\bar{\sigmoid^\prime}^2\epsilon^2t+4\epsilon t+\sqrt{8{\aCt B_p^2\log\frac{\pi^2t^2\mathcal{N}(\mathcal{B}_g,\epsilon, \|\cdot\|_\infty)}{3\delta}}{}}$. 
\end{lemma}

\begin{proof}
    For any fixed function $\hat{g}$, we have,
\begin{align*}
&\log{\Prob_{\hat{g}}}((x_\tau, \onetau)_{\tau\in\Ct})-\log{\Prob_{{g}}}((x_\tau, \onetau)_{\tau\in\Ct})\\
=&\sum_{\tau\in\Ct}\left(\log{\Prob_{\hat{g}}}((x_\tau,\onetau))-\log{\Prob_{{g}}}((x_\tau,\onetau))\right)\\
=&\sum_{\tau\in\Ct}\left(\onetau\left(\log{\hat{p}_\tau}-\log{p}_\tau\right)+(1-\onetau)\left(\log{(1-\hat{p}_\tau)}-\log(1-{p}_\tau)\right)\right),
\end{align*}
where $\hat{p}_\tau=\sigmoid(\hat{g}(x_\tau))$ and ${p}_\tau=\sigmoid({g}(x_\tau))$.
We have, 
\begin{equation}
\log y \leq \log x +\frac{1}{x}(y-x)-H_\sigmoid(y-x)^2, \forall x,y\in[\underline{\sigmoid}, \bar{\sigmoid}], 
\end{equation}
where $H_\sigmoid=\frac{1}{2\bar{\sigmoid}^2}$.
Hence,
\begin{align*}
&\log{\Prob_{\hat{g}}}((x_\tau,\onetau)_{\tau\in\Ct})-\log{\Prob_{{g}}}((x_\tau,\onetau)_{\tau\in\Ct})\\
=&\sum_{\tau\in\Ct}\left(\onetau\left(\log{\hat{p}_\tau}-\log{p}_\tau\right)+(1-\onetau)\left(\log{(1-\hat{p}_\tau)}-\log(1-{p}_\tau)\right)\right)\\
\leq&\sum_{\tau\in\Ct}\left(\onetau\left(\frac{\hat{p}_\tau-p_\tau}{p_\tau}-H_\sigmoid\left(\hat{p}_\tau-p_\tau\right)^2\right)+(1-\onetau)\left(\frac{{p}_\tau-\hat{p}_\tau}{1-p_\tau}-H_\sigmoid\left(\hat{p}_\tau-p_\tau\right)^2\right)\right)
\end{align*}
Rearrangement gives,
\begin{align*}
&H_\sigmoid\sum_{\tau\in\Ct}\left(\hat{p}_\tau-p_\tau\right)^2+\log{\Prob_{\hat{g}}}((x_\tau, \onetau)_{\tau\in\Ct})-\log{\Prob_{{g}}}((x_\tau,\onetau)_{\tau\in\Ct})\\
\leq&\sum_{\tau\in\Ct}\left(\onetau\frac{\hat{p}_\tau-p_\tau}{p_\tau}+(1-\onetau)\frac{{p}_\tau-\hat{p}_\tau}{1-p_\tau}\right).
\end{align*}
Since $\E\left[\onetau\frac{\hat{p}_\tau-p_\tau}{p_\tau}+(1-\onetau)\frac{{p}_\tau-\hat{p}_\tau}{1-p_\tau}|\mathcal{F}_{\tau-1}\right]=\E\left[p_\tau\frac{\hat{p}_\tau-p_\tau}{p_\tau}+(1-p_\tau)\frac{{p}_\tau-\hat{p}_\tau}{1-p_\tau}|\mathcal{F}_{\tau-1}\right]=0$ and with probability one, 
\begin{align}
\left\lvert{\onetau\frac{\hat{p}_\tau-p_\tau}{p_\tau}+(1-\onetau)\frac{{p}_\tau-\hat{p}_\tau}{1-p_\tau}}\right\rvert&\leq\onetau\left\lvert{\frac{\hat{p}_\tau-p_\tau}{p_\tau}}\right\rvert+(1-\onetau)\left\lvert{\frac{{p}_\tau-\hat{p}_\tau}{1-p_\tau}}\right\rvert\\
&=\onetau\left\lvert{\frac{\hat{p}_\tau}{p_\tau}-1}\right\rvert+(1-\onetau)\left\lvert{\frac{1-\hat{p}_\tau}{1-p_\tau}-1}\right\rvert\\
&\leq \frac{\sigmoid(B_\gl)}{\sigmoid(-B_\gl)}-\frac{\sigmoid(-B_\gl)}{\sigmoid(B_\gl)}=B_p.
\end{align}
By Azuma–Hoeffding inequality, we have, $\forall\xi>0$,
\begin{align}
&\Prob\left(\sum_{\tau\in\Ct}\left(\onetau\frac{\hat{p}_\tau-p_\tau}{p_\tau}+(1-\onetau)\frac{{p}_\tau-\hat{p}_\tau}{1-p_\tau}\right)\leq\xi\right)\geq 1-\exp\left\{-\frac{2\xi^2}{|\Ct|B_p^2}\right\}.
\end{align}
We set $\exp\left\{-\frac{2\xi^2}{\aCt B_p^2}\right\}=\delta_t>0$, and derive
\begin{align}
&\Prob\left(H_\sigmoid\sum_{\tau\in\Ct}\left(\hat{p}_\tau-p_\tau\right)^2+\log{\Prob_{\hat{g}}}((x_\tau, \onetau)_{\tau\in\Ct})-\log{\Prob_{{g}}}((x_\tau,\onetau)_{\tau\in\Ct})\leq \sqrt{\frac{\aCt B_p^2\log\frac{1}{\delta_t}}{2}}\right)\\ 
\geq&\Prob\left(\sum_{\tau\in\Ct}\left(\onetau\frac{\hat{p}_\tau-p_\tau}{p_\tau}+(1-\onetau)\frac{{p}_\tau-\hat{p}_\tau}{1-p_\tau}\right)\leq\sqrt{\frac{\aCt B_p^2\log\frac{1}{\delta_t}}{2}}\right)\\
 \geq& 1-\delta_t.    
\end{align}
    

We use $\mathcal{E}_t$ to denote the event $H_\sigmoid\sum_{\tau\in\Ct}\left(\hat{p}_\tau-p_\tau\right)^2\leq\log{\Prob_{{g}}}((x_\tau, \onetau)_{\tau\in\Ct})-\log{\Prob_{\hat{g}}}((x_\tau,\onetau)_{\tau\in\Ct})+\sqrt{\frac{\aCt B_p^2\log\frac{1}{\delta_t}}{2}}$.
We pick $\delta_t=\nicefrac{(6\delta)}{(\pi^2 t^2)}$.
We have,
\begin{align*}
  &\Prob\left(H_\sigmoid\sum_{\tau\in\Ct}\left(\hat{p}_\tau-p_\tau\right)^2\leq\log{\Prob_{{g}}}((x_\tau, \onetau)_{\tau\in\Ct})-\log{\Prob_{\hat{g}}}((x_\tau, \onetau)_{\tau\in\Ct})+\sqrt{\frac{\aCt B_p^2\log\frac{1}{\delta_t}}{2}}, \forall t\geq1\right) \\
  =& 1-\Prob\left(\overline{\cap_{t=1}^\infty \mathcal{E}_t}\right)\\
  =& 1-\Prob\left({\cup_{t=1}^\infty \overline{\mathcal{E}_t}}\right)\\
  \geq& 1-\sum_{t=1}^\infty\Prob\left( \overline{\mathcal{E}_t}\right)\\
  =& 1-\sum_{t=1}^\infty\left(1-\Prob\left({\mathcal{E}_t}\right)\right)\\
 =& 1-\sum_{t=1}^\infty\left(1-\Prob\left({H_\sigmoid\sum_{\tau\in\Ct}\left(\hat{p}_\tau-p_\tau\right)^2\leq\log{\Prob_{{g}}}((x_\tau, \onetau)_{\tau\in\Ct})-\log{\Prob_{\hat{g}}}((x_\tau,\onetau)_{\tau\in\Ct})+\sqrt{\frac{\aCt B_p^2\log\frac{1}{\delta_t}}{2}}}\right)\right)\\
 \geq&1-\sum_{t=1}^\infty\delta_t\\
 = &1-\frac{6\delta}{\pi^2}\sum_{t=1}^\infty\frac{1}{t^2}\\
= & 1-\delta.\label{inq:p_square_bound} 
\end{align*}

Resetting the `$\delta$' to be $\nicefrac{\delta}{\mathcal{N}(\mathcal{B}_g, \epsilon, \|\cdot\|_\infty)}$, we can guarantee the inequality~\eqref{inq:p_square_bound} holds for all the functions in an $\epsilon$-covering of $\mathcal{B}_g$.   

For any $\hat{g}_{t+1}\in\mathcal{B}_g^{t+1}$, there exists $\hat{g}$ in the $\epsilon$-covering of $\mathcal{B}_g$, such that $\|\hat{g}^{t+1}-\hat{g}\|_\infty\leq\epsilon$. We use the notations $\hat{p}^{t+1}_\tau=\hat{g}_{t+1}(x_\tau)$, and $\hat{p}_\tau=\hat{g}(x_\tau)$.
Thus, we have,
\begin{align*}
  &  H_\sigmoid\sum_{\tau\in\Ct}\left(\hat{p}^{t+1}_\tau-p_\tau\right)^2\\
  =&  2H_\sigmoid\sum_{\tau\in\Ct}\left(\hat{p}^{t+1}_\tau-\hat{p}_\tau\right)^2+2H_\sigmoid\sum_{\tau\in\Ct}\left(\hat{p}_\tau-{p}_\tau\right)^2 \\
  =& 2H_\sigmoid\bar{\sigmoid^\prime}^2\sum_{\tau\in\Ct}\left(\hat{g}^{t+1}(x_\tau)-\hat{g}(x_\tau)\right)^2+2H_\sigmoid\sum_{\tau\in\Ct}\left(\hat{p}_\tau-{p}_\tau\right)^2\\
  \leq& 8H_\sigmoid\bar{\sigmoid^\prime}^2\sum_{\tau\in\Ct}\epsilon^2+2H_\sigmoid\sum_{\tau\in\Ct}\left(\hat{p}_\tau-{p}_\tau\right)^2\\
  \leq&8H_\sigmoid\bar{\sigmoid^\prime}^2\sum_{\tau\in\Ct}\epsilon^2+2H_\sigmoid\sum_{\tau\in\Ct}\left(\hat{p}_\tau-{p}_\tau\right)^2\\
  \leq&8H_\sigmoid\bar{\sigmoid^\prime}^2\epsilon^2\aCt+\sqrt{2{\aCt B_p^2\log\frac{\pi^2t^2\mathcal{N}(\mathcal{B}_g,\epsilon, \|\cdot\|_\infty)}{6\delta}}{}}+2\left(
 \log{\Prob_{{g}}}((x_\tau,\onetau)_{\tau\in\Ct})-\log{\Prob_{\hat{g}}}((x_\tau,\onetau)_{\tau\in\Ct})\right)\\
 \leq&C(\epsilon, \delta, \aCt, t)+2\left(\log{\Prob_{\hat{g}_{t+1}^\mathrm{MLE}}}((x_\tau,\onetau)_{\tau\in\Ct})-\log{\Prob_{\hat{g}_{t+1}}}((x_\tau,\onetau)_{\tau\in\Ct})\right)\\
 &+2\left(\log{\Prob_{\hat{g}_{t+1}}}((x_\tau, \onetau)_{\tau\in\Ct})-\log{\Prob_{\hat{g}}}((x_\tau, \onetau)_{\tau\in\Ct})\right)\\
 \leq&C(\epsilon, \delta, \aCt, t)+4\epsilon t+2\betag_1(\epsilon, \delta,\aCt,t)\\
 =&\betag_2(\epsilon,\delta,\aCt, t)+2\betag_1(\epsilon, \delta,\aCt, t),
\end{align*}
where $C(\epsilon, \delta,\aCt, t)=8H_\sigmoid\bar{\sigmoid^\prime}^2\epsilon^2t+\sqrt{2{\aCt B_p^2\log\frac{\pi^2t^2\mathcal{N}(\mathcal{B}_g,\epsilon, \|\cdot\|_\infty)}{6\delta}}{}}$ and $\betag_2(\epsilon, \delta, \aCt, t)=C(\epsilon, \delta,\aCt, t)+4\epsilon t$. 

Furthermore,
\begin{align*}
  \sum_{\tau=1}^t\left(\hat{p}^{t+1}_\tau-p_\tau\right)^2\geq\sum_{\tau=1}^t\lb{\underline{\sigmoid}^\prime}\rb^2\left(\hat{g}^{t+1}(x_\tau)-{g}(x_\tau)\right)^2. 
\end{align*}
The conclusion then follows.

\end{proof}

\subsection{Bound Point-Wise Error}

\begin{lemma}[\textbf{Point-wise Error Bound}]
\label{lem:bound_pointwise_error} 
 For any estimate $\tilde{g}\in\mathcal{B}_g^{t+1}$ measurable with respect to $\mathcal{F}_{t}$, we have, with probability at least $1-{\delta}$, $\forall t\geq1, x\in\mathcal{X}$,  
    \begin{align}
     \big|{\tilde{g}(x)-\tg(x)}\big|
     \leq\;\;2\left(2B_\gl+\regvar^{-\nicefrac{1}{2}}\sqrt{\betag(\epsilon, \nicefrac{\delta}{2}, \aCt, t)}\right)\sigma_{\gtp}(x). 
    \end{align}    
where $\sigma_{\gtp}(x)=\sqrt{k_\gl(x,x) -k_{\gl}(X_{\Ct}, x)^\top(K_{\Ct}+\regvar I)^{-1}k_{\gl}(X_{\Ct}, x)}$.
\end{lemma}

    

\begin{proof}
    We use $\phi(x)$ to denote the function $k_\gl(x,\cdot)$, where $\phi: \mathbb{R}^d\to \mathcal{H}_{k_\gl}$ maps a finite dimensional point $x\in\mathbb{R}^d$ to the RKHS $\mathcal{H}_{k_\gl}$. For notation simplicity, we set $k(\cdot, \cdot)=k_\gl(\cdot, \cdot)$ in this proof. For simplicity, we use $h_1^\top h_2$ to denote the inner product of two functions $h_1, h_2$ from the RKHS $\Hil_{k_\gl}$. Therefore, $h(x)=\langle h, k(x, \cdot)\rangle_{k_\gl}=h^\top\phi(x)$ and
$k_\gl(x, x^\prime)=\langle k_\gl(x, \cdot), k_\gl(x^\prime, \cdot)\rangle=\phi(x)^\top\phi(x^\prime)$, $\forall x, x^\prime\in \mathcal{X}$. We can introduce the feature map $$\Phi_t \defeq\left[\phi(x_\tau)^\top\right]^\top_{\tau\in\Ct},$$ 
we then get the kernel matrix $K_t = \Phi_t\Phi_t^\top$, $k_t(x)= \Phi_t\phi(x)$ for all $x\in\mathcal{X}$ and $h_{\Ct}=\Phi_th$.

Note that when the Hilbert space $\Hil_{k_\gl}$ is a finite-dimensional Euclidean space, $\Phi_t$ is interpreted as the normal finite-dimensional matrix. In the more general setting where $\Hil_{k_\gl}$ can be an infinite-dimensional space, $\Phi_t$ is the evaluation operator $\Hil_{k_\gl}\to\mathbb{R}^{\aCt}$ defined as $\Phi_t h=[h(x_\tau)]^\top_{\tau\in\Ct},\forall h\in\Hil$, with $\Phi_t^\top$ as its adjoint operator.

Since the matrices $(\Phi_t^\top\Phi_t + \regvar I):\mathcal{H}_{k_\gl}\to\mathcal{H}_{k_\gl}$ and $(\Phi_t\Phi_t^\top + \regvar I):\mathbb{R}^{\aCt}\to\mathbb{R}^{\aCt}$ are strictly positive definite and $$(\Phi_t^\top\Phi_t + \regvar I)\Phi_t^\top = \Phi_t^\top(\Phi_t\Phi_t^\top + \regvar I),$$ we have
\begin{equation}
    \Phi_t^\top(\Phi_t\Phi_t^\top + \regvar I)^{-1} = (\Phi_t^\top\Phi_t + \regvar I)^{-1}\Phi_t^\top.
\label{eqn:dim-change}
\end{equation}
Also from the definitions above $(\Phi_t^\top\Phi_t+\regvar I)\phi(x)=\Phi_t^\top k_t(x) + \regvar \phi(x)$, and thus from Eq.~\eqref{eqn:dim-change} we deduce that
\begin{equation}
    \phi(x)=\Phi_t^\top(\Phi_t\Phi_t^T+\regvar I)^{-1}k_t(x)+\regvar (\Phi_t^\top\Phi_t + \regvar I)^{-1}\phi(x),
\end{equation}
which gives
\begin{equation}
    \phi(x)^\top\phi(x)= k_t(x)^\top(\Phi_t\Phi_t^\top+\regvar I)^{-1}k_t(x)+\regvar \phi(x)^\top(\Phi_t^\top\Phi_t + \regvar I)^{-1}\phi(x).
\end{equation}
This implies
\begin{equation}
    \regvar \phi(x)^\top(\Phi_t^\top\Phi_t + \regvar I)^{-1}\phi(x) = k(x,x) -k_t(x)^\top(K_t+\regvar I)^{-1}k_t(x),
\label{eqn:variance}
\end{equation}
which is by definition the posterior variance $(\sigma_{\gtp}(x))^2$.
Now we can observe that
\begin{align*} 
&\abs{g(x)-k_t(x)^\top(K_t+\regvar I)^{-1}g_{\Ct}} \\
=& \abs{\phi(x)^\top g- \phi(x)^\top\Phi_t^\top(\Phi_t\Phi_t^\top+\regvar I)^{-1}\Phi_t g}\\
=& \abs{\phi(x)^\top g-\phi(x)^\top(\Phi_t^\top\Phi_t + \regvar I)^{-1}\Phi_t^\top\Phi_t g}\\
=& \abs{\phi(x)^\top(\Phi_t^\top\Phi_t + \regvar I)^{-1}(\Phi_t^\top\Phi_t+\regvar I) g-\phi(x)^\top(\Phi_t^\top\Phi_t + \regvar I)^{-1}\Phi_t^\top\Phi_t g}\\
=& \abs{\regvar \phi(x)^\top(\Phi_t^\top\Phi_t+\regvar I)^{-1}g}\\
\le& \norm{\regvar(\Phi_t^\top\Phi_t+\regvar I)^{-1}\phi(x)}_{k_\gl}\norm{g}_{k_\gl}\\
=& \norm{g}_{k_\gl} \sqrt{\regvar \phi(x)^\top(\Phi_t^\top\Phi_t+\regvar I)^{-1}\regvar I(\Phi_t^\top\Phi_t+ \regvar I)^{-1}\phi(x)}\\
\le & B_\gl \sqrt{\regvar \phi(x)^\top(\Phi_t^\top\Phi_t+\regvar I)^{-1}(\Phi_t^\top\Phi_t+\regvar I)(\Phi_t^\top\Phi_t+\regvar I)^{-1}\phi(x)}\\
=&B_\gl\; \sigma_{\gtp}(x), 
\end{align*}
where the second equality uses Eq.~\eqref{eqn:dim-change}, the first inequality is by Cauchy-Schwartz and the final equality is from Eq.~\eqref{eqn:variance}. 
We define $\epsilon_{\Ct}=\tilde{g}_{\Ct}-{g}_{\Ct}$, where $\tilde{g}_\tau=\tilde{g}(x_\tau)$. We have,
\begin{align*}
   & |{k_t(x)^\top(K_t+\regvar I)^{-1}\epsilon_{\Ct}}|\\ =&\abs{\phi(x)^\top\Phi_t^\top(\Phi_t\Phi_t^\top + \regvar I)^{-1}\epsilon_{\Ct}}\\
=& \abs{\phi(x)^\top(\Phi_t^\top\Phi_t + \regvar I)^{-1}\Phi_t^\top\epsilon_{\Ct}} \\
\le &\norm{(\Phi_t^\top\Phi_t + \regvar I)^{-1/2}\phi(x)}_{k_\gl} \norm{(\Phi_t^\top\Phi_t + \regvar I)^{-1/2}\Phi_t^\top\epsilon_{\Ct}}_{k_\gl}\\
= &\sqrt{\phi(x)^\top(\Phi_t^\top\Phi_t + \regvar I)^{-1}\phi(x)}\sqrt{(\Phi_t^\top\epsilon_{\Ct})^\top(\Phi_t^\top\Phi_t + \regvar I)^{-1}\Phi_t^\top\epsilon_{\Ct}}\\
=&\regvar^{-1/2}\sigma_{\gtp}(x)\sqrt{\epsilon_{\Ct}^\top\Phi_t\Phi_t^\top(\Phi_t\Phi_t^\top +\regvar I)^{-1}\epsilon_{\Ct}}\\
=&\regvar^{-1/2}\sigma_{\gtp}(x)\sqrt{\epsilon_{\Ct}^\top K_t(K_t+\regvar I)^{-1}\epsilon_{\Ct}}\\
\leq&\regvar^{-1/2}\sigma_{\gtp}(x)\sqrt{\epsilon_{\Ct}^\top\epsilon_{\Ct}}\\
\leq&\regvar^{-1/2}\alpha_t^{1/2}\sigma_{\gtp}(x),
\end{align*}
where the second equality is from Eq.~\eqref{eqn:dim-change}, the first inequality is by Cauchy-Schwartz and the last inequality follows by Eq.~\eqref{eq:bound_over_hist} and $\alpha_t=\betag(\epsilon, \nicefrac{\delta}{2}, \aCt, t)$.
\begin{align*}
         &|\fg(x)-g(x)| \\
     \le & \left|\left({k_t(x)^\top(K_t+\regvar I)^{-1}(\tilde{g}_{\Ct}-g_{\Ct})}\right) - \left({g(x)-k_t(x)^\top(K_t+\regvar I)^{-1}g_{\Ct}}\right) + \left({\fg(x)-k_t(x)^\top(K_t+\regvar I)^{-1}\fg_{\Ct}}\right)\right|\\
    \le & |{k_t(x)^\top(K_t+\regvar I)^{-1}(\tilde{g}_{\Ct}-g_{\Ct})}| + |{g(x)-k_t(x)^\top(K_t+\regvar I)^{-1}g_{\Ct}}| + |{\fg(x)-k_t(x)^\top(K_t+\regvar I)^{-1}\fg_{\Ct}}|\\
    \le& \sigma_{\gtp}(x)\Big(2B_\gl + \regvar^{-1/2}\betag_t^{1/2}\Big).
\end{align*}
\end{proof}

\subsection{Efficient Computations of Confidence Range for the Latent \rev{\Gcost} function $g$}
\label{app_sec:comp_conf_g}
Leveraging the representer theorem \cite{scholkopf2001generalized, xu2024principled} thanks to the RKHS property, the MLE problem and confidence range computation problem can be reduced to an $\mathcal{O}(|\Ct|)$-dimensional, tractable optimisation problem~\eqref{eqn:MLE}, problem~\eqref{eqn:upper_g} and problem~\eqref{eqn:lower_g}.

\begin{equation}
\begin{aligned}
    \label{eqn:MLE}
\ell_t(\hat{g}_t^\mathrm{MLE})=\min_{Z_{\Ct}\in\mathbb{R}^{|\Ct|}}&\quad\sum_{\tau\in\Ct} Z_\tau\one_\tau-\sum_{\tau\in\Ct}\log\left(1+e^{Z_\tau}\right)\\
   \text{subject to}&\quad 
Z_{\Ct}K_{\Ct}^{-1}Z_{\Ct}    \leq B_g^2, 
\end{aligned}
\end{equation}
where $K_{\Ct} := (k_g(x_{\tau_1}, x_{\tau_2}))_{\tau_1, \tau_2\in\Ct}$.

\begin{equation}
\begin{aligned}
    \label{eqn:upper_g}
\bar{g}_t(x)=\max_{Z_{\Ct}\in\mathbb{R}^{|\Ct|}, \,\, z\in\mathbb{R}, \,\, x\in\mathcal{X}}&\quad z\\
   \text{subject to}&\quad \left[\begin{array}{l}
Z_{\Ct} \\
z
\end{array}\right]^{\top}K_{\Ct,x}^{-1}\left[\begin{array}{l}
Z_{\Ct} \\
z
\end{array}\right]   \leq B_g^2, \\
   &\quad \ell(Z_{\Ct} \mid \mathcal{D}^g_t)\geq \ell_t(\hat{g}^\mathrm{MLE}_t)-\betag_1(\epsilon, \delta, \aCt, t),
\end{aligned}
\end{equation}
where $K_{\Ct,x} := \left(k_g(\tilde{x}, \tilde{x}^\prime)\right)_{\tilde{x},\tilde{x}^\prime \in X_{\Ct}\cup\{x\}}$, and $\ell(Z_{\Ct} \mid \mathcal{D}^g_t)=\sum_{\tau\in\Ct} Z_\tau\one_\tau-\sum_{\tau\in\Ct}\log\left(1+e^{Z_\tau}\right)$ is the LL value when the function value at $x_\tau$ is $Z_\tau$.

\begin{equation}
\begin{aligned}
    \label{eqn:lower_g}
\underline{g}_t(x)=\min_{Z_{\Ct}\in\mathbb{R}^{|\Ct|}, \,\, z\in\mathbb{R}, \,\, x\in\mathcal{X}}&\quad z\\
   \text{subject to}&\quad \left[\begin{array}{l}
Z_{\Ct} \\
z
\end{array}\right]^{\top}K_{\Ct,x}^{-1}\left[\begin{array}{l}
Z_{\Ct} \\
z
\end{array}\right]   \leq B_g^2, \\
   &\quad \ell(Z_{\Ct} \mid \mathcal{D}^g_t)\geq \ell_t(\hat{g}^\mathrm{MLE}_t)-\betag_1(\epsilon, \delta, \aCt, t),
\end{aligned}
\end{equation}
where $K_{\Ct,x} := \left(k_g(\tilde{x}, \tilde{x}^\prime)\right)_{\tilde{x},\tilde{x}^\prime \in X_{\Ct}\cup\{x\}}$, and $\ell(Z_{\Ct} \mid \mathcal{D}^g_t)=\sum_{\tau\in\Ct} Z_\tau\one_\tau-\sum_{\tau\in\Ct}\log\left(1+e^{Z_\tau}\right)$ is the LL value when the function value at $x_\tau$ is $Z_\tau$.

\subsection{Bound Cumulative Standard Deviation over Sample Trajectory}

\begin{lemma}[Lemma 4,~\cite{chowdhury2017kernelized}\footnote{Appears in the arXiv version:~\url{https://arxiv.org/pdf/1704.00445}.}]
\begin{equation}
    \label{lem:bound_cumu_sd_f}
\sum_{t\in\GT} {\sigma}_{\ft}\left(x_t\right) \leq \sqrt{4(|\GT|+2) {\gamma}^{f}_{|\GT|}}=\mathcal{O}\left(\sqrt{|\GT| {\gamma}^{f}_{|\GT|}}\right).
\end{equation}
\end{lemma} 

Similarly, we have,
\begin{equation}
    \label{lem:bound_cumu_sd_g}
\sum_{t\in\CT} {\sigma}_{\gt}\left(x_t\right) \leq \sqrt{4(|\CT|+2) {\gamma}^{g}_{|\CT|}}=\mathcal{O}\left(\sqrt{|\CT| {\gamma}^{g}_{|\CT|}}\right).
\end{equation}

\subsection{Bound Cumulative Regret}
We can then analyze the regret of our algorithm. We use $\CoT$ to denote the set $\lig t\in[T]| x_t=x_t^c\rig$. 
\begin{align*}
 R_{\GT}=& \sum_{t\in\GT}[f(x_t)-f(x^\star)] \\
 =& \sum_{t\in\GT\cap\mathcal{C}_T}[f(x_t)-f(x^\star)]+\sum_{t\in\GT\setminus\mathcal{C}_T}[f(x_t)-f(x^\star)] \\
 =& \sum_{t\in\GT\cap\mathcal{C}_T}[f(x^c_t)-f(x^\star)]+\sum_{t\in\GT\setminus\mathcal{C}_T}[f(x^u_t)-f(x^\star)] \\ 
\end{align*} 

For the first part, we have, 
\begin{align*}
&\sum_{t\in\GT\cap\mathcal{C}_T}[f(x^c_t)-f(x^\star)]\\
 = &\sum_{t\in\GT\cap\mathcal{C}_T}[f(x^c_t)-\underline{f}_t(x^c_t)+\underline{f}_t(x^c_t)-f(x^\star)]\\
 \leq &\sum_{t\in \GT\cap\mathcal{C}_T}[f(x^c_t)-\underline{f}_t(x^c_t)+\underline{f}_t(x^c_t)-\underline{f}_t(x^\star)]\\
= &\sum_{t\in \GT\cap\mathcal{C}_T}[f(x^c_t)-\underline{f}_t(x^c_t)+\underline{f}_t(x^c_t)-\underline{f}_t(x^u_t)+\underline{f}_t(x^u_t)-\underline{f}_t(x^\star)]\\
\leq & \sum_{t\in\GT\cap\mathcal{C}_T}2\beta_{\ft}\sigma_{\ft}(x_t)+ \sum_{t\in \GT\cap\mathcal{C}_T}[\underline{f}_t(x^c_t)-\underline{f}_t(x^u_t)]\\
\leq & 2\beta_{\fT} \sum_{t\in\GT\cap\mathcal{C}_T} \sigma_{\ft}(x_t)+\sum_{t\in \GT\cap\mathcal{C}_T}[\underline{f}_t(x^c_t)-\underline{f}_t(x^u_t)], 
\end{align*}
where $\sigma_{\ft}$ is as given in Eq.~\eqref{eq:f_sigma_def}, the first inequality follows by Lem.~\ref{lem:conf_int}, the second inequality follows by Lem.~\ref{lem:conf_int} and the line~\ref{alg_line:uncon_opt_lcb} of Alg.~\ref{alg:HAIBO}.   

Furthermore, we have,
\begin{align}
 &\sum_{t\in \GT\cap\mathcal{C}_T}[\underline{f}_t(x^c_t)-\underline{f}_t(x^u_t)]\\
\leq& \sum_{t\in \GT\cap\mathcal{C}_T}[\bar{f}_t(x^u_t)-\underline{f}_t(x^u_t)]\\
\leq & \sum_{t\in \GT\cap\mathcal{C}_T}2\beta_{\ft}\sigma_{\ft}(x^u_t)\\
\leq & \sum_{t\in \GT\cap\mathcal{C}_T}2\beta_{\ft}\eta\sigma_{\ft}(x^c_t)\\
= & \sum_{t\in \GT\cap\mathcal{C}_T}2\beta_{\ft}\eta\sigma_{\ft}(x_t)
\end{align}
where the first inequality follows by the condition in line~\ref{alg_line:constr_cond} of the Alg.~\ref{alg:HAIBO}, the second inequality follows by the Lem.~\ref{lem:conf_int}, and the third inequality follows by the condition in line~\ref{alg_line:constr_cond} of the Alg.~\ref{alg:HAIBO}.

For the second part, we have,
\begin{align}
&\sum_{t\in\GT\setminus\mathcal{C}_T}[f(x^u_t)-f(x^\star)]\\
 = &\sum_{t\in\GT\setminus\mathcal{C}_T}[f(x^u_t)-\underline{f}_t(x^u_t)+\underline{f}_t(x^u_t)-f(x^\star)]\\
 \leq &\sum_{t\in \GT\setminus\mathcal{C}_T}[f(x^u_t)-\underline{f}_t(x^u_t)+\underline{f}_t(x^u_t)-\underline{f}_t(x^\star)]\\
\leq & \sum_{t\in\GT\setminus\CoT}2\beta_{\ft}\sigma_{\ft}(x_t)\\
\leq & 2\beta_{\fT} \sum_{t\in\GT\setminus\CoT} \sigma_{\ft}(x_t), \end{align}
where the first inequality follows by that $f(x^\star)\geq\underline{f}_t(x^\star)$, the second inequality follows by the optimality of $x_t^u$ for the problem in line~\ref{alg_line:uncon_opt_lcb} and the Lem.~\ref{lem:conf_int}, and the third inequality follows by the monotonicity of $\beta_{\ft}$ in $t$. 

Hence, 
\begin{align*}
 R_{\GT}\leq &2(2+\eta)\beta_{\fT} \sum_{t\in\GT} \sigma_{\ft}(x_t)\\  
 \leq&   2(2+\eta)\beta_{\fT} \sqrt{4(\aGT+2) \gamma^f_{\aGT}}\\
= &  \mathcal{O}\left(\gamma^f_{\aGT}\sqrt{\aGT}\right).
\end{align*}

\subsection{Bound Cumulative Queries to Labeler}
We can then analyze the cumulative queries to the expert. 
We notice that, $\forall t\in\CT$,

\begin{equation}
   \bar{g}_t(x_t)-\underline{g}_t(x_t)\geq \gthr 
\end{equation}

Meanwhile, by Lem.~\ref{lem:bound_pointwise_error}, 
\begin{equation}
   \bar{g}_t(x_t)-\underline{g}_t(x_t)\leq 4\left(2B_\gl+\regvar^{-\nicefrac{1}{2}}\sqrt{\betag_t}\right)\sigma_{\gt}(x).
\end{equation}

Hence,
\begin{equation}
   \gthr\leq 4\left(2B_\gl+\regvar^{-\nicefrac{1}{2}}\sqrt{\betag_t}\right)\sigma_{\gt}(x).
\end{equation}

Therefore,
\begin{align}
 \CTc= &|\CT|\\
 =&\sum_{t\in\CT} 1\\
 \leq&\frac{1}{\gthr}\sum_{t\in\CT} \gthr\\
 \leq&  \frac{1}{\gthr}\sum_{t\in\CT} 4\left(2B_\gl+\regvar^{-\nicefrac{1}{2}}\sqrt{\betag_t}\right)\sigma_{\gt}(x_t)\\
\leq&  \frac{4}{\gthr}\left(2B_\gl+\regvar^{-\nicefrac{1}{2}}\sqrt{\betag_T}\right)\sum_{t\in\CT} \sigma_{\gt}(x_t)\\
= & \mathcal{O}\left(\sqrt{\betag_T\gamma^{g}_{\aCT} \aCT}\right).
\end{align}

Dividing by $\sqrt{\aCT}$, we obtain, 
\begin{equation}
    \sqrt{|\CT|}=\mathcal{O}(\sqrt{\betag_T\gamma^{g}_{\aCT}}).
\end{equation}

Hence,
\begin{align}
\CTc= |\CT|=\mathcal{O}({\betag_T\gamma^{g}_{\aCT}}).
\label{eq:divide_1}
\end{align}

By setting $\epsilon=\frac{1}{T}$, we have
\begin{align}
\betag_T=\mathcal{O}\left(\sqrt{\aCT\log\frac{T\mathcal{N}(\mathcal{B}_g, \nicefrac{1}{T},\|\cdot\|_\infty)}{\delta}}\right).
\end{align}

Hence, dividing by $\sqrt{\aCT}$ on Eq.~\eqref{eq:divide_1} again, we obtain,
\begin{align}
\CTc= |\CT|=\mathcal{O}\left(\left(\gamma^{g}_{\aCT}\right)^2{\log\frac{T\mathcal{N}(\mathcal{B}_g, \nicefrac{1}{T},\|\cdot\|_\infty)}{\delta}}\right)\leq \mathcal{O}\left(\left(\gamma^{g}_{T}\right)^2{\log\frac{T\mathcal{N}(\mathcal{B}_g, \nicefrac{1}{T},\|\cdot\|_\infty)}{\delta}}\right).
\label{eq:divide_2}
\end{align}

\rev{
\section{Detailed Discussions on The Significance of Thm~\ref{thm:RQ_bound}}
\label{app_sec:det_thm}
\textbf{Order-wise improvement can not be attained under current mild assumption}. $g$ may contain no information (e.g., $g=0$) or even adversarial. Even if human expertise is helpful, we can not guarantee an \emph{order-wise} improvement either. For example, consider the following $g$,
\[
g(x)=
\begin{cases}
 f(x^\star)+c, & \text{if } f(x)-f(x^\star)\leq c, \\
 f(x) & \text{otherwise}, 
\end{cases}
\]
where $c>0$ is a positive constant. In practice, such a scenario means the human expert has some rough idea in a near-optimal region but not exactly sure where the exact optimum is. This is common in practice. In this case, human expert is helpful in identifying the region with $f(x)\leq f(x^\star)+c$ but no longer helpful for further optimization inside the region $\{x\in\mathcal{X}|f(x)\leq f(x^\star)+c\}$. However, convergence rate is defined in the asymptotic sense. Hence, an order-wise improvement can not be guaranteed.

\textbf{Assumption becomes unrealistic if we {really} want it}. 
Some papers that show theoretical superiority [2, 6], yet the assumptions are unrealistic. For example, [6] assumed that the human knows the true kernel hyperparameters while GP is misspecified, and [2] assumed the human belief function $g$ has better and tighter confidence intervals over the entire domain. We can derive the better convergence rate of our algorithm than AI-only ones if we use [2] assumption, but this is unlikely to be true in reality. In fact, our method outperforms these method empirically (see Figure 5). This supports the superiority based on unrealistic conditions is not meaningful in practice. 

\textbf{Empirical success can be achieved without order-wise improvement on worst-case convergence.}
Our assumption is more natural; following [37], we posit humans have better prior knowledge than GP and are only useful at the beginning as a warm starter. This assumption is widely accepted by the community and practitioners, which leads to real-world impact (e.g. Nature [42]). The warm-starting-based papers [36, 37, 44] have been published in reputable venues without such a theory. In our manuscript, real-world applications also empirically demonstrate that our method not only improves the convergence of BO, but also maintains robustness despite varying labelling accuracy.  

\textbf{Worst-case convergence and hand-over guarantees matter.}
We believe that the value of theory is the worst-case guarantee. To be clear, starting point of human-AI collaborative BO is that \emph{the experts are not currently using BO}. The scientific experts do very expensive tasks, which often cost millions of dollars and weeks to months to test one design (e.g. battery design). They are reluctant to employ BO due to its opaque and untrustworthy nature. The experts want to be involved in the AI decision-making process, otherwise they are forced to work as a robot feeding experimental results to the AI. But, they are also in the middle of trial and error, so their advice is not always reliable. Our worst-case guarantee assures that at least their involvement does not harm the AI-only results, and also assures the automation in the later round. Thus, we believe our approach can extend the applicable range of BO to high-stakes optimisation tasks. Furthermore, our handover guarantee assures that only limited human labeling effort is needed, which is also meaningful because the motivation to use BO is to alleviate the tedious human effort in the first place.  
}

\rev{

\section{Proof of the Kernel-Specific Bounds in Tab.~\ref{tab:kern_spec_bounds}}
For the cumulative regret part, we have,
\begin{itemize}
\item If the kernel function is linear, $\gamma^f_{\aGT}=\mathcal{O}(\log \aGT)$, and thus $R_{\aGT}=\mathcal{O}\left(\sqrt{\aGT}\log \aGT\right)$.
\item If the kernel function is squared exponential, $\gamma^f_{|\GT|}=\mathcal{O}((\log\aGT)^{d+1})$, $R_{\GT}=\mathcal{O}(\sqrt{\aGT}(\log \aGT)^{d+1})$.
\item If the kernel function is {M\'atern}, $\gamma^f_{\aGT}=\mathcal{O}\left(\aGT^{\frac{d }{2 \nu+d}}\log^{\frac{2\nu}{2\nu+d}}(\aGT)\right) (\left(\nu>\frac{d}{2}\right))$, $R_{\GT}=\mathcal{O}\left(\aGT^{\frac{2\nu+3d}{4\nu+2d}}\log^{\frac{2\nu}{2\nu+d}}(\aGT)\right)$. 
\end{itemize}

To bound the cumulative queries, we have, 
\begin{enumerate}
    \item $k_\gl$ is a linear kernel, then $\log\mathcal{N}(\mathcal{B}_g, T^{-1},\|\cdot\|_\infty)=\mathcal{O}\left(\log\frac{1}{\epsilon}\right)=\mathcal{O}\left(\log T\right)$. By Thm. 5 in~\cite{srinivas2012information},
    $$
    \gamma_T^{g}=\mathcal{O}(\log T). 
    $$
    Hence, 
    $$
    \CTc=\mathcal{O}\left((\log T)^2\log T\right)=\mathcal{O}\left((\log T)^3\right). 
    $$

 \item $k_\gl$ is a squared exponential kernel, then $\log\mathcal{N}(\mathcal{B}_g, T^{-1},\|\cdot\|_\infty)=\mathcal{O}\left((\log\frac{1}{\epsilon})^{d+1}\right)=\mathcal{O}\left((\log T)^{d+1}\right)$~(Example 4,~\cite{zhou2002covering}). By Thm. 4 in~\cite{kandasamy2015high}, we have,
    $$
    \gamma_T^{g}=\mathcal{O}((\log T)^{d+1}). 
    $$
   Hence, 
    $$
    \CTc=\mathcal{O}\left((\log T)^{2(d+1)}(\log T)^{d+1}\right)=\mathcal{O}\left((\log T)^{3(d+1)}\right). 
    $$

 \item $k_\gl$ is a Matern kernel, then $\log\mathcal{N}(\mathcal{B}_g, T^{-1},\|\cdot\|_\infty)=\mathcal{O}\left((\frac{1}{\epsilon})^{\nicefrac{d}{\nu}}\log\frac{1}{\epsilon}\right)=\mathcal{O}\left(T^{\nicefrac{d}{\nu}}\log T\right)$~(by Thm. 5.1 and Thm. 5.3 in \cite{xu2024lower}). By Thm. 4 in~\cite{kandasamy2015high}, we have,
    $$
    \gamma_T^{g}=\mathcal{O}\left(T^{\frac{d(d+1)}{2\nu+d(d+1)}}\log T\right). 
    $$
    Hence, 
    $$
    \CTc=\mathcal{O}\left(T^{\frac{2d(d+1)}{2\nu+d(d+1)}}(\log T)^2T^{\frac{d}{\nu}}\log T\right)=\mathcal{O}(T^{\frac{2d(d+1)}{2\nu+d(d+1)}}T^{\frac{d}{\nu}}(\log T)^3), 
    $$
    where $\nu>\frac{d\lb d+3+\sqrt{d^2+14d+17}\rb}{4}$.
\end{enumerate}

\section{Theoretical improvement of convergence rate}\label{app:improvement}
\begin{figure}[ht]
\begin{algorithm}[H]
\caption{\textbf{CO}llaborative \textbf{B}ayesian \textbf{O}ptimization with \textbf{H}elpful \textbf{L}abelling Experts (\textbf{COBOHL}).}
\label{alg:cobohl}
\begin{algorithmic}[1]
\normalsize
\State \rev{\textbf{Input and Initialization}: function space ball $\mathcal{B}_g$, and uncertainty threshold $\gthr$. }
\State Set $\mathcal{B}^1_g=\mathcal{B}_g$\rev{, $\mathcal{Q}^f_0=\emptyset$, and $\mathcal{Q}^g_0=\emptyset$}. 
\For{$t\in[T]$} 

\State Generate $x_t$ by solving the constrained auxiliary optimization problem $\min_{x\in\mathcal{X}}\underline{f}_t(x)\text{ subject to } \underline{g}_t(x)\leq0$. \Comment{Expert-constrained LCB}
\label{alg_line:colhbo_generate_xc}
    
\If{$\bar{g}_t(x_t)-\underline{g}_t(x_t)>\gthr$} \Comment{Handover guarantee}
 \State  Query the expert's label to get the feedback $\mathbf{1}_t$.
 \State Update $\Ct=\Ctm\cup\{t\}$ and the posterior confidence set $\mathcal{B}^{t+1}_g$. Set $\Gt=\Gtm$.
\Else  
\State Evaluate the black-box function at the point $x_t$, and set $\Gt=\Gtm\cup\{t\}$. Set $\Ct=\Ctm$.
 \State Update the posterior mean/variance of the objective $f$.
\EndIf
 
\EndFor
\end{algorithmic}
\end{algorithm}
\vspace{-2em}
\end{figure}

Here, we give the analysis on the regret of COBOHL,  
\begin{align}
    \sum_{t\in\GT} (f(x_t)-f(x^\star))&= \sum_{t\in\GT}(f(x_t)-\underline{f}_t(x_t)+\underline{f}_t(x_t)-f(x^\star)) \\
  & \leq \sum_{t\in\GT}(f(x_t)-\underline{f}_t(x_t)) \\
  & \leq \sum_{t\in\GT}2\beta_{\ft}\sigma_{\ft}(x_t)\\
  & \leq 2\beta_{\fT}\sum_{t\in\GT}\sigma_{\ft}(x_t)\\
  & = \mathcal{O}\left(\gamma^{f, \mathcal{X}^g}_{\aGT}\sqrt{\aGT}\right),
\end{align}
where the maximum information gain is defined over the set $\mathcal{X}^g\defeq\{x\in\mathcal{X}|g(x)\leq\gthr\}$. Meanwhile, the regret bound of vanilla LCB has a similar form of $\mathcal{O}\left(\gamma^{f, \mathcal{X}}_{\aGT}\sqrt{\aGT}\right)$. Notably, the regret bound for vanilla LCB has a maximum information gain defined over the region $\mathcal{X}$. For commonly used kernel functions, the maximum information gain is proportional to the volume of the set. Since $\mathcal{X}^g\subset\mathcal{X}$, $\mathrm{vol}(\mathcal{X}^g)\leq\mathrm{vol}(\mathcal{X})$ and the maximum information gain gets reduced by a ratio of $\frac{\mathrm{vol}(\mathcal{X}^g)}{\mathrm{vol}(\mathcal{X})}$. Therefore, the regret bound gets improved by a ratio of $\frac{\mathrm{vol}(\mathcal{X}^g)}{\mathrm{vol}(\mathcal{X})}$.

}

{
\section{Estimating norm bound online}
\label{sec:est_norm_bound}
By Assumption~\ref{assump:bounded_norm}, there exists a large enough constant $B_\gl$ that upper bounds the norm of the ground-truth latent black-box \rev{function} $g$. However, a tight estimate of this upper bound may be unknown to us in practice, while the execution of our algorithm explicitly relies on knowing a bound $B_\gl$ (in Prob.~\eqref{eqn:reform_inner_prob_to_fin}, $B_\gl$ is a key parameter). 

So it is necessary to estimate the norm bound $B_\gl$ using the online data. Suppose our guess is $\hat{B}$. It is possible that $\hat{B}$ is even smaller than the ground-truth function norm $\|g\|$. To detect this underestimate, we observe that, with the correct setting of $B_\gl$ such that $B_\gl\geq\|g\|$, we have that by Lemma~\ref{thm:conf_set_g} and the definition of maximum likelihood estimate,
$$
\ell_t(\hat{g}^{\mathrm{MLE}}_{t|\hat{B}})\geq\ell_t(g)\geq\ell_t(\hat{g}^{\mathrm{MLE}}_{t|B})-\betag_1(\epsilon, \delta, \aCt, t|\hat{B}),
$$
where $\hat{g}^{\mathrm{MLE}}_{t|\hat{B}}$ is the maximum likelihood estimate function with function norm bound $\hat{B}$ and  $\betag_1(\epsilon, \delta,\aCt, t|\hat{B})$ is the corresponding parameter as defined in Lemma~\ref{thm:conf_set_g} with norm bound $\hat{B}$. We also have $2\hat{B}$ is a valid upper bound on $\|g\|$ and thus,     
$$
\ell_t(\hat{g}^{\mathrm{MLE}}_{t|2\hat{B}})\geq\ell_t(g)\geq\ell_t(\hat{g}^{\mathrm{MLE}}_{t|2\hat{B}})-\betag_1(\epsilon, \delta,\aCt, t|2\hat{B}).
$$
Therefore, 
$$
\ell_t(\hat{g}^{\mathrm{MLE}}_{t|\hat{B}})\geq\ell_t(g)\geq\ell_t(\hat{g}^{\mathrm{MLE}}_{t|2\hat{B}})-\betag_1(\epsilon, \delta,\aCt, t|2\hat{B}).
$$
That is to say, $\ell_t(\hat{g}^{\mathrm{MLE}}_{t|\hat{B}})$ needs to be greater than or equal to $\ell_t(\hat{g}^{\mathrm{MLE}}_{t|2\hat{B}})-\betag_1(\epsilon, \delta,\aCt, t|2\hat{B})$ when $\hat{B}$ is a valid upper bound on $\|g\|$. 

Therefore, we can use the heuristic: every time we find that
$$
\ell_t(\hat{g}^{\mathrm{MLE}}_{t|\hat{B}})<\ell_t(\hat{g}^{\mathrm{MLE}}_{t|2\hat{B}})-\betag_1(\epsilon, \delta, \aCt, t|2\hat{B}),
$$
we double the upper bound guess $\hat{B}$.
}

\section{Related Work}\label{app:related}
\begin{table}
    \centering
    \caption{Comparisons between our algorithm with the existing baseline methods.}
    \label{tab:comparison}
    \resizebox{1\textwidth}{!}{
    \begin{tabular}{lccccccc}
    \toprule
         baselines &
         \begin{tabular}{@{}l@{}}blackbox \\ human model?\end{tabular}&
         \begin{tabular}{@{}l@{}}no-rankability\\ assumption?\end{tabular}&
         \begin{tabular}{@{}l@{}}continuous\\ guarantee?\end{tabular}&
         \begin{tabular}{@{}l@{}}no-harm \\ guarantee?\end{tabular}&
         \begin{tabular}{@{}l@{}}data-driven \\ trust?\end{tabular}&
         \begin{tabular}{@{}l@{}}handpver\\ guarantee?\end{tabular}&
         \\
    \midrule
         AV et al. (2022) \cite{av2022human}& \ding{51} & \xmark & \xmark & \xmark & \xmark & \xmark\\
         Hvarfner et al. (2022) \cite{hvarfner2022pi}& \xmark & \xmark&\ding{51} & \ding{51} & \xmark & \xmark\\
         Gupta et al. (2023) \cite{gupta2023bo}& \ding{51} & \xmark  &\ding{51} & \xmark & \xmark & \xmark\\
         Khoshvishkaie et al. (2023) \cite{khoshvishkaie2023cooperative}&\ding{51} & \xmark &\xmark & \xmark & \xmark & \xmark\\
         Cisse et al. (2023) \cite{cisse2023hypbo}& \xmark & \xmark & \xmark & \xmark & \xmark & \xmark\\
         Adachi et al. (2023) \cite{adachi2023looping}& \ding{51} & \xmark  &\xmark & \ding{51} & \xmark & \xmark\\
         Rodemann et al. (2024) \cite{rodemann2024explaining}& \ding{51} & \xmark &\xmark & \xmark & \xmark & \xmark\\
         AV et al. (2024) \cite{av2024enhanced}& \ding{51} & \xmark & \xmark & \xmark & \xmark & \xmark\\
         Hvarfner et al. (2024) \cite{hvarfner2024a}& \xmark & \xmark & \xmark & \xmark & \xmark & \xmark\\
    \midrule
         \textbf{Ours} & \ding{51} & \ding{51} & \ding{51} & \ding{51} & \ding{51} & \ding{51}\\
    \bottomrule
    \end{tabular}
    }
\end{table}
We summarized the baseline comparison in terms of five factors in Table~\ref{tab:comparison}. Our algorithm is the first to offer a data-driven trust level no-harm guarantee and a handover guarantee under no rankability assumption. 

We briefly introduce the baseline methods used in the real-world experiments::
\begin{enumerate}
    \item AV. et al., NeurIPS 2022 \cite{av2022human}: This algorithm initially proposed the human-AI collaborative setting. The approach is straightforward: human experts can intervene in the optimization process if they find the next query location suggested by the vanilla LCB BO to be unpromising. This method can be described as a 'human as constraint' approach, where the BO must adhere to the experts' recommendations regardless of the quality of their advice. This approach assumes that human experts are at least better than the vanilla LCB, thus requiring a high level of trust in the experts. As shown in Figure~\ref{fig:human}, experts' input is not always reliable.
    \item Khoshvishkaie et al., ECML 2023 \cite{khoshvishkaie2023cooperative}: This setting assumes that the querying budget is equally divided between human experts and the vanilla LCB BO. This means that once a point is selected by human experts, the BO will alternately select the next query. This method can select the vanilla LCB regardless of what the human expert selected, making it likely to achieve a no-harm guarantee, although no theoretical proof is provided. The trust level in experts in this method is low, as all expert inputs are treated equally regardless of their quality. Therefore, while this method performs well in unreliable settings, it is not as effective when experts are good advisors. To be fair, their work focuses more on imperfect cases and does not consider scenarios with effective experts.
    \item Adachi et al., AISTATS 2024 \cite{adachi2023looping}: This setting assumes that the BO provides two possible candidates, from which the human selects one. Both candidates have convergence guarantees, thus ensuring a no-harm guarantee, although their proof is limited to discrete settings. However, the human must ultimately choose one of the candidates, maintaining a high level of trust in human experts. They introduced a discounting function that hand-tunes the decaying rate of trust, gradually generating the same candidates. Although their work initiated the no-harm guarantee concept, the trust level adjustment is not data-driven and the proof is limited to discrete cases. To be fair, their main focus is on the explainability of black-box optimizers, which we did not consider in this work. Their method can be integrated into the GP surrogate model as a plug-and-play feature, making it easy to extend our work.
\end{enumerate}

We did not compare against the following papers due to difficulty in aligning assumptions and similarity.
\begin{enumerate}
    \item \cite{hvarfner2022pi, hvarfner2024a, cisse2023hypbo}: These works assume that humans can explicitly express their beliefs as a probability distribution, such as a Gaussian distribution centered at the most promising location. This assumption is too strong and incompatible with our black-box assumption of human belief.
    \item \cite{gupta2023bo, av2024enhanced}: These methods are nearly identical to \cite{khoshvishkaie2023cooperative}. Therefore, we selected \cite{khoshvishkaie2023cooperative} as a representative work for this pessimistic approach.
    \item \cite{rodemann2024explaining}: This method is almost identical to \cite{av2022human}. Thus, we selected \cite{av2022human} as a representative work for this pessimistic approach.
\end{enumerate}

\rev{
\section{Comparison and Generalization to Other Feedback Forms.} \label{app:feedback}
\subsection{Other feedback forms}
\begin{enumerate}
    \item[(a)] \textbf{Pinpoint form:} \cite{av2022human, gupta2023bo, khoshvishkaie2023cooperative} adopt this form that the algorithm asks the humans to directly pinpoint the next query location.
    \item[(b)] \textbf{Pairwise comparison:} \cite{adachi2023looping} adopts this form that the algorithm presents paired candidates, and the human selects the preferred one.
    \item[(c)] \textbf{Ranking:} \cite{av2024enhanced} adopts this form that the algorithm proposes a list of candidates, and the human provides a preferential ranking.
    \item[(d)] \textbf{Belief function:} \cite{hvarfner2022pi, hvarfner2024a} adopt a Gaussian distribution as expert input. Unlike the others, this form assumes an offline setting where the input is defined at the beginning and remains unchanged during the optimization. Human experts must specify the mean and variance of the Gaussian, which represent their belief in the location of the global optimum and their confidence in this estimation, respectively.
\end{enumerate}

\subsection{Adaptation}
Slight modification can adapt these forms to our method.
\begin{enumerate}
    \item[(a)] \textbf{Pinpoint form:} We can simply replace the expert-augmented LCB in line~\ref{alg_line:colhbo_generate_xc} of Algorithm~\ref{alg:HAIBO} with the pinpointed candidate.
    \item[(b)] \textbf{Pairwise comparison:} By adopting the Bradley-Terry-Luce (BTL) model \cite{bradley1952rank}, we can extend our likelihood ratio model to incorporate preferential feedback. This allows us to obtain the surrogate , while the other parts of our algorithm remain unchanged.
    \item[(c)] \textbf{Ranking:} Ranking feedback can be decomposed into multiple pairwise comparisons. Therefore, we can apply the same method as in the pairwise comparison.
    \item[(d)] \textbf{Belief function:} We can use this Gaussian distribution model as the surrogate.
\end{enumerate}

\subsection{Comparison}
\begin{figure}%
    \centering
    \subfloat[\centering (a) Primal-dual vs pinpointing]{{\includegraphics[width=0.5\hsize]{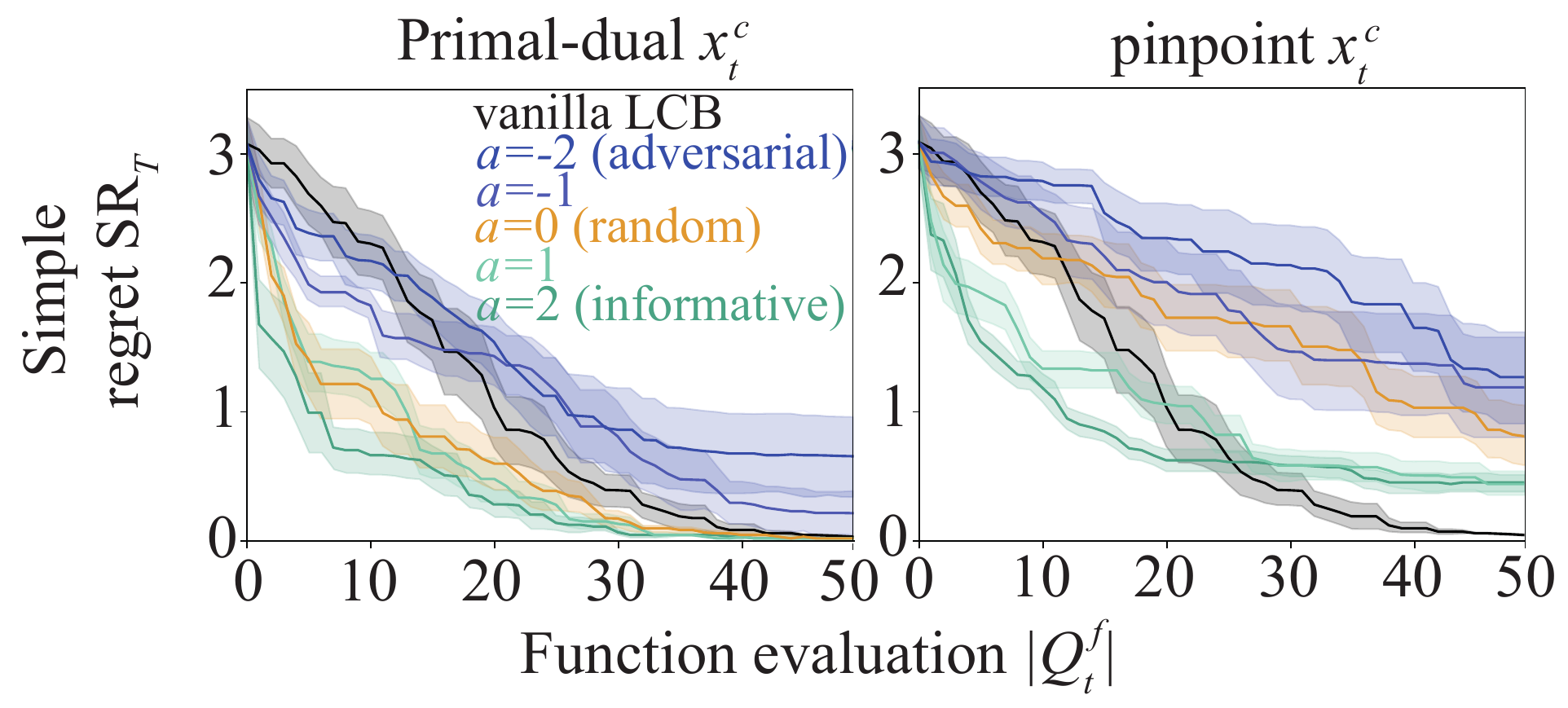} }}%
    \qquad
    \subfloat[\centering (b) MVN human belief model]{{\includegraphics[width=0.4\hsize]{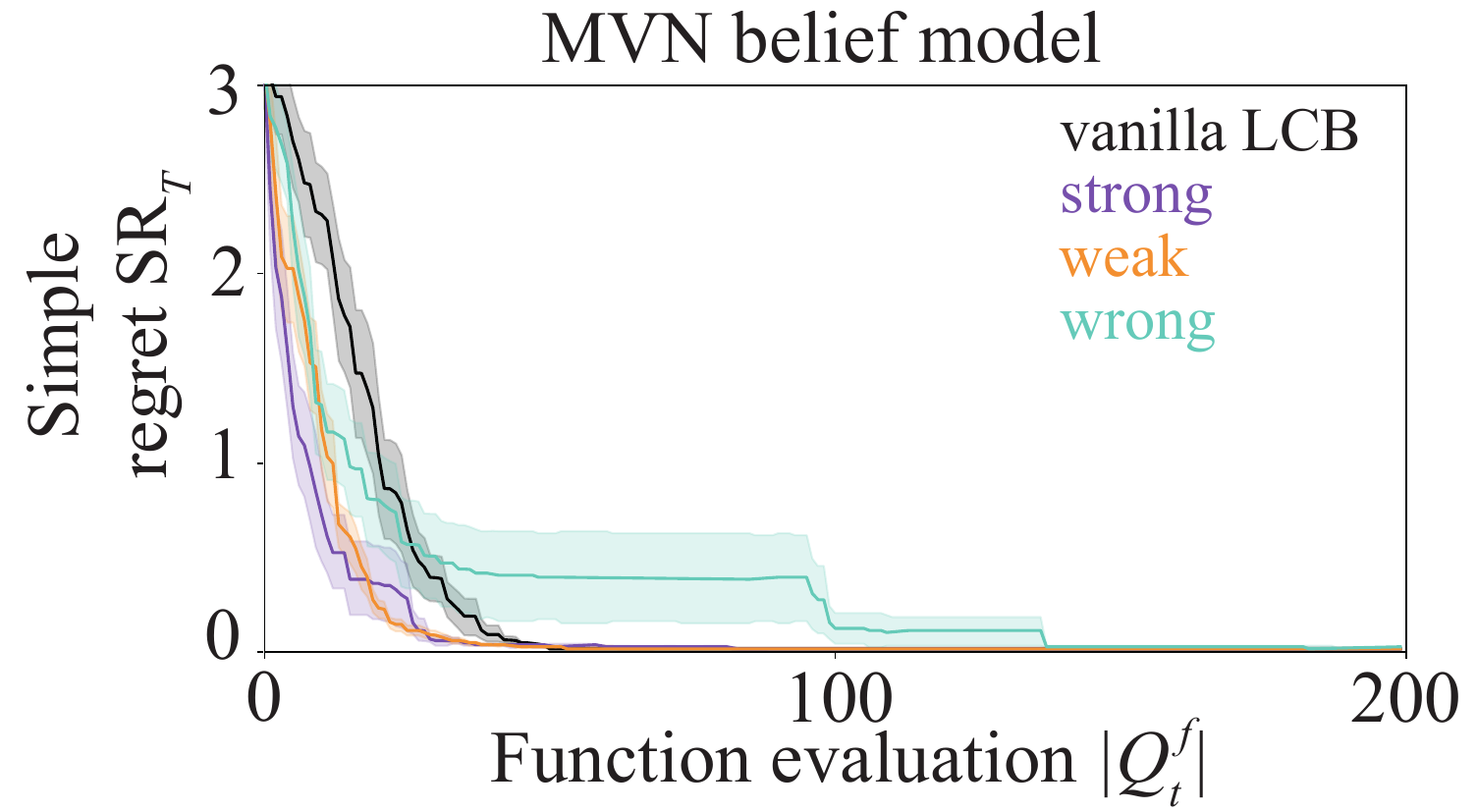} }}%
    \caption{Different forms of human feedback
    }%
    \label{fig:other_forms}
\end{figure}
We demonstrate the adaptation of (a) pinpoint and (d) belief function forms in Fig.~\ref{fig:other_forms}. 
The pinpoint strategy employs a sample from the expert belief function as $x_c$ on line~\ref{alg_line:colhbo_generate_xc} in Algorithm~\ref{alg:HAIBO}, while keeping the remaining lines the same as the original. It performs worse than the original primal-dual approaches, particularly in later iterations. This is because expert sampling does not incorporate GP information. Generally, humans excel at exploration in the beginning, while GP excels at finding precise locations in the later stages. This finding is supported by other literature, such as \cite{kanarik2023human}, involving human expert studies. 

In Fig.~\ref{fig:other_forms}(b), we employed the multivariate normal distribution (MVN) belief model proposed by \cite{hvarfner2022pi}. This model represents the human belief function as $\tilde p = \mathcal{N}(x; \mu, \Sigma)$, where $\mu$ is the mean vector representing the estimated location of the global optimum $x^*$, and $\Sigma$ is the covariance matrix, representing the confidence of the estimation. We use $\Sigma = \textbf{I}$, the identity matrix $\textbf{I}$, as suggested by \cite{hvarfner2022pi}. We transform: $[0, {|2\pi\Sigma|}^{-1/2}] \rightarrow [0,1]$, and we use this normalised belief function as the acceptance probability of a Bernoulli distribution $1 - p$ at given location $x$ (note that $p=0$ is acceptance). Following  \cite{hvarfner2022pi}, we set three levels of beliefs: strong, weak, and wrong. These levels are established by adjusting the mean vector to be offset from $x^*$. `Strong' aligns with $x^*$, `wrong' is the furthest possible location from $x^*$, and `weak' is an intermediate location. Our algorithm robustly converges for any level of trust.

As such, the primary reason we adopted binary labelling is due to its empirical success, as demonstrated in Fig.~\ref{fig:human} and Fig.~\ref{fig:other_forms}. None of the other formats, including (a) pinpoint form \cite{av2022human, khoshvishkaie2023cooperative} and (b) pairwise comparison \cite{adachi2023looping}, outperforms our method. In the experiments by \cite{adachi2023looping}, the authors showed that (a) pairwise comparison outperforms both (d) belief form \cite{hvarfner2022pi}. Therefore, it logically follows that our binary labeling format yields the best performance. 

The main reasons why the binary format works better are as follows:
\begin{enumerate}
    \item[(a)] \textbf{Pinpoint form:} The accuracy of pinpointing is generally lower than that of kernel-based models. Humans excel at qualitative comparison rather than estimating absolute quantities \cite{kahneman1979interpretation}. Numerous studies \cite{av2022human, kanarik2023human, khoshvishkaie2023cooperative, rodemann2024explaining} have confirmed that manual search (pinpointing) by human experts only outperforms in the initial stages, with standard BO with GP performing better in later rounds. \cite{gupta2023bo} shows that this type of feedback only outperforms when the expert's manual sampling is consistently superior to the standard BO. However, such cases are rare in our examples (e.g., Rosenbrock), and \cite{av2022human, rodemann2024explaining} corroborate this conclusion.
    \item[(b)] \textbf{Pairwise comparison:} This format relies on two critical assumptions: transitivity and completeness. Transitivity assumes no inconsistencies, which are often referred to as a "rock-paper-scissors" relationship. However, real-world human preferences frequently exhibit this issue \cite{chau2022learning}. Completeness assumes that humans can always rank their preferences at any given points. In practice, when a user is unsure which option is better, this assumption does not hold. Our imprecise probability approach avoids these issues by not relying on an absolute ranking structure \cite{augustin2014introduction, hullermeier2021aleatoric}.
    \item[(c)] \textbf{Ranking:} Ranking is an extension of pairwise comparison and has been classically researched as the Borda count, which is known not to satisfy all rational axioms. Theoretically, the Condorcet winner in pairwise comparison is the only method that is known to identify the global maximum of ordinal utility.
    \item[(d)] \textbf{Belief function:} This is another form of absolute quantity, which humans are generally not proficient at estimating. Additionally, the offline nature of this method does not allow for knowledge updates.
\end{enumerate}
}

\rev{
\section{Potential Extensions for Future Work}
\subsection{Extension to Time-varying Human Feedback Model}
\label{app:dynamic}
\begin{figure}
    \centering
    \includegraphics[width=0.95\textwidth]{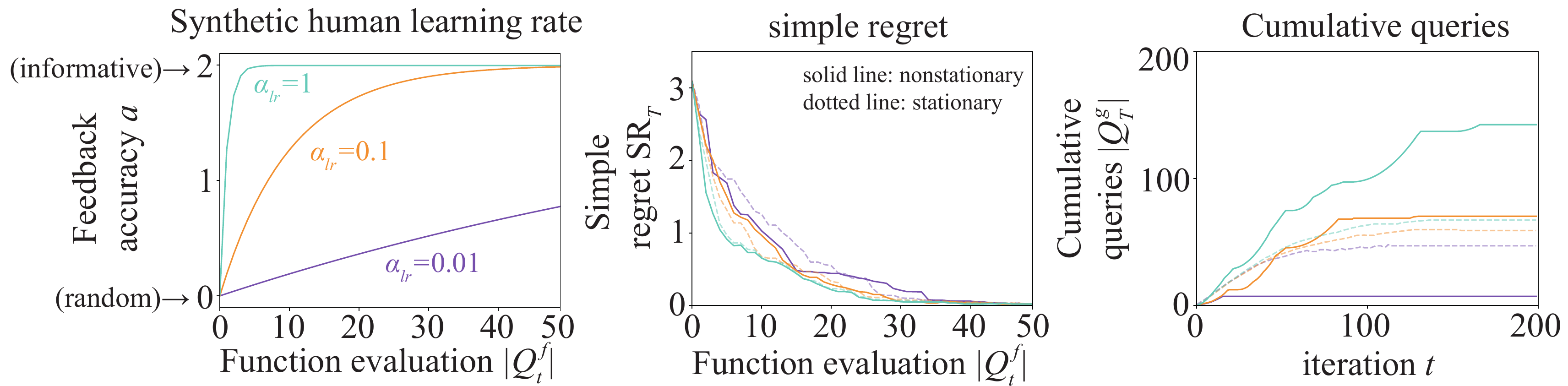}
    \caption{
    Non-stationary human accuracy.
    }
    \label{fig:nonstationary}
\end{figure}
In practice, human's belief in the black-box function may be influenced by the online evaluation results of the ground-truth black-box function. To further incorporate such online influence, we need to model the change of human feedback model. 

\textbf{Simple extension, yet not promising performance gain.} The most naïve approach for non-stationary model is windowing, i.e., forgetting the previous queried dataset. This can be very easy to apply to our setting, as it simply removes the old data outside the predefined iteration window. 

Fig.~\ref{fig:nonstationary} shows the scenario where the accuracy of human experts' labelling improves over time, represented by $a = 2(1 - \exp(-\alpha_{lr} / | Q_t^f |))$, where $\alpha_{lr}$ controls the learning rate. The non-stationary model employs windowing, retaining only the most recent $w$-th data points, with $w=5$. The stationary model does not use windowing, thereby retaining all labelled datasets. The plots represent the average of 10 runs without standard error for improved visibility. While simple regret showed slight improvement initially, the performance gain varied depending on $\alpha_{lr}$. In contrast, the cumulative number of queries $|Q_t^g|$ significantly increased due to the increased uncertainty introduced by windowing.

\textbf{More sophisticated extension.} Another more sophisticated approach is modelling the dynamics of behavioural change. A potential idea is modelling the human behaviour change as an implicit online learning process of the latent function $g$. That is, $g_{t+1}=F(g_t, x_t, y_t)$, where $g_t$ is the human latent function at step $t$. The forward dynamics $F$ captures the update of human latent function $g$ when observing the new data point. One potential $F$ is gradient ascent of log-likelihood as shown in $g_{t+1}=g_t+\lambda\nabla_g\log p_g(x_t, y_t)$, where $p_g(x_t, y_t)$ is the probability of observing $y_t$ at the input $x_t$ given the black-box objective function is $g$. We can then combine this dynamic with our likelihood ratio model. Since this part requires significantly different analysis and experiments, we leave it as future work.

\subsection{Extension to Adaptive Trust Weight $\eta$}
\label{app:adapt_eta}
In line~\ref{alg_line:no-harm guarantee} of Alg.~\ref{alg:cobohl}, the weight $\eta$ is fixed. An adaptive $\eta$ could offer better resilience to adversity. However, even without such a scheme, our no-harm guarantee holds, both theoretically and empirically. 


\textbf{Adaptation through the posterior standard deviation.} Although $\eta$ is set to be a constant in our current design of the algorithm, there is still adaptation on trusting human or the vanilla BO algorithm through the time-varying posterior standard deviation. Intuitively, if originally the expert-augmented solution $x_t^c$ is trusted more, more samples are allocated to human-preferred region and $\sigma_t(x_t^c)$ drops quickly. Intuitively, if we keep sampling $x_t^c$ and $x_t^u\neq x_t^c$, $\sigma_t(x_t^u)$ would finally be larger than $\eta\sigma_t(x_t^c)$ and we switch to sampling $x_t^u$. 

\textbf{Choice of $\eta$ does not need to be very large in practice.} Intuitively, $\eta$ captures the belief on the expertise level of the human. The more trust we have on the expertise of the human, the larger $\eta$ we can choose. But larger $\eta$ increases the risk of higher regret due to potential over-trust in adversarial human labeler. In our experience, $\eta$ does not need to be very large. Indeed, $\eta=3$ already achieves superior performance in our experiment (see Fig.~\ref{fig:robust}). 
\subsection{Extension to Different Acquisition Function}
\label{app:ext_acq}
Our algorithm can be easily extended to other acquisition functions. For example, we can indeed use similar idea to extend expected improvement (EI) acquisition function to human constrained expected improvement (HCEI) to generate $x_t^c$.
\begin{equation}
   x_t^c\in\arg\max_{x\in\mathcal{X}}\Prob(x \textrm{ is accepted by human})\mathrm{EI}(x). 
\end{equation}

}

\section{Experiments}\label{appendix:experiments}
\subsection{Hyperparameters}\label{app:hypers}
\begin{table}
    \centering
    \caption{The complete list of hyperparameters and their settings.}
    \label{tab:hypers}
    \resizebox{1\textwidth}{!}{
    \begin{tabular}{lccccccc}
    \toprule
         hyperparameters &
         initial value&
         data-driven optimisation?&
         tuning method
         \\
    \midrule
         $f$ kernel hyperparamters & BoTorch default & \ding{51} & maximising the marginal likelihood\\
         $g$ kernel hyperparamters & BoTorch default & \ding{51} & copying $f$ kernel values\\
         $r$ in Eq.\ref{eq:f_sigma_def} & 1e-4 & fixed & --\\
         $\gamma_{\aGt}^f$ in Eq.\ref{eq:max_inf_gain} & -- & \ding{51} & algorithm using \cite{hong2023optimization}\\
         $B_{f}$ in Lemma~\ref{lem:conf_int} & standardised (=1) & fixed & --\\
         $\sigma$ in Lemma~\ref{lem:conf_int} & $\sigma = r$ & fixed & --\\
         $\delta$ in Lemma~\ref{lem:conf_int} & 0.01 & fixed & --\\
         $\beta_{f_t}$ in Lemma~\ref{lem:conf_int} & 1 & \ding{51} & using the equation in Lemma~\ref{lem:conf_int}\\
         $\lambda_{t}$ in Eq.~\ref{eqn:reform_inner_prob_to_fin} & 1 & \ding{51} & using dual update in Eq.~\ref{eq:primal_dual}\\
         $\xi$ in Eq.~\ref{eq:primal_dual} & 0.02 & fixed & --\\
         $B_g$ in Eq.~\ref{eqn:reform_inner_prob_to_fin} & 1 & \ding{51} & the method in Appendix~\ref{sec:est_norm_bound}\\
         $\alpha_1$ in Eq.~\ref{eqn:reform_inner_prob_to_fin} & 0.01 & \ding{51} & the method in Appendix~\ref{sec:est_norm_bound}\\
         $\eta$ in line.~\ref{alg_line:no-harm guarantee} in Alg.~\ref{alg:HAIBO} & 3 & fixed & --\\
         $g_\text{thr}$ in line.~\ref{alg_line:no-elicitation guarantee} in Alg.~\ref{alg:HAIBO} & 1e-5 & fixed & --\\
    \bottomrule
    \end{tabular}
    }
\end{table}
We summarized the comprehensive list of hyperparameters used in this work and their settings in Table~\ref{tab:hypers}. Most of these are standard in typical GP-UCB approaches. The newly introduced hyperparameters are primarily tunable in a data-driven manner, and we provided a sensitivity analysis in the experiment section for those that are not.

\subsection{Synthetic Function Details}\label{sec:synthetic}
\subsubsection{Task Definitions}
\paragraph{Ackely}
Ackley funciton is defined as:
\begin{align}
f(x) := - a \exp \left[ -b \sqrt{\frac{1}{d} \sum_{i=1}^d x_i^2} \right] - \exp \left[
\frac{1}{d} \sum_{i=1}^d \cos (c x_i) \right] + a + \exp(1)
\end{align}
where $a = 20, c = 2\pi, d = 4$. We take the negative Ackley function as the objective of BO to make this optimisation problem maximisation. This is a 4-dimensional function bounded by $x \in [-1, 1]^d$. The global optimum is $x^* = [0,0,0,0]$ and $f(x^*) = 0$.

\paragraph{Hölder Table}
Hölder Table funciton is defined as:
\begin{align}
f(x) := \Bigg\lvert \sin(x_1) \cos(x_2) \exp\left( \Bigg\lvert 1 - \frac{\sqrt{x_1^2 + x_2^2}}{\pi} \Bigg\rvert \right) \Bigg\rvert
\end{align}
where $x_i$ is the $i$-th dimensional input. This is a 2-dimensional function bounded by $x \in [0, 10]^d$. The global optimum is $x^* = [8.05502, 9.66459]$ and $f(x^*) = 19.2085$.

\paragraph{Rastringin}
Rastringin function is defined as:
\begin{align}
f(x) := 10 d \sum_{i=1}^d \left[ x^2_i - 10 \cos(2 \pi x_i) \right]
\end{align}
where $x_i$ is the $i$-th dimensional input. This is a 2-dimensional function bounded by $x \in [-5.12, 5.12]^d$. The global optimum is $x^* = [0,0]$ and $f(x^*) = 0$.

\paragraph{Michalewicz}
Michalewicz funciton is defined as:
\begin{align}
f(x) := \sum_{i=1}^d \sin(x_i) \sin^{2m} \left(\frac{i x_i^2}{\pi} \right)
\end{align}
where $x_i$ is the $i$-th dimensional input and $m =10$. This is a 5-dimensional function bounded by $x \in [0, \pi]^d$. The global optimum is $f(x^*) = -4.687658$.

\paragraph{Rosenbrock}
Rosenbrock funciton is defined as:
\begin{align}
f(x) :=  \sum_{i=1}^{d-1} \left[ 100 (x_{i+1} - x_i^2)^2 + (x_i - 1)^2 \right]
\end{align}
where $x_i$ is the $i$-th dimensional input. This is a 3-dimensional function bounded by $x \in [-5, 10]^d$. The global optimum is $x^* = [1]^d$ and $f(x^*) = 0$.

\subsubsection{Computational time and elicitation efficiency}\label{app:synthetic_time}
\begin{figure}
  \centering
  \includegraphics[width=1\hsize]{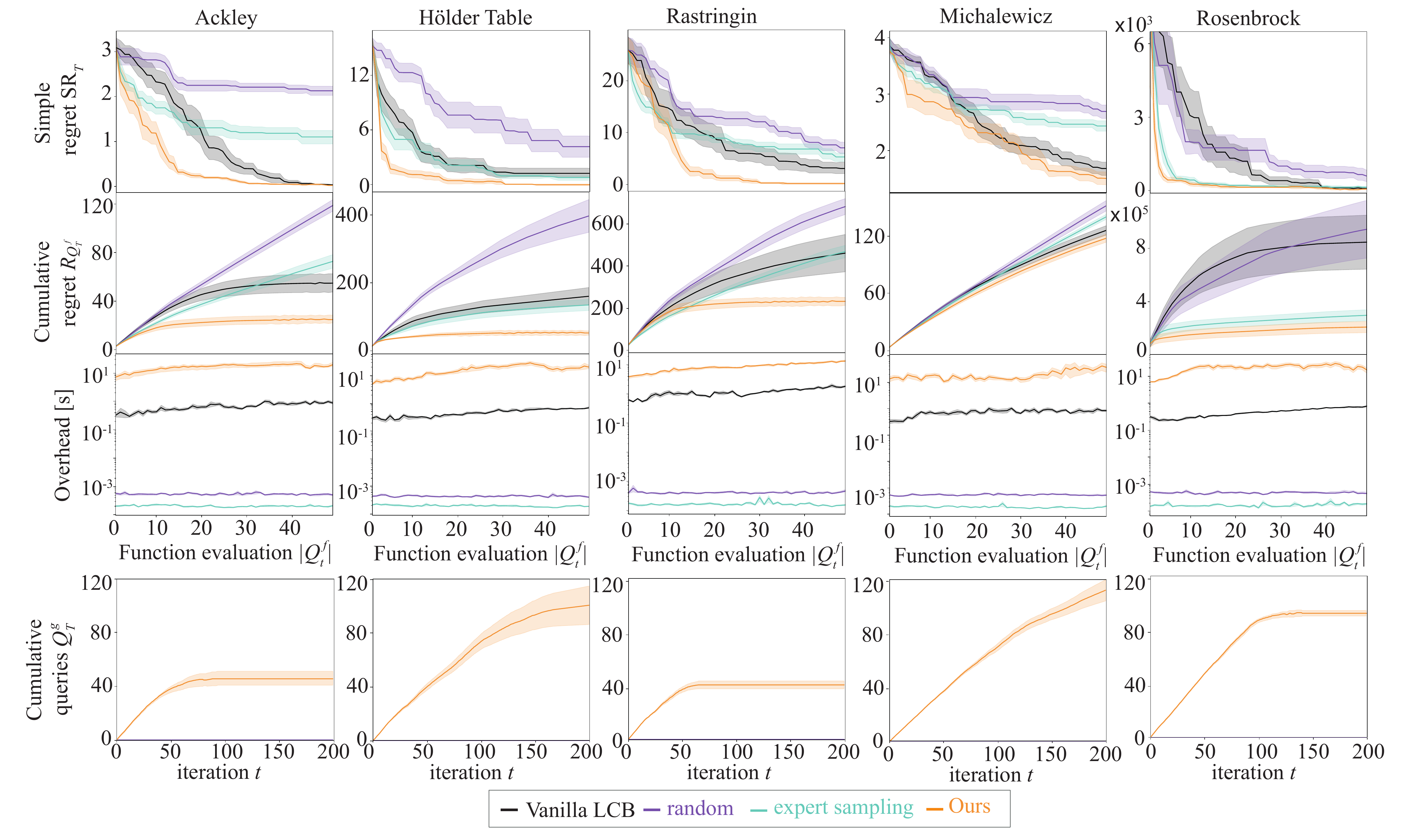}
  \caption{Simple and cumulative regrets, overhead, and cumulative queries for synthetic experiments.}
  \label{fig:synth_time}
  \vspace{-1em}
\end{figure}
Figure \ref{fig:synth_time} presents the comprehensive experimental results, including overhead and cumulative queries. Overhead refers to the wall-clock time in seconds required to generate the next query location. While the time taken to query the objective function is excluded, the time to query human (or synthetic) experts is included. Our overhead is the largest among the simple baselines; however, an average of around 10 seconds per query is reasonable when compared to more computationally expensive algorithms, such as information-theoretic acquisition functions, which typically require several hours per query. In most experiments, we observe a plateau in cumulative queries, indicating a handover guarantee. In the case of the Michalewicz function, a plateau has not yet been reached due to its high-dimensional nature. Nevertheless, we observe convergence acceleration in both simple and cumulative regrets.

\rev{
\subsubsection{Comprehensive check for no-harm guarantee}\label{app:no-harm}
\begin{figure}%
    \centering
    \includegraphics[width=1\hsize]{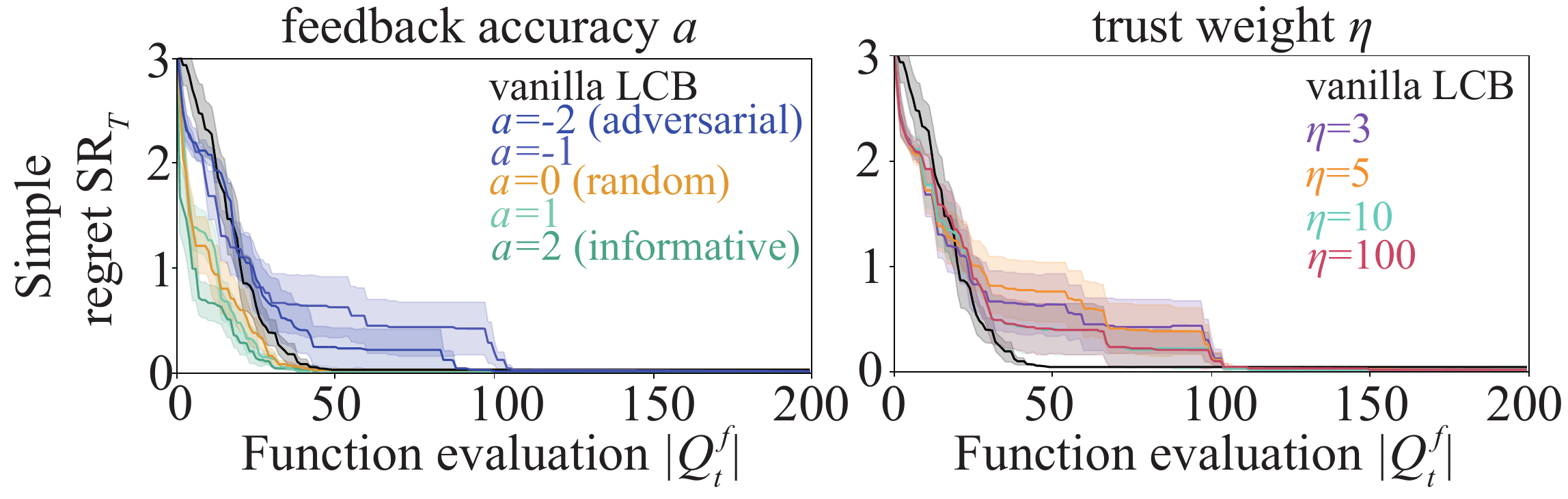}
    \caption{
    Confirming no-harm guarantee.
    }%
\end{figure}

We examine the no-harm guarantee by extending the iterations to confirm that our algorithm can converge at a rate comparable to the vanilla LCB. We tested with the two adversarial cases; (1) varying feedback accuracy $a \in \{-2, -1, 0, 1, 2\}$ for the fixed trust weight $\eta = 3$ and (2) varying trust weights $\eta \in \{ 3, 5, 10, 100\}$ for the fixed accuracy $a = -2$. Our algorithm converges to the same regret as the vanilla LCB over multiple iterations in both cases. We observed saturation behavior, where the convergence drop starts at similar locations among larger $\eta$, indicating that the no-harm guarantee is assured regardless of how large $\eta$ becomes. Particularly, the convergence curves of $\eta=10$ and $\eta = 100$ are almost identical, supporting the saturation perspective.
}

\subsection{Human experiment details}\label{sec:human}
\subsubsection{Task definitions}
The task involves identifying the optimal electrolyte material combination to maximize ionic conductivity in lithium-ion batteries. Ionic conductivity is crucial for reducing internal resistance, which is essential for fast charging. Slow charging remains one of the biggest challenges for the widespread adoption of electric vehicles. Therefore, finding the best electrolyte combination is crucial to advancing electric vehicle development and realizing a sustainable society.

In our study, we considered four types of electrolyte materials. For demonstration purposes, we did not conduct physical experiments. Instead, we utilized an open dataset and fitted functions to interpolate between data points, creating a continuous search space. Experiments were then performed on this synthetic data using software and four human experts. In real-world development, researchers and engineers synthesize these materials, which is expensive, making the expert's labeling process significantly cheaper than objective queries.

\paragraph{Li$^+$ standard design}
The first task involves the EC-DMC-EMC-LiPF$_6$ system \cite{dave2022autonomous, adachi2023looping}, where EC, DMC, and EMC are ethylene carbonate, dimethyl carbonate, and ethyl methyl carbonate, respectively, and LiPF$_6$ is lithium hexafluorophosphate. Ionic conductivity depends on both lithium salt molarity and cosolvent composition. Using the dataset from \cite{dave2022autonomous}, we fitted the Casteel-Amis equation \cite{casteel1972specific} and extended it to a continuous space. The input features are (1) LiPF$_6$ molarity, (2) DMC vs. EMC cosolvent ratio, and (3) EC vs. carbonates cosolvent ratio, with inputs bounded as $x_1 \in [0, 2]$, $x_2 \in [0, 1]$, and $x_3 \in [0, 1]$. The output is generated by adding i.i.d. zero-mean Gaussian noise with a variance of 1 to the noiseless function. We take the negative of the ionic conductivity in log mS/cm as the minimization objective.

\paragraph{Li$^+$ methyl-acetate}
The second task involves the MA-DMC-EMC-LiPF$_6$ system \cite{logan2018study, adachi2023looping}, with MA being methyl acetate. Using the dataset from \cite{logan2018study}, we fitted the Casteel-Amis equation and extended it to continuous space. The input features are (1) LiPF$_6$ molarity, (2) DMC vs. EMC cosolvent ratio, and (3) MA vs. carbonates cosolvent ratio, with inputs bounded as $x_1 \in [0, 2]$, $x_2 \in [0, 1]$, and $x_3 \in [0, 1]$. The output is generated by adding i.i.d. zero-mean Gaussian noise with a variance of 1 to the noiseless function. We take the negative of the ionic conductivity in log mS/cm as the minimization objective.

\paragraph{Li$^+$ polymer-nanocomposite}
The third task involves the PEO-LLZTO nanocomposite electrolyte system \cite{zhang2016flexible}, where PEO is polyethylene oxide, and LLZTO is lithium garnet (Li$_6.4$La$_3$Zr$_1.4$Ta$_0.6$O$_12$) nanoparticles. Using the dataset from\cite{zhang2016flexible}, we fitted a GP model and extended it to continuous space. The input features are (1) PEO volume \%, (2) LLZTO volume \%, and (3) LLZTO particle size in micrometers, with inputs bounded as $x_1 \in [70, 95]$, $x_2 \in [5, 30]$, and $x_3 \in [0.04, 10]$. The output is generated by adding i.i.d. zero-mean Gaussian noise with a variance of 1 to the noiseless function. We take the negative of the ionic conductivity in log mS/cm as the minimization objective.

\paragraph{Li$^+$ Ionic liquid}
The fourth task involves the bmimSCN-LiClO$_4$-LiTFSI ionic liquid \cite{rosol2009solubility}, where bmimSCN is 1-butyl-3-methylimidazolium thiocyanate, LiClO$_4$ is lithium perchlorate, and LiTFSI is lithium bis(trifluoromethanesulfonyl)imide. Using the dataset from \cite{rosol2009solubility}, we fitted a GP model and extended it to continuous space. The input features are (1) LiClO$_4$ molarity, (2) LiTFSI molarity, and (3) bmimSCN molarity, with inputs bounded as $x_1 \in [0, 4]$, $x_2 \in [0, 1.5]$, and $x_3 \in [3, 5]$. The output is generated by adding i.i.d. zero-mean Gaussian noise with a variance of 1 to the noiseless function. We take the negative of the ionic conductivity in log mS/cm as the minimization objective.

\subsection{How Do Human Experts Reason?}\label{app:reason}
We explore how experts reason through these optimization tasks. Ionic conductivity is roughly estimated by the product of movable ion density and diffusivity, as described by the Nernst-Einstein equation. Experts base their evaluations on this relationship.

\paragraph{Li$^+$ standard design}
In this system, EC plays a crucial role in both factors. LiPF$_6$ provides movable ions (Li$^+$ and PF$_6^-$), but these ions are not mobile in their raw state due to strong electrostatic forces. EC, a highly polarized but non-charged solvent, dissolves LiPF$_6$ through solvation. Increasing EC concentration can raise movable ion density, but EC's high viscosity slows diffusivity, creating a convex curve. Experts generally agree that the global maximum is around 30\% EC and 1 M LiPF$_6$, but the optimal EMC/DMC ratio remains uncertain. EMC and DMC are similar, with EMC being larger and asymmetric, and DMC being smaller and symmetric. Smaller molecules tend to be more diffusive, so a higher DMC ratio is expected to be better, although the asymmetric structure of EMC could disrupt higher-order solvation networks, contributing to diffusivity.

In summary, experts vaguely know the whole function shape and possible global optimum location for two variables, yet others are unknown.

\paragraph{Li$^+$ methyl-acetate}
his task involves replacing EC with MA from the first task, making the overall system similar. However, MA is an unusual material, and none of the participants are familiar with it. We will explain how experts reasoned this change in the optimization task.

EC plays a central role in dissolving LiPF$_6$, increasing movable ion density, although it is viscous. While no one knows methyl acetate, it can be inferred that it also dissolves LiPF$_6$. The challenge lies in determining its polarization ability and viscosity. EC is a planar molecule with a five-membered ring, resembling a `small sheet magnet' with strong magnetic power but easy stacking. Conversely, MA is a small, non-ring-structured, asymmetric molecule. This asymmetry prevents MA molecules from stacking, enhancing diffusivity. However, the asymmetry also reduces polarization, leading to a weaker solvation effect and lower movable ion density.

Thus, MA has a mix of positive and negative effects, making it difficult for experts to predict the exact shape of the convex curve. Nonetheless, in most "less viscous" solvent systems, the peak typically occurs around 1.5 M of LiPF$_6$. Experts can roughly estimate this position, and this estimation is fairly accurate, as the true position is at 1.35 M.

\paragraph{Li$^+$ polymer-nanocomposite}
This task is completely different from the previous two tasks. Our electrolyte is now solid-state rather than liquid, so the Nernst-Einstein equation may not be applicable. However, the core idea remains the same. PEO is a framework material without ionic conductivity, whereas LLZTO has ionic conductivity. Generally, a higher LLZTO content should result in greater conductivity. Other factors are less certain.

We can anticipate the effects in both directions. Smaller particle sizes might be better because they distribute more evenly within the PEO, increasing ionic conductive paths. However, smaller particles might also be worse due to increased grain boundaries and aggregation caused by electric forces. Thus, most experts expected a convex relationship with particle size and a monotonic increase with LLZTO ratio.

In reality, experimental results showed that conductivity improved monotonically with smaller particle sizes and displayed a convex relationship with LLZTO volume. Therefore, the experts' advice was somewhat inaccurate.

Looking back, experts were partially correct. Aggregation did create the convex shape in LLZTO volume ratio, indicating their understanding of the phenomenon. However, they did not identify the correct input dimension where aggregation mattered. For particle size, the thorough mixing procedure with ball milling used in the dataset prevented aggregation, leading to misconceptions about the function shape.

\paragraph{Na$^+$ Ionic liquid}
This task is completely different from the previous tasks. Although our electrolyte is liquid, all materials are ionically conductive. As the name suggests, ionic liquids are special materials that can dissolve themselves without the need for a cosolvent. Consequently, the movable ion density factor remains almost unchanged, as all components are conductive regardless of composition. Therefore, diffusivity becomes the dominant factor. Diffusivity primarily depends on two factors: molecule size and electric interaction. Smaller molecules are generally more mobile, but they also have stronger electric interactions when the charge is the same (all ions in this system are monovalent).

This dual dependence leads to different expectations: if size is the dominant factor, smaller molecules (like LiCl$_4$) are expected to perform best. Conversely, if electric interaction is dominant, the results will differ.

Most experts anticipated a monotonic change in all dimensions, expecting both LiCl$_4$ and LiTFSI to show increased performance due to their smaller size compared to bmimSCN. However, experimental results showed a double peak shape for LiTFSI vs. bmimSCN and a convex shape for LiCl$_4$ and bmimSCN. Thus, the experts' advice was inaccurate.
The real physical phenomena were more complex than initially thought, with electric interactions playing a more dominant role.

\subsubsection{Computational time and elicitation efficiency}\label{app:real_time}
\begin{figure}
  \centering
  \includegraphics[width=1\hsize]{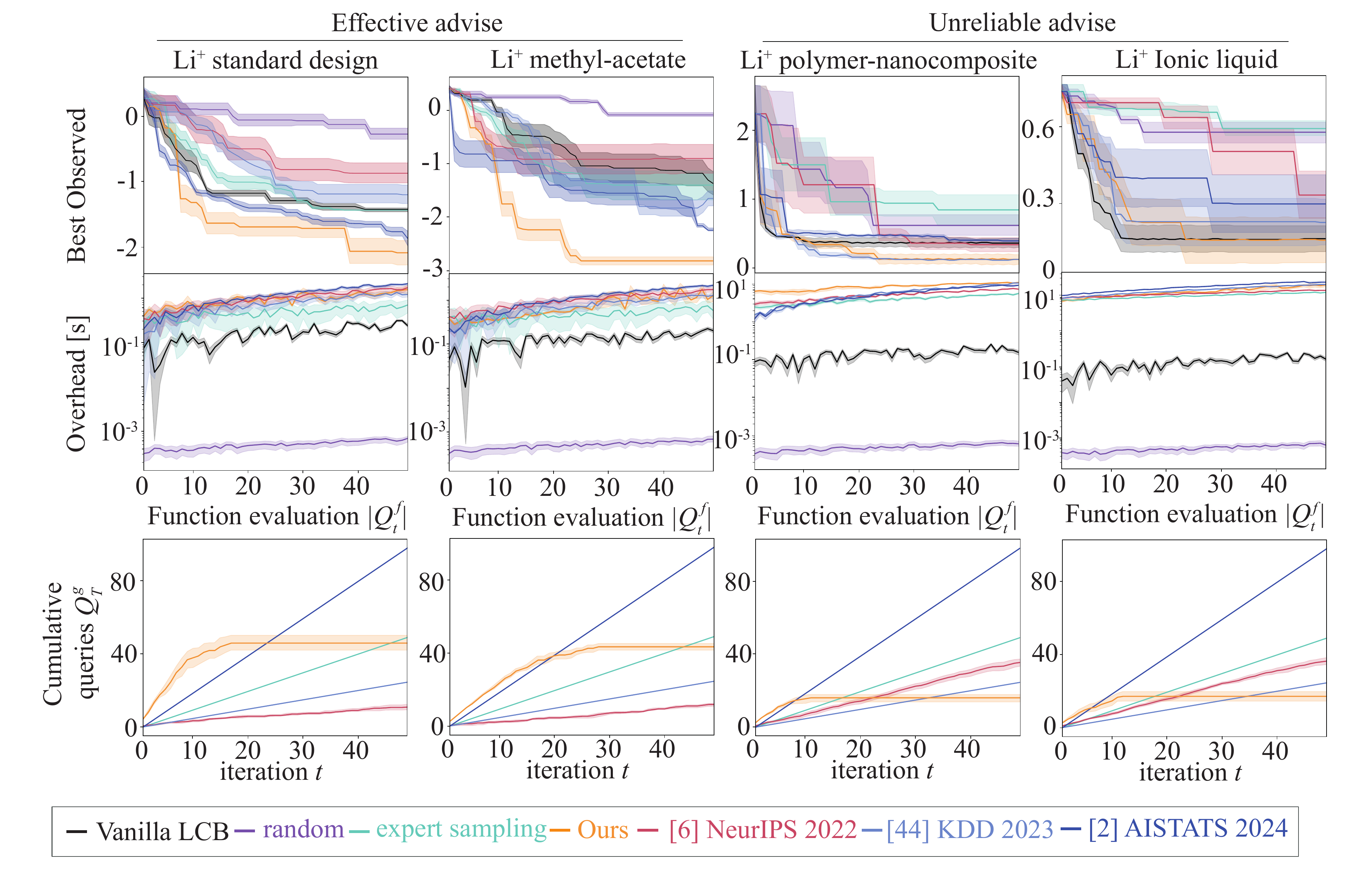}
  \caption{Simple and cumulative regrets, overhead, and cumulative queries for real-world experiments.}
  \label{fig:real_time}
  \vspace{-1em}
\end{figure}

Figure~\ref{fig:real_time} illustrates the full experimental results, including the best observed values $\max Y_{\aGT}$, overhead, and cumulative queries $\Ct$. The overhead definition remains consistent with that in the synthetic experiments. Note that these experiments include only four human trials, resulting in noisier data compared to the synthetic experiments, which used 10 random seeds. The overhead for our method and the baselines is approximately the same, around 10 seconds per query. This is manageable compared to the significantly slower methods, such as information-theoretic acquisition functions, which take several hours per query.

Regarding cumulative queries, only our method demonstrates a handover guarantee. While baseline methods continue to request human intervention even as the experiments conclude, our method stops requesting input midway through the experiments, thereby freeing the human expert from the task. Our approach allows for more effective input from experts in cases where their advice is beneficial and reduces input in unreliable cases. In contrast, the baselines request input regardless of the quality of the advice. Notably, the method described in \cite{av2022human} increases the frequency of requests when experts provide incorrect information. This occurs because disagreements between the surrogate $f$ and human beliefs prompt human experts to intervene, aiming to prevent the BO from proceeding in the wrong direction. Unfortunately, this intervention can act as an adversarial response. In contrast, our algorithm avoids such scenarios through active learning constraints (as highlighted in line~\ref{alg_line:no-harm guarantee}), thus achieving a no-harm guarantee in unreliable cases.


\newpage
\section*{NeurIPS Paper Checklist}

\begin{enumerate}

\item {\bf Claims}
    \item[] Question: Do the main claims made in the abstract and introduction accurately reflect the paper's contributions and scope?
    \item[] Answer: \answerYes{} 
    \item[] Justification: Elicitation efficiency and no-harm guarantee are proved in Theorem~\ref{thm:RQ_bound}. Experiments shows empirical efficacy in Figures~\ref{fig:robust}, \ref{fig:synth}, \ref{fig:human}, \ref{fig:synth_time}, \ref{fig:real_time}.

\item {\bf Limitations}
    \item[] Question: Does the paper discuss the limitations of the work performed by the authors?
    \item[] Answer: \answerYes{} 
    \item[] Justification: See Conclusion and Limitation section.

\item {\bf Theory Assumptions and Proofs}
    \item[] Question: For each theoretical result, does the paper provide the full set of assumptions and a complete (and correct) proof?
    \item[] Answer: \answerYes{}
    \item[] Justification: Assumptions 2.1-2.6, Proofs in Appendices A-C.

    \item {\bf Experimental Result Reproducibility}
    \item[] Question: Does the paper fully disclose all the information needed to reproduce the main experimental results of the paper to the extent that it affects the main claims and/or conclusions of the paper (regardless of whether the code and data are provided or not)?
    \item[] Answer: \answerYes{}
    \item[] Justification: Experimental section, Appendix~\ref{appendix:experiments}, open-sourced code (anonymous) \url{https://github.com/ma921/COBOL/}

\item {\bf Open access to data and code}
    \item[] Question: Does the paper provide open access to the data and code, with sufficient instructions to faithfully reproduce the main experimental results, as described in supplemental material?
    \item[] Answer: \answerYes{}
    \item[] Justification: Provide data and code on Anonymised repository \url{https://github.com/ma921/COBOL/}

\item {\bf Experimental Setting/Details}
    \item[] Question: Does the paper specify all the training and test details (e.g., data splits, hyperparameters, how they were chosen, type of optimizer, etc.) necessary to understand the results?
    \item[] Answer: \answerYes{}
    \item[] Justification: See experimental section and Appendix~\ref{appendix:experiments}.

\item {\bf Experiment Statistical Significance}
    \item[] Question: Does the paper report error bars suitably and correctly defined or other appropriate information about the statistical significance of the experiments?
    \item[] Answer: \answerYes{}
    \item[] Justification: All experiments report the $\pm$ standard errors.

\item {\bf Experiments Compute Resources}
    \item[] Question: For each experiment, does the paper provide sufficient information on the computer resources (type of compute workers, memory, time of execution) needed to reproduce the experiments?
    \item[] Answer: \answerYes{}
    \item[] Justification: See experimental section and footnote 2. See also Appendix~\ref{appendix:experiments}.
    
\item {\bf Code Of Ethics}
    \item[] Question: Does the research conducted in the paper conform, in every respect, with the NeurIPS Code of Ethics \url{https://neurips.cc/public/EthicsGuidelines}?
    \item[] Answer: \answerYes{}
    \item[] Justification: Our research is pure algorithm development.

\item {\bf Broader Impacts}
    \item[] Question: Does the paper discuss both potential positive societal impacts and negative societal impacts of the work performed?
    \item[] Answer: \answerYes{}
    \item[] Justification: See Conclusion and Limitation.
    
\item {\bf Safeguards}
    \item[] Question: Does the paper describe safeguards that have been put in place for responsible release of data or models that have a high risk for misuse (e.g., pretrained language models, image generators, or scraped datasets)?
    \item[] Answer: \answerNA{}.
    \item[] Justification: Our paper is a pure algorithm study for blackbox optimization for small, lower dimensional tasks.

\item {\bf Licenses for existing assets}
    \item[] Question: Are the creators or original owners of assets (e.g., code, data, models), used in the paper, properly credited and are the license and terms of use explicitly mentioned and properly respected?
    \item[] Answer: \answerYes{}
    \item[] Justification: See experimental section.

\item {\bf New Assets}
    \item[] Question: Are new assets introduced in the paper well documented and is the documentation provided alongside the assets?
    \item[] Answer: \answerYes{}
    \item[] Justification: Documents are available on Anonymised repository \url{https://anonymous.4open.science/r/COBOL-9B8B/}

\item {\bf Crowdsourcing and Research with Human Subjects}
    \item[] Question: For crowdsourcing experiments and research with human subjects, does the paper include the full text of instructions given to participants and screenshots, if applicable, as well as details about compensation (if any)? 
    \item[] Answer: \answerYes{}
    \item[] Justification: See experimental section and Appendix~\ref{appendix:experiments}.

\item {\bf Institutional Review Board (IRB) Approvals or Equivalent for Research with Human Subjects}
    \item[] Question: Does the paper describe potential risks incurred by study participants, whether such risks were disclosed to the subjects, and whether Institutional Review Board (IRB) approvals (or an equivalent approval/review based on the requirements of your country or institution) were obtained?
    \item[] Answer: \answerYes{}
    \item[] Justification: 
    While we do not have IRB approval, our institution has reviewed and approved our study as low-risk followed by the protocol indicated in \url{https://researchsupport.admin.ox.ac.uk/governance/ethics/apply}, considering that all experiments involve running software on open-source datasets. According to the NeurIPS 2024 ethics guidelines, adherence to existing protocols at the authors' institution is required, with IRB approval being just one form of such protocols. Our institution follows its own policy, and we adhered to the standard procedure. 

\end{enumerate}

\end{document}